\documentclass[10pt,twocolumn,letterpaper]{article}

\usepackage[pagenumbers]{cvpr} 

\usepackage[dvipsnames]{xcolor}

\definecolor{lightblue}{rgb}{0.2, 0.5, 0.9}

\usepackage{graphicx}
\usepackage{amsmath}
\usepackage{amssymb}
\usepackage{booktabs}

\usepackage{times}
\usepackage{epsfig}
\usepackage{graphicx}
\usepackage{amsmath}
\usepackage{amssymb}
\usepackage{color}
\usepackage{paralist} 
\usepackage{enumitem}
\usepackage{colortbl}
\usepackage{nicefrac}
\usepackage{multirow}
\usepackage{booktabs}
\usepackage{makecell}
\usepackage{gensymb}
\usepackage{microtype}
\usepackage{cancel}
\usepackage{wasysym}

\usepackage[accsupp]{axessibility}  

\definecolor{cvprblue}{rgb}{0.21,0.49,0.74}
\usepackage[pagebackref,breaklinks,colorlinks,citecolor=cvprblue]{hyperref}

\usepackage[capitalize]{cleveref}
\crefname{section}{Sec.}{Secs.}
\Crefname{section}{Section}{Sections}
\Crefname{table}{Table}{Tables}
\crefname{table}{Tab.}{Tabs.}

\newcommand\blfootnote[1]{%
  \begingroup
  \renewcommand\thefootnote{}\footnote{#1}%
  \addtocounter{footnote}{-1}%
  \endgroup
}

\title{Restoration by Generation with Constrained Priors}

\author{Zheng Ding$^{1\dagger}$, Xuaner Zhang$^{2}$, Zhuowen Tu$^{1}$, Zhihao Xia$^{2}$\\
$^{1}$UC San Diego \quad \quad $^{2}$Adobe
}

\begin{document}
\twocolumn[{%
\renewcommand\twocolumn[1][]{#1}%
\maketitle
\vspace{-2.5em}
\begin{center}
    \centering
    \captionsetup{type=figure}
    \includegraphics[width=\textwidth]{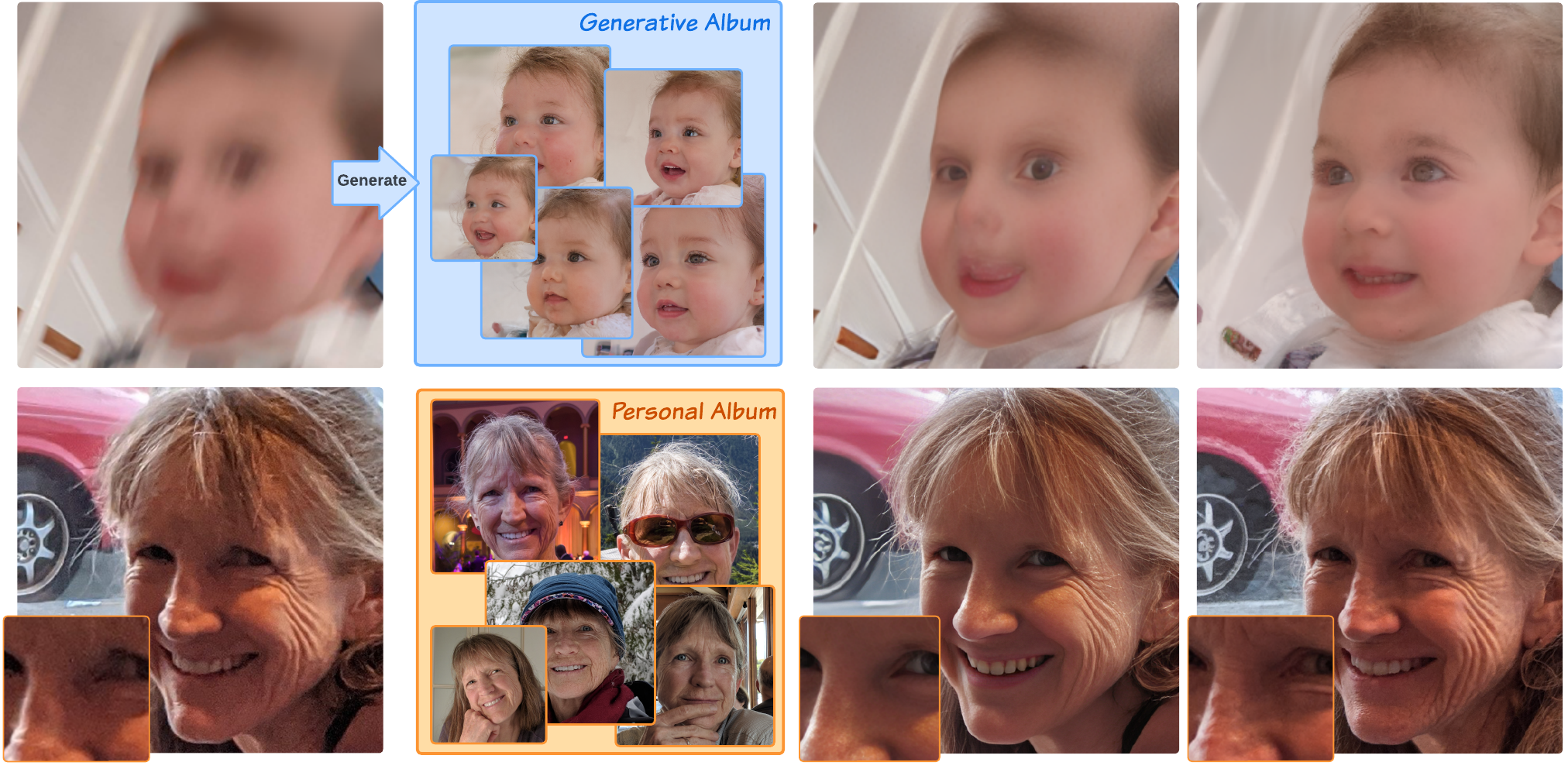}
    \begin{tabular}{cccc}
         \hspace{3mm} Input \hspace{29mm} & Anchor \hspace{26mm} & CodeFormer~\cite{zhou2022towards} \hspace{25mm} & Ours
    \end{tabular}
    \vspace{-2mm}
    \captionof{figure}{We harness the generative capacity of a diffusion model for image restoration. By constraining the generative space with a generative or personal album, we can directly use a pre-trained diffusion model to produce a high-quality and realistic image that is also faithful to the input identity. Without any assumption on the degradation type, we are able to generalize to real-world images that exhibit complicated degradation. We compare our restoration result with CodeFormer, a state-of-the-art baseline~\cite{zhou2022towards}. Our method generalizes better to different types of degradation while more faithfully preserving the input identity. Images are best viewed zoomed in on a big screen.}
    \label{fig:teaser}
\end{center}%
}]

\blfootnote{$\dagger$ Work done during an internship at Adobe.}

\begin{abstract}
\vspace{-3mm}

The inherent generative power of denoising diffusion models makes them well-suited for image restoration tasks where the objective is to find the optimal high-quality image within the generative space that closely resembles the input image.
We propose a method to adapt a pretrained diffusion model for image restoration by simply adding noise to the input image to be restored and then denoise. Our method is based on the observation that the space of a generative model needs to be constrained. We impose this constraint by finetuning the generative model with a set of anchor images that capture the characteristics of the input image. With the constrained space, we can then leverage the sampling strategy used for generation to do image restoration. We evaluate against previous methods and show superior performances on multiple real-world restoration datasets in preserving identity and image quality. 
We also demonstrate an important and practical application on personalized restoration, where we use a personal album as the anchor images to constrain the generative space. This approach allows us to produce results that accurately preserve high-frequency details, which previous works are unable to do. Project webpage: \href{https://gen2res.github.io}{https://gen2res.github.io}.

\end{abstract}

\vspace{-5mm}
\section{Introduction}
\label{sec:intro}

Image restoration involves recovering a high-quality natural image $x$ from its degraded observation $y=H(x)$ is a fundamental task in low-level vision. The challenge lies in finding a solution that 1) matches the observation through a set of degradation steps; and 2) aligns with the distribution of $x$. In scenarios where the degradation process $H$ is unknown, the problem becomes a blind image restoration problem.

Discriminative learning approaches \cite{gu2022vqfr, zhou2022towards, wang2021towards, yang2021gan} aim to solve this inverse problem directly by training an inverse model $F(y)$, typically a neural network, using datasets of low- and high-quality image pairs ${(x, y)}$. However, the trained model is limited to restoring images with degradations $H$ present in the training set. This limitation places the burden of generalization on the construction of the training set. The effectiveness of these methods also heavily depends on the capacity of the inversion model and the characteristics of the loss function. 
Model-based optimization methods \cite{tv, zhang2021plug, red, kawar2022denoising, chung2022diffusion}, on the other hand, assume that the degradation model is only known at inference time. They focus on learning the image prior $p(x)$, which can be represented as regularization terms \cite{tv}, denoising networks \cite{zhang2017learning, red}, or more recently pre-trained diffusion models \cite{kawar2022denoising, chung2022diffusion}. However, these methods generally assume that the degradation process is known at inference time, limiting their practicality and often relegating them to synthetic evaluations.

In this paper, we adopt a markedly different approach to the image restoration problem. 
We observe that humans are able to recognize a degraded image (i.e., a `bad photo') and envision a fix without knowing the imperfections in the image formation process. Such insights rely on our inherent understanding of what constitutes a high-quality image. 
Building on this observation, we propose to approach image restoration using the recent success of large generative models, which possess the capacity of forming high-quality imagery.
Unlike prior works, we do not make any assumption on the degradation process. Our method solely relies on a well-trained denoising diffusion model.

The challenges then arise in how to project the input image into the generative process given the models are trained on mostly clean images. And once projected, how to constrain the generation to preserve the useful features in the input, e.g., the identity. We address the input projection by adding Gaussian noise to the low-quality image to be restored, matching the distribution of clean images added with noise. Once projected, we can then denoise the image as is normally done in the generation process of a diffusion model. To handle the second challenge of preserving useful signals in the input, we propose to constrain the generative space by finetuning the model with anchor images that share characteristic features with the low-quality input. When the anchor is given, such as from an album of other photos of the same identity, we can simply finetune the model with the provided images. When the anchor is missing, as in most single-image restoration scenarios, we propose to use a generative album as the anchor. The generative album is a set of clean images generated from the diffusion model with the low-quality input image imposing soft guidance, and thus closely resembles the input image.

\begin{figure}[tph]
    \centering
        \includegraphics[width=1.05\linewidth]{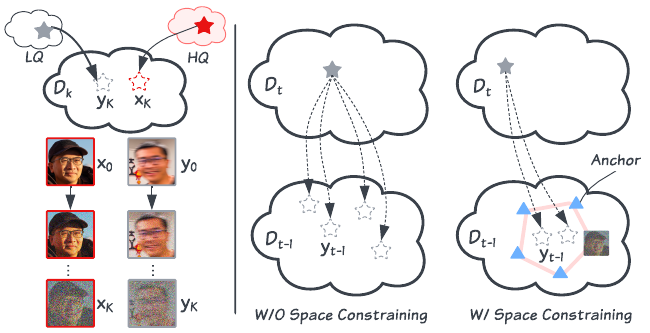}
    \vspace{-2em}
    \caption{\textbf{Left: Image projection.} When sufficient Gaussian noise is added to the low- and high-quality image, we can bring them to the same distribution. The low-quality image can thus be denoised with a pre-trained diffusion model. \textbf{Right: With and without space constraining.} A regular diffusion step lands $y_t$ in an arbitrary position in the generative space; with space constraining, the path of generation becomes more constrained towards the space defined by the anchor images.}
    \label{fig:motivation}
    \vspace{-4mm}
\end{figure}

Surprisingly, we find that our straightforward approach yields high-quality results on blind image restoration. Unlike previous methods, our approach does not rely on paired training data or assumptions about the degradation process. 
It thus generalizes well to real-world images with unknown degradation types, such as noise, motion blur, and low resolution. By effectively harnessing the generative capacity of a pre-trained diffusion model, our generation-based restoration approach produces high-quality and realistic images that are faithful to the input identity.

\section{Related Works}
\label{sec:formatting}

\paragraph{Supervised Learning for Image Restoration.} The trend of leveraging advanced neural network architectures for image restoration has spanned from CNNs \cite{DnCNN, cho2021rethinking, zhang2018ffdnet, tian2020attention, zhang2019residual} to GANs \cite{srgan, kupyn2018deblurgan, kupyn2019deblurgan}, and more recently, to transformers \cite{restormer, swinir, wang2022uformer} and diffusion models \cite{saharia2022image, whang2022deblurring, saharia2022palette}. One aspect remains unchanged: these methods are trained on datasets comprising pairs of high-quality and low-quality images. Typically, these image pairs are synthetically generated, depicting a single type of degradation, leading to task-specific models for denoising \cite{DnCNN, zhang2018ffdnet, xia2020identifying, tian2020attention, guo2019toward}, deblurring \cite{kupyn2018deblurgan, kupyn2019deblurgan, whang2022deblurring, cho2021rethinking}, or super-resolution \cite{srgan, wang2021real, saharia2022image}. However, they fall short when applied to real-world low-quality images, which often suffer from diverse, unknown degradations.

In specific domains, particularly with facial images, numerous works have focused on training blind restoration models that simulate various degradation types during training. For instance, GFPGAN \cite{wang2021towards} and GPEN \cite{yang2021gan} enhance pretrained GAN networks with modules to leverage generative priors for blind face restoration. Recent approaches like CodeFormer \cite{zhou2022towards}, VQFR \cite{gu2022vqfr} and RestoreFormer\cite{restormer} exploit the low-dimensional space of facial images to achieve impressive results. Emerging works have also started building upon the success of diffusion models \cite{ddpm, ddim, adm}. For example, IDM \cite{zhao2023towards} trains a conditional diffusion model for face image restoration by injecting low-quality images at different layers of the model. Conversely, DR2 \cite{wang2023dr2} combines the generative capabilities of pre-trained diffusion models with existing face restoration networks. Another line of works \cite{li2018learning, li2020enhanced} seeks to enhance the results by incorporating additional information present in a guide image or photo album, which is often available in practice. Nevertheless, these methods rely on a synthetic data pipeline for training, which limits their generalizability. Diverging from these methodologies, our approach does not use paired data, synthetic or real, allowing it to generalize naturally to real data without succumbing to artifacts.
\vspace{-4mm}

\paragraph{Model-based Image Restoration.} Unlike supervised learning methods, model-based methods often form a posterior of the underlying clean image given the degraded image, with a likelihood term from the degradation process and an image prior. \cite{zhang2021plug, red} proposed using denoising networks as the image prior. These priors are integrated with the known degradation process during inference, and the Maximum A Posteriori (MAP) problem is addressed through approximate iterative optimization methods. DGP \cite{pan2021exploiting} proposes image restoration through GAN inversion, searching for a latent code that generates an image closely matching the input image after processing it through the known degradation. The recent success of pre-trained foundational diffusion models has inspired works \cite{kadkhodaie2021stochastic, kawar2021snips, choi2021ilvr, kawar2021stochastic} to utilize diffusion models as such priors. Kawar \etal \cite{kawar2022denoising} and Wang \etal \cite{wang2022zero} proposed an unsupervised posterior sampling method using a pre-trained denoising diffusion model to solve linear inverse problems.  Chung \etal\cite{chung2022diffusion} extends diffusion solvers to general noise inverse problems. Despite these advancements, these methods generally assume that the degradation process is known at inference, limiting their practicality to synthetic evaluations. In contrast, our method does not assume any knowledge of the degradation model at training or inference. 
\vspace{-4mm}

\paragraph{Personalized Diffusion Models.} Personalization methods aim to adapt pre-trained diffusion models to specific subjects or concepts by leveraging data unique to the target case. In text-to-image synthesis, many works opt for customization by fine-tuning with personalized data, adapting token embeddings of visual concepts \cite{gal2022image, gal2023encoder}, the entire denoising network \cite{dreambooth}, or a subset of the network \cite{kumari2023multi}. Recent studies \cite{jia2023taming, shi2023instantbooth, xiao2023fastcomposer} propose bypassing per-object optimization by training an encoder to extract embeddings of subject identity and injecting them into the diffusion model's sampling process. In other domains, 
DiffusionRig \cite{diffusionrig} learns personalized facial editing by fine-tuning a 3D-aware diffusion model on a personal album. In this work, we demonstrate that a personalized diffusion model represents a constrained generative space, directly usable for sampling high-quality images to restore images of a specific subject, without additional complexities. For \textit{single-image restoration}, unlike previous instance-based personalization methods \cite{jia2023taming, shi2023instantbooth, xiao2023fastcomposer}, we generate an album of images close to the input and then constrain the diffusion model using this generative album. This approach enables restoration by directly sampling from the fine-tuned model, eliminating the need for guidance.

\begin{figure*}[ht]
    \centering
        \includegraphics[width=0.95\textwidth]{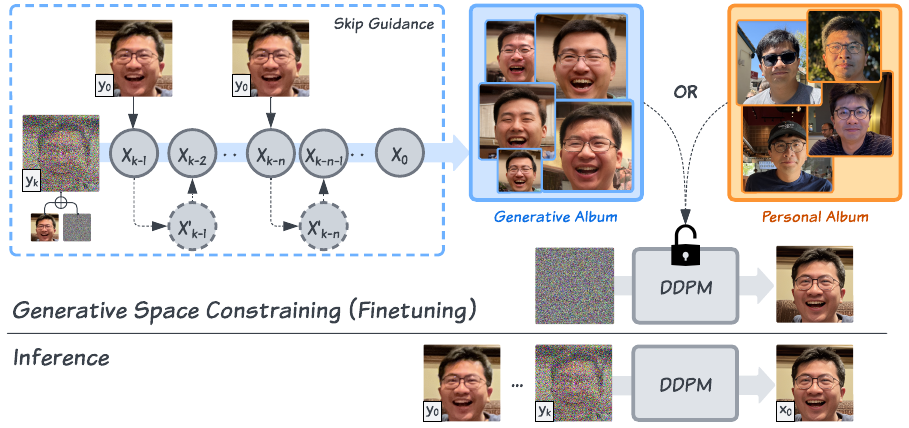}
    \caption{\textbf{An illustration of our finetuning and inference stage.} The core of our method is to constrain the generative space by fine-tuning a pre-trained diffusion model with either a generative album or a personal album. The generative album is generated from the input low-quality image with skip guidance to loosely follow the characteristics of the input. Once the generative space is constrained, at inference time, we can simply add noise to the input low-quality image and pass it through the diffusion model to do restoration.}
    \label{fig:pipeline}
    \vspace{-3mm}
\end{figure*}

\section{Method}

\subsection{Preliminaries}
A diffusion model approximates its training image distribution $p_\theta(x_0)$ by learning a model $\theta$ that effectively reverses the process of adding noise. The commonly used Denoising Diffusion Probabilistic Models (DDPM) gradually introduce Gaussian noise into a clean image $x_0$:
\begin{equation}
\label{eq:xt}
\vspace{-2mm}
x_t = \sqrt{\alpha_t}x_0 + \sqrt{1-\alpha_t}\epsilon, \quad \text{where} \quad \epsilon \sim \mathcal{N}(\mathbf{0}, \mathbf{I})
\end{equation}
The reverse generative process aims to progressively denoise $x_t$ until it is free from noise.
Once a diffusion model is trained, for any given time $t$ and the corresponding noisy image $x_t$, it can iteratively denoise by sampling from $p(x_0 | x_t)$ using the trained model.

The objective of image restoration, on the other hand, is to recover the latent high-quality image $x_0$ from a low-quality, partially observed image $y_0$. Contrary to previous methods that decompose the posterior distribution into the likelihood $p(y_0 | x_0)$ and the prior $p(x_0)$ to solve a MAP problem, we propose to recover the complete observation by directly sampling from the posterior:
\vspace{-2mm}
\begin{equation}
\hat{x} \sim p(x_0 | y_0)
\end{equation}

\subsection{Restoration by Generation}

We aim to maximally leverage the generative capacity of the diffusion model by using its iterative sampling process for restoration.
A critical observation underlies this approach: when sufficient Gaussian noise is added to the degraded observation $y_0$, the resultant image $y_t$:
\begin{equation}
\label{eq:yt}
y_t = \sqrt{\alpha_t}y_0 + \sqrt{1-\alpha_t}\epsilon, \quad \text{where} \quad \epsilon \sim \mathcal{N}(\mathbf{0}, \mathbf{I})
\end{equation}
becomes indistinguishable from the underlying clean image $x_0$ with the same noise. That is, there exists a large enough $K$ such that 
\vspace{-2mm}
\begin{equation}
y_K \approx x_K
\end{equation}
This phenomenon becomes apparent from Eq~\ref{eq:xt} and~\ref{eq:yt} as $\alpha$ decreases and when the same noise $\epsilon$ is sampled. It is also demonstrated in Fig~\ref{fig:motivation}, where adding noise to high-quality and low-quality images can progressively align their distributions, making them more similar over time, this suggests:
\vspace{-2mm}
\begin{equation}
p(x_0 | y_K) \approx p(x_0 | x_K)
\vspace{-2mm}
\end{equation}
Based on this observation, we can sample a clean image $x_0$ from $p(x_0 | y_K)$ using the same sampling process as from $p(x_0 | x_K)$; in other words, we can denoise $y_K$ iteratively directly with the pre-trained diffusion model. Since the sampling process remains unchanged, the resultant image should match the quality of the images generated from the original diffusion model. 

We find it critical to select the optimal time $K$, which determines the amount of noise added to the low-quality input image to start the sampling process. If too little noise is added, the discrepancy between $x_K$ and $y_K$ becomes large, yielding low-quality samples as $y_K$ does not align with the training distribution $p(x_K)$ of the diffusion model. 
On the other hand, with too excessive noise added, the original contents in the input $y_K$ are hardly discernible. The generated sample, though with high quality, will not be faithful to the input.
We aim to produce high-quality samples, while mitigating the information loss, and achieve so by constraining the generative space of the pre-trained diffusion model.

\begin{table*}[h]
\centering
\scalebox{0.9}{
\begin{tabular}{lcc|cc|cc|cc}
\hline
\toprule
& \multicolumn{2}{c|}{\textbf{Wider-Test}} & \multicolumn{2}{c|}{\textbf{WebPhoto-Test}} & \multicolumn{2}{c|}{\textbf{LFW-Test}} &  \multicolumn{2}{c}{\textbf{Deblur-Test}} \\
 & FID $\downarrow$ & MUSIQ $\uparrow$ & FID $\downarrow$ & MUSIQ $\uparrow$  & FID $\downarrow$ & MUSIQ $\uparrow$ & FID $\downarrow$ & MUSIQ $\uparrow$ \\
 \hline
Input & 183.03 & 15.68 & 161.82 & 20.26  & 131.68 & 27.51 & 169.43 & 27.53 \\
GFPGAN\cite{wang2021towards} & 59.38 & 56.48 & 114.15 & 55.13  & 64.10 & 60.46 & 178.40 & 58.03\\
CodeFormer\cite{zhou2022towards} & 48.57 & 55.70 &  98.55 & 55.20 &  66.31 & 58.72 & 163.47 & 57.09 \\
VQFR\cite{gu2022vqfr} & 52.64 & 54.23 &  105.94 & 52.44 &  63.73 & 57.52 & 168.36& 54.45\\
DR2(+VQFR)\cite{wang2023dr2} & 69.40 & 53.62 & 143.96 & 51.92 & 67.70 & 57.42 & 173.33 & 55.34 \\
\hline
Ours & \textbf{46.38} & \textbf{58.73} &  \textbf{96.44} & \textbf{57.71} &  \textbf{56.32} & \textbf{60.68} & \textbf{135.33} & \textbf{60.20} \\
\bottomrule
\hline
\end{tabular}
}
\caption{\textbf{Quantitative comparison on real-world single-image blind face restoration on four datasets.}}
\vspace{-4mm}
\label{tab:standard}
\end{table*}

\subsection{Generative Space Constraining}

The loss of information is inherent in the diffusion process. Due to the stochasticity of the forward Markov chain, the clean image generated using the reverse process from $x_t$ may not match the original $x_0$.
The larger $t$ is, the larger the generative space $p(x_0|x_t)$ spans. The learned score functions guide $x_t$ to the clean image space without constraining its content.
This property is desirable for a generative model where the diversity of generation is valued.
However, this is not ideal for image restoration where the input contents also need to be preserved. The goal is thus to constrain the generative space to a small subspace that tightly surrounds the underlying clean image.

We propose to use a set of anchor images to fine-tune the diffusion model, thus imposing the generative space. 
These anchor images can be given in the form of a \textit{personal album}, or be generated as a \textit{generative album} in the common scenario of single image restoration.
\vspace{-3mm}

\paragraph{Personal Album as Additional Information.} In many real-world scenarios, additional information about the underlying clean image beyond a single degraded observation is available, such as an album of different clean images of the same subject. We personalize the pre-trained diffusion model in this case — fine-tuning it with the personal album. This approach naturally addresses the ill-posed nature of single-image restoration, producing results containing authentic high-frequency details absent in the degraded observation. This is demonstrated in identity preservation in face restoration tasks (Sec~\ref{subsec:personal}).
\vspace{-4mm}

\paragraph{Generative Album from a Single Degraded Observation.} For single-image restoration, due to its ill-posed nature, we can only constrain the generative space to a subspace of high-quality realistic images close to the degraded observation. To generate this album of high-quality images, we follow approaches similar to previous works on guided image generation \cite{chung2022diffusion, song2023loss, Bansal_2023_CVPR}. Specifically, given a degraded image $y_0$, we first add noise $\epsilon_K$ to obtain $y_K$, then denoise it progressively with the pre-trained diffusion model. For the denoised image $x_t$, we apply a simple $L_1$ guidance that computes the distance between the input degraded image and the generated image:
\vspace{-2mm}
\begin{equation}
x_t' = x_t - \lambda \nabla_{x_t}||y_0-\hat{x}_{0,t}||_2^2
\end{equation}
Unlike previous methods where the guidance needs to be strongly followed, our guidance, the low-quality input, is an approximation. Instead of applying the guidance at every step \cite{song2023loss, chung2022diffusion}, we propose to apply this approximated guidance periodically at every $n$ steps. The proposed \textit{Skip Guidance} enforces the generated image to loosely follow the information in the degraded input while retaining the quality of images in the generative steps. We repeat this process multiple times to generate a set of images that form a generative album, which is used to fine-tune the diffusion model.

Once the diffusion model is fine-tuned with a personal or generative album, we restore a degraded image $y_0$ by adding noise $\epsilon_K$. Then, we iteratively denoise $y_K$ using the fine-tuned model for $K$ steps, \textit{without} further guidance. Notably, our approach does not rely on paired data for training and makes no assumptions about the degradation process at training or inference.

\begin{figure*}[ht]
    \centering
    \setlength{\tabcolsep}{1pt}
    \def\imW{0.15\linewidth}
    \begin{tabular}{cccccc}
        \raisebox{-.0\height}{\includegraphics[width=\imW]{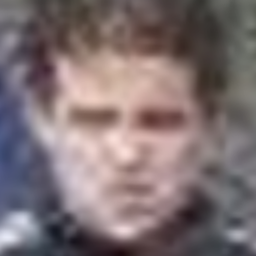}} &
        \raisebox{-.0\height}{\includegraphics[width=\imW]{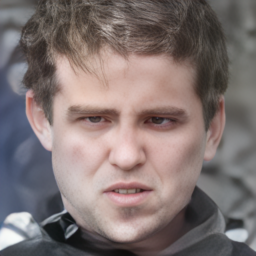}} &
        \raisebox{-.0\height}{\includegraphics[width=\imW]{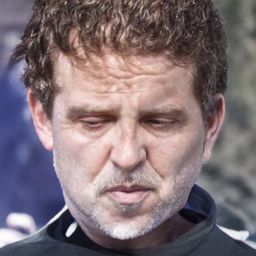}} &
        \raisebox{-.0\height}{\includegraphics[width=\imW]{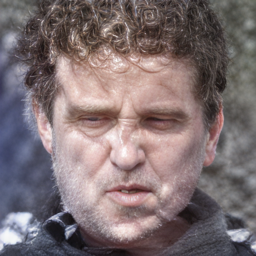}} &
        \raisebox{-.0\height}{\includegraphics[width=\imW]{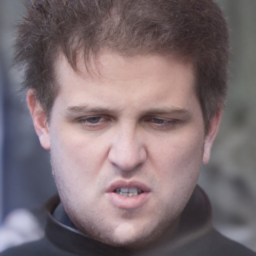}} &
        \raisebox{-.0\height}{\includegraphics[width=\imW]{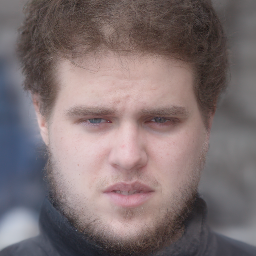}} \\
        \raisebox{-.0\height}{\includegraphics[width=\imW]{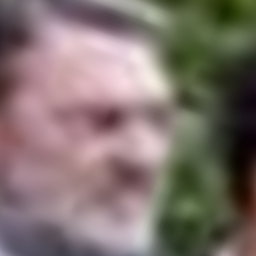}} &
        \raisebox{-.0\height}{\includegraphics[width=\imW]{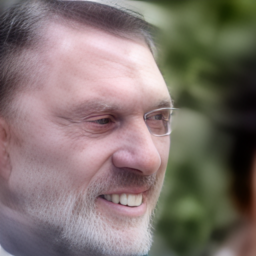}} &
        \raisebox{-.0\height}{\includegraphics[width=\imW]{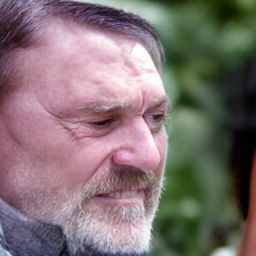}} &
        \raisebox{-.0\height}{\includegraphics[width=\imW]{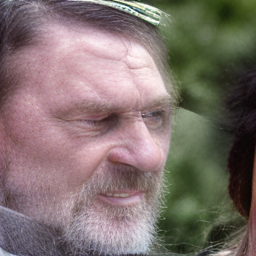}} &
        \raisebox{-.0\height}{\includegraphics[width=\imW]{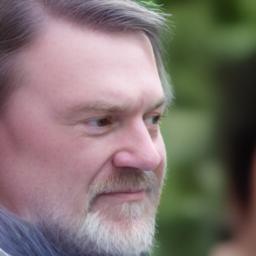}} &
        \raisebox{-.0\height}{\includegraphics[width=\imW]{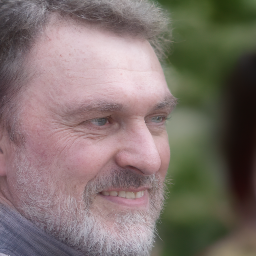}} \\
        \raisebox{-.0\height}{\includegraphics[width=\imW]{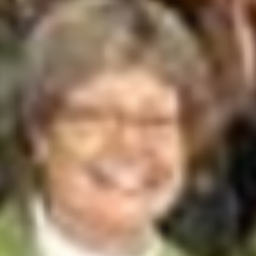}} &
        \raisebox{-.0\height}{\includegraphics[width=\imW]{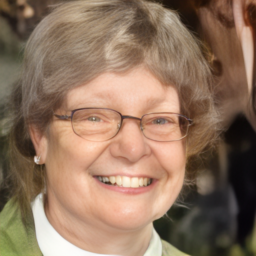}} &
        \raisebox{-.0\height}{\includegraphics[width=\imW]{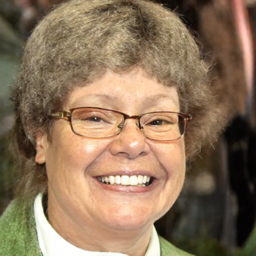}} &
        \raisebox{-.0\height}{\includegraphics[width=\imW]{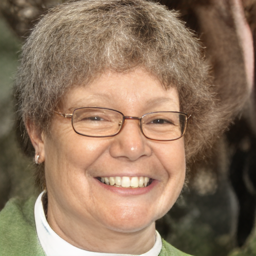}} &
        \raisebox{-.0\height}{\includegraphics[width=\imW]{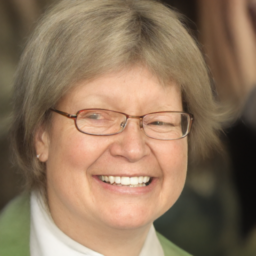}} &
        \raisebox{-.0\height}{\includegraphics[width=\imW]{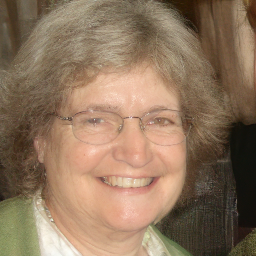}} \\
        \raisebox{-.0\height}{\includegraphics[width=\imW]{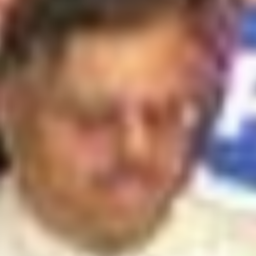}} &
        \raisebox{-.0\height}{\includegraphics[width=\imW]{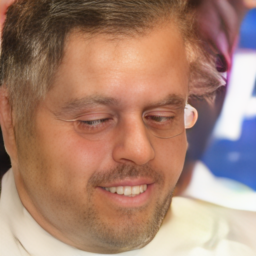}} &
        \raisebox{-.0\height}{\includegraphics[width=\imW]{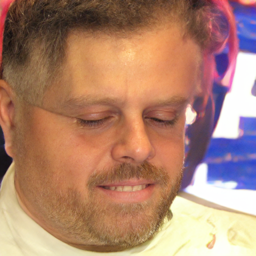}} &
        \raisebox{-.0\height}{\includegraphics[width=\imW]{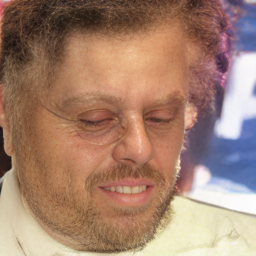}} &
        \raisebox{-.0\height}{\includegraphics[width=\imW]{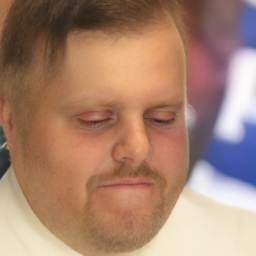}} &
        \raisebox{-.0\height}{\includegraphics[width=\imW]{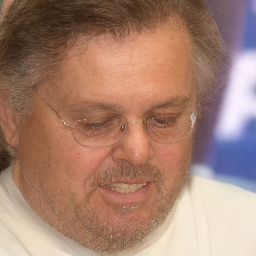}} \\
        Input & GFPGAN & VQFR & CodeFormer & DR2(+VQFR)  & Ours \\
    \end{tabular}
    \vspace{-7pt} 
    \caption{\textbf{Qualitative comparison with baselines on Wider-Test.} With strong generative capacity of the diffusion model, our method performs well on severely degraded images. We are able to produce high-quality and realistic images while prior works suffer from unrealistic artifacts.}
    \vspace{-4mm} 
    \label{fig:standard}
\end{figure*}

\section{Experiments}

With the core observation that generation can be directly applied for restoration, our method requires only a pre-trained unconditional diffusion model and is applicable to any image domain for which the diffusion model has been trained. We first show results of our restoration-by-generation approach on the standard task of single-image blind face restoration in Sec~\ref{subsec:blind}. 
In Sec~\ref{subsec:personal}, we extend our approach to personalized face restoration. Here, the objective is to restore a degraded image of a subject using other clean images of the same identity. 
Sec~\ref{subsec:pets} presents the adaptation of our method to different image categories, such as dogs and cats, by simply swapping the pre-trained diffusion model. Notably, as our method does not presume any specific form of degradation, all our evaluations are conducted on real images with unknown degradation.

\subsection{Blind Face Restoration with Generative Album}
\label{subsec:blind}

For the task of single-image blind face restoration, we utilize an unconditional diffusion model pretrained on the FFHQ dataset \cite{ffhq}. We first assess our approach on three widely-used real-world face benchmarks with degradation levels ranging from heavy to mild: Wider-Test (970 images) \cite{zhou2022towards}, LFW-Test (1771 images) \cite{wang2021towards}, and Webphoto-Test (407 images) \cite{wang2021towards}. These datasets are collections of in-the-wild images aligned using the method employed in FFHQ \cite{ffhq}.

Our approach uses a \textit{generative album} as the anchor for restoring these in-the-wild images. For each input low-quality image, we generate 16 images with skip guidance to form the album. We then fine-tune the diffusion model using this album to constrain the generative space. The process involves adding noise to the input low-quality image and denoising it for $K$ steps with the fine-tuned model, where $K = 200$. The model is fine-tuned for 3,000 iterations with a batch size of 4 and a learning rate of 1e-5.

We benchmark our method against state-of-the-art supervised alternatives for blind face restoration, including the GAN-based GFPGAN \cite{wang2021towards}, two codebook-based approaches (Codeformer \cite{zhou2022towards} and VQFR \cite{gu2022vqfr}), and a diffusion-based approach DR2 \cite{wang2023dr2}. Except for DR2, which combines a diffusion model with the pretrained supervised face restoration model VQFR \cite{gu2022vqfr}, all methods utilize supervised training with synthetic low-quality images from FFHQ.

Quantitative and qualitative results are provided. For the former, we use FID \cite{fid} and MUSIQ(Koniq) \cite{musiq} as metrics following CodeFormer~\cite{zhou2022towards}. The quantitative scores are in Table~\ref{tab:standard}. Previous methods, except for DR2 \cite{wang2023dr2}, are trained on FFHQ-512$\times$512 for restoration. For a fair comparison, we downsize the outputs of these methods to 256$\times$256 for metric calculation. Our results surpass all previous methods in terms of FID and MUSIQ across all datasets, despite not undergoing a supervised training approach for image restoration. Qualitative comparisons in Figure~\ref{fig:standard} illustrate that our method produces high-quality restoration results akin to those from an unconditional diffusion model, even with severely degraded input images.

\begin{figure}[ht]
    \centering
    \setlength{\tabcolsep}{1pt}
    \def\imW{0.25\linewidth}
    \scalebox{0.95}{
    \begin{tabular}{cccc}
        \raisebox{-.0\height}{\includegraphics[width=\imW]{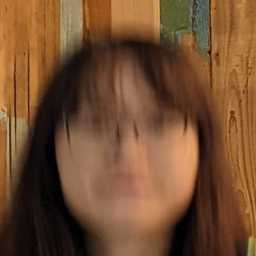}} &
        \raisebox{-.0\height}{\includegraphics[width=\imW]{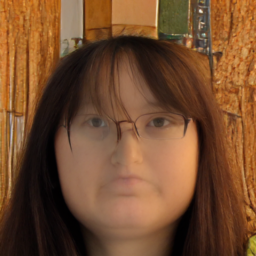}} &
        \raisebox{-.0\height}{\includegraphics[width=\imW]{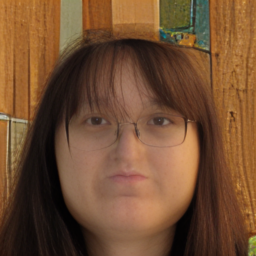}} &
        \raisebox{-.0\height}{\includegraphics[width=\imW]{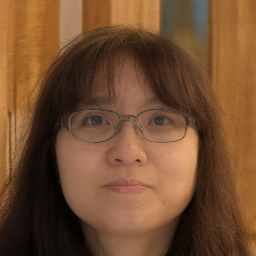}} \\
        \raisebox{-.0\height}{\includegraphics[width=\imW]{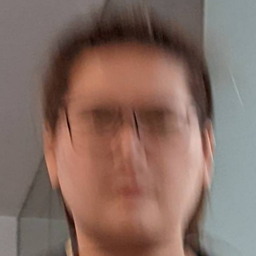}} &
        \raisebox{-.0\height}{\includegraphics[width=\imW]{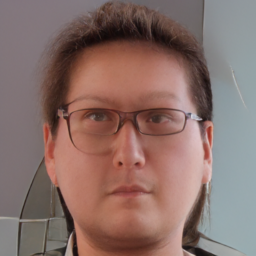}} &
        \raisebox{-.0\height}{\includegraphics[width=\imW]{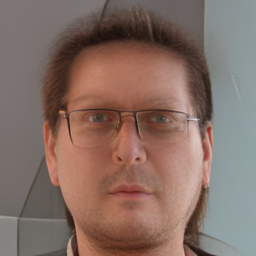}} &
        \raisebox{-.0\height}{\includegraphics[width=\imW]{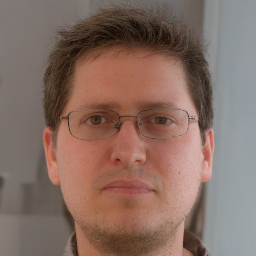}} \\
        Input & VQFR & CodeFormer  & Ours \\
    \end{tabular}
    }
    \vspace{-7pt}
    \caption{\textbf{Comparison with previous methods on Deblur-Test.} Previous methods do not include motion blur as part of the degradation simulation for training, and thus fail to restore the images. In contrast, our method does not make assumptions on the degradation types and generalizes more robustly.}
    \vspace{-5mm} 
    \label{fig:deblur}
\end{figure}

Our method's agnosticism to the degradation process leads to superior generalization capabilities. To further demonstrate this, we constructed a motion blur dataset (Deblur-Test) by selecting 67 images from \cite{lai2022face} featuring moderate to severe real motion blur. The synthetic data pipeline in other supervised approaches does not model motion blur, resulting in poor performance on this out-of-distribution dataset. In contrast, our method consistently restores clean images from complex non-uniform motion blur, as seen in Figure~\ref{fig:deblur}, outperforming previous methods significantly, as shown in Table~\ref{tab:standard}.

\begin{figure*}[ht]
    \centering
    \setlength{\tabcolsep}{1pt}
    \def\imW{0.2\linewidth}
    \scalebox{0.9}{
    \begin{tabular}{ccccccc}
        \raisebox{-.0\height}{\includegraphics[width=\imW]{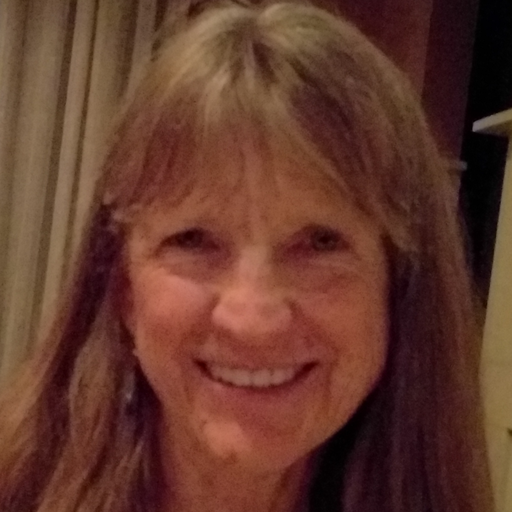}} &
        \raisebox{-.0\height}{\includegraphics[width=\imW]{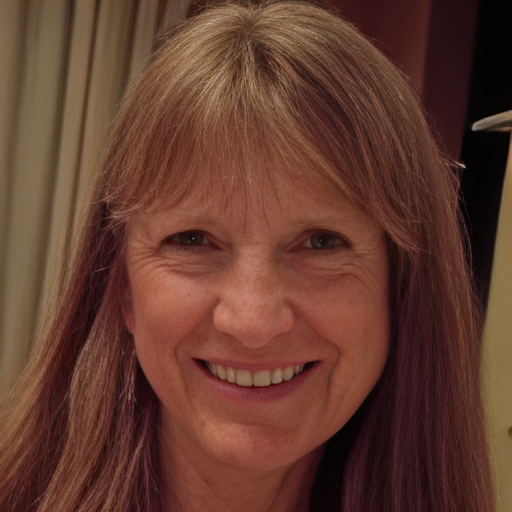}} & 
        \raisebox{-.0\height}{\includegraphics[width=\imW]{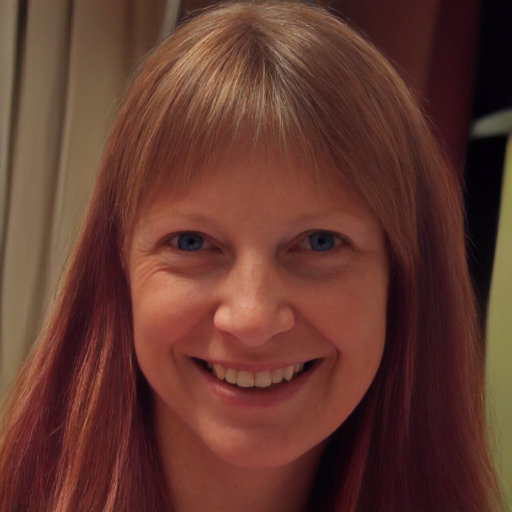}} &
        \raisebox{-.0\height}{\includegraphics[width=\imW]{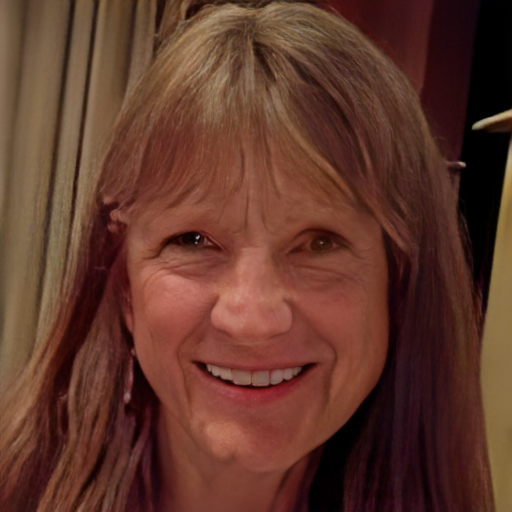}} & 
        \raisebox{-.0\height}{\includegraphics[width=\imW]{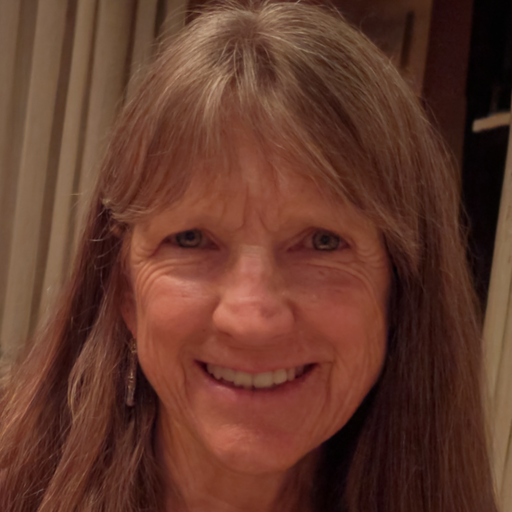}} \\
        \raisebox{-.0\height}{\includegraphics[width=\imW]{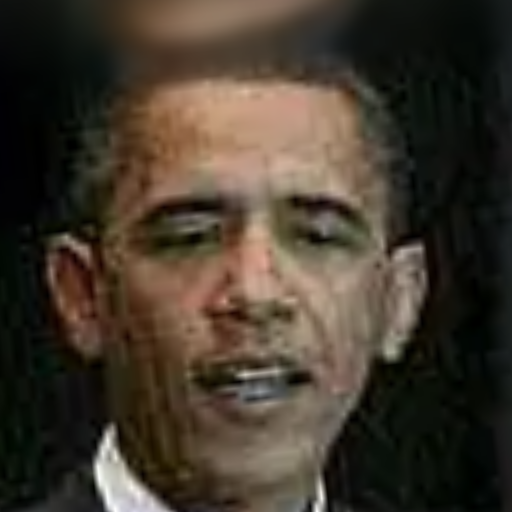}} &
        \raisebox{-.0\height}{\includegraphics[width=\imW]{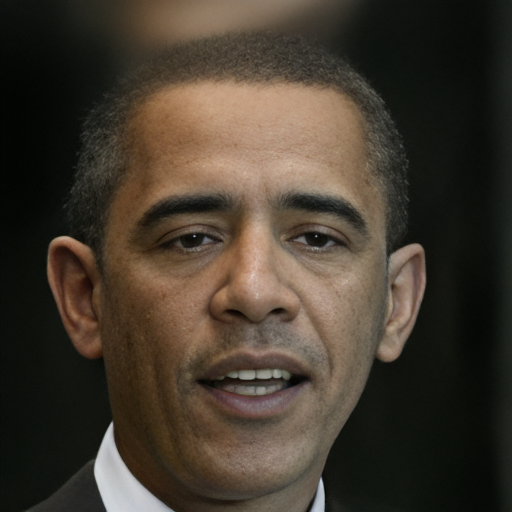}} &
        \raisebox{-.0\height}{\includegraphics[width=\imW]{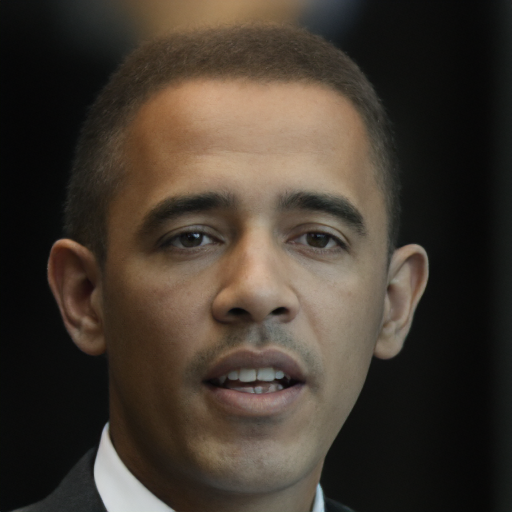}} &
        \raisebox{-.0\height}{\includegraphics[width=\imW]{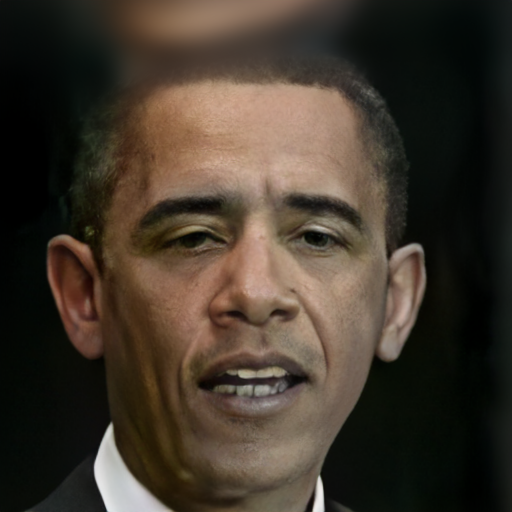}} &
        \raisebox{-.0\height}{\includegraphics[width=\imW]{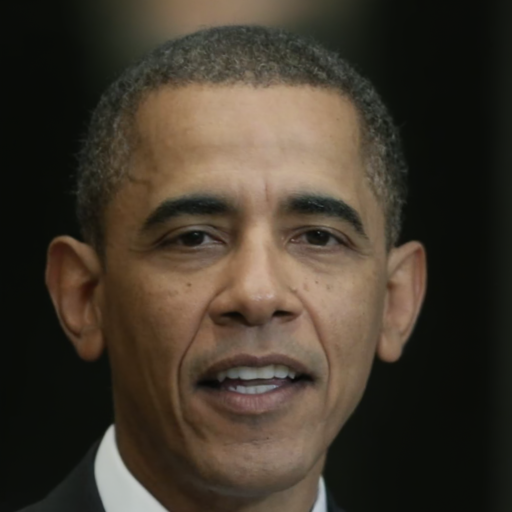}} \\
        \raisebox{-.0\height}{\includegraphics[width=\imW]{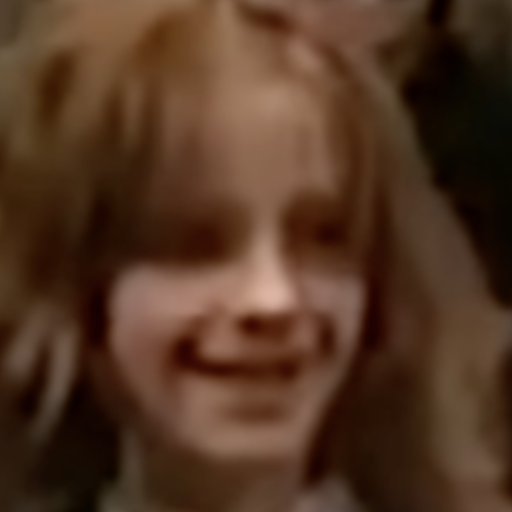}} &
        \raisebox{-.0\height}{\includegraphics[width=\imW]{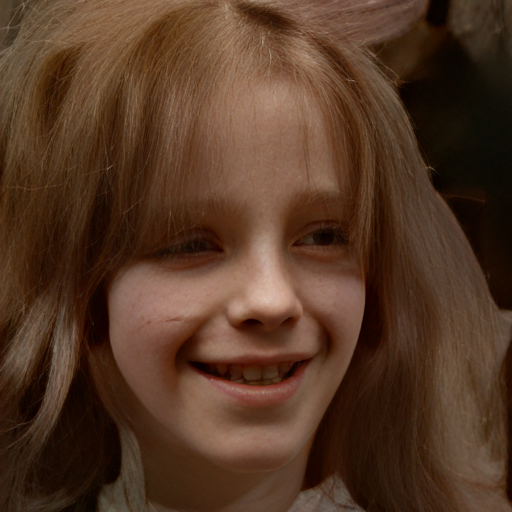}} &
        \raisebox{-.0\height}{\includegraphics[width=\imW]{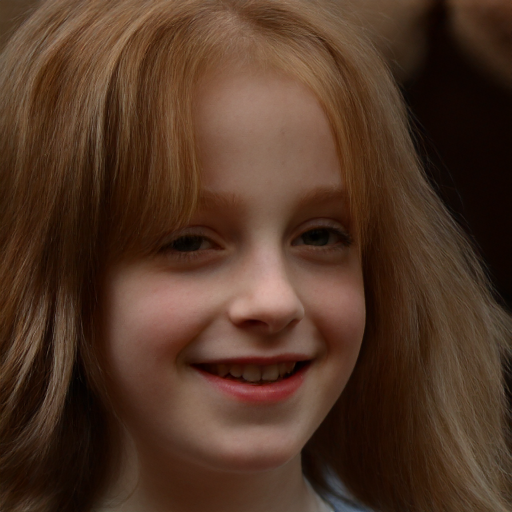}} &
        \raisebox{-.0\height}{\includegraphics[width=\imW]{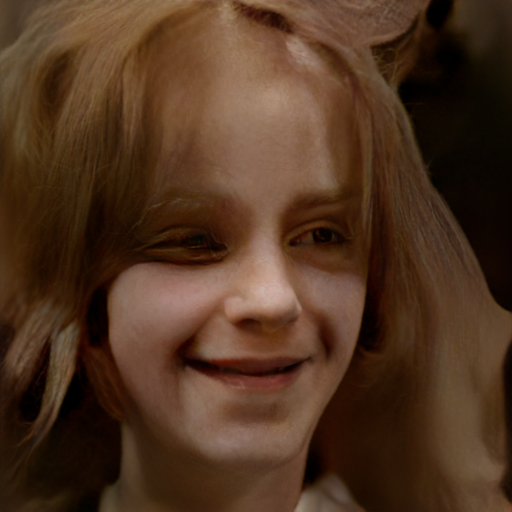}} &
        \raisebox{-.0\height}{\includegraphics[width=\imW]{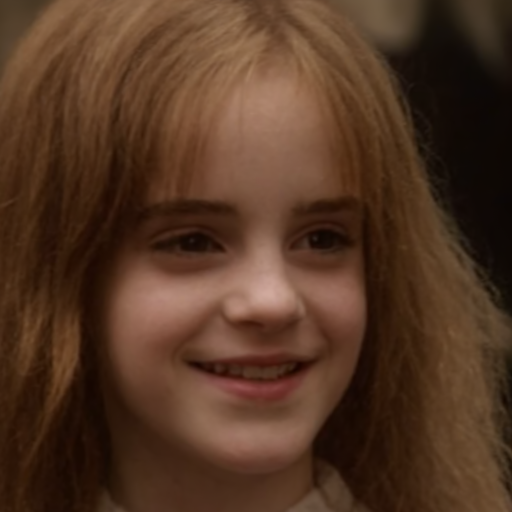}} \\
        Input  & CodeFormer & DR2(+VQFR) & ASFFNet & Ours \\
    \end{tabular}
    }
    \vspace{-7pt}
    \caption{\textbf{Qualitative Comparison on personalized face restoration.} From top to bottom: Subject A, Obama and Hermione. With a personal album as anchor, we are able to restore images with faithful preservation of the input identity. Previous single-image methods alter the identity with lost details; previous reference-based methods fail to produce high-quality images and are prone to artifacts.}
    \vspace{-4mm}
    \label{fig:personal}
\end{figure*}

\subsection{Personalized Face Restoration}
\label{subsec:personal}
We now evaluate our method on personalized restoration. Given a set of clean images of a subject, the goal is to restore any degraded image of the same subject using personalized features to preserve identity and recover high-frequency details that may have been lost in the degraded image. Our method naturally incorporates the personal album as the anchor. We use a personal album that contains around 20 images with diversity in pose, hairstyle, accessories, lighting, etc. We fine-tune on the personal album for 5,000 iterations. The model can then be used to restore any low-quality images of the same subject through direct sampling.

We compare our method against three single-image-based works: Codeformer \cite{zhou2022towards}, VQFR \cite{gu2022vqfr}, DR2 \cite{wang2023dr2}, as well as an exemplar-based approach ASFFNet \cite{li2020enhanced} which also incorporates a personal album for additional information. We evaluate our approach on three subjects: an elderly woman (Subject A), Obama and Hermione. We present the qualitative comparison in Figure~\ref{fig:personal}. Single-image-based methods struggle to preserve identity -- for example, wrinkles and other facial structures are often missing in the results of CodeFormer or DR2 for the elderly subject, altering their age and identity. By using a photo album as reference, ASFFNet preserves identity better, but fails to produce high-quality results. Our method, on the other hand, directly samples from the personalized generative space to do restoration, and thus produces faithful and high-quality results. 

We also provide quantitative evaluation in Table \ref{tab:personal} where we focus on the identity preservation. We use the identity score which uses the cosine similarity of the features given by a face recognition network ArcFace\cite{arcface}. For each subjects, we collect around 20 test images and compute their average identity scores. Table~\ref{tab:personal} shows that our method preserves the identity of the subject much better than both single-image-based methods and the exemplar-based approach ASFFNet. 

\begin{table}[h]
\centering
\scalebox{0.7}{
\begin{tabular}{l|ccc}
\toprule
& Subject A & Obama & Hermione \\
\hline
Input & 0.721 & 0.502 & 0.483 \\
CodeFormer\cite{zhou2022towards} & 0.633 & 0.558 & 0.518 \\
VQFR\cite{gu2022vqfr} & 0.560 & 0.527 & 0.483 \\
DR2(+VQFR)\cite{wang2023dr2} & 0.384 & 0.400 & 0.392  \\
ASFFNet\cite{li2020enhanced} & 0.694 & 0.574 & 0.522 \\
Ours & \textbf{0.731} & \textbf{0.716} & \textbf{0.664}  \\
\bottomrule
\end{tabular}
}
\caption{\textbf{IDS comparison on three subjects.} We use the cosine similarity of the features given by ArcFace\cite{arcface} to compute identity score.}
\label{tab:personal}
\end{table}

\subsection{Beyond Face Restoration}
\label{subsec:pets}

\begin{figure}[h]
    \centering
    \setlength{\tabcolsep}{1pt}
    \def\imW{0.24\linewidth}
    \begin{tabular}{cc@{\hskip 8pt}cc}
        \raisebox{-.0\height}{\includegraphics[width=\imW]{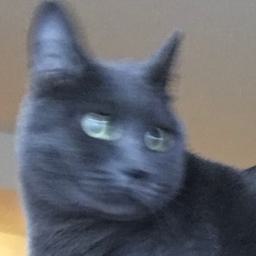}} &
        \raisebox{-.0\height}{\includegraphics[width=\imW]{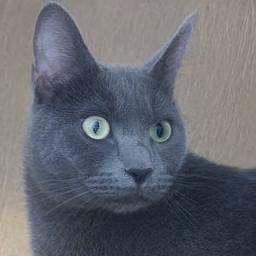}} &
        \raisebox{-.0\height}{\includegraphics[width=\imW]{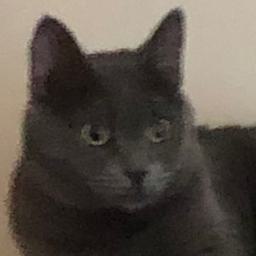}} &
        \raisebox{-.0\height}{\includegraphics[width=\imW]{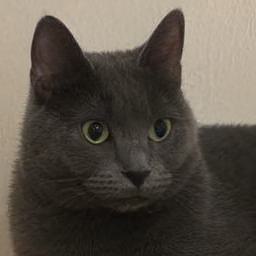}}\\
        \raisebox{-.0\height}{\includegraphics[width=\imW]{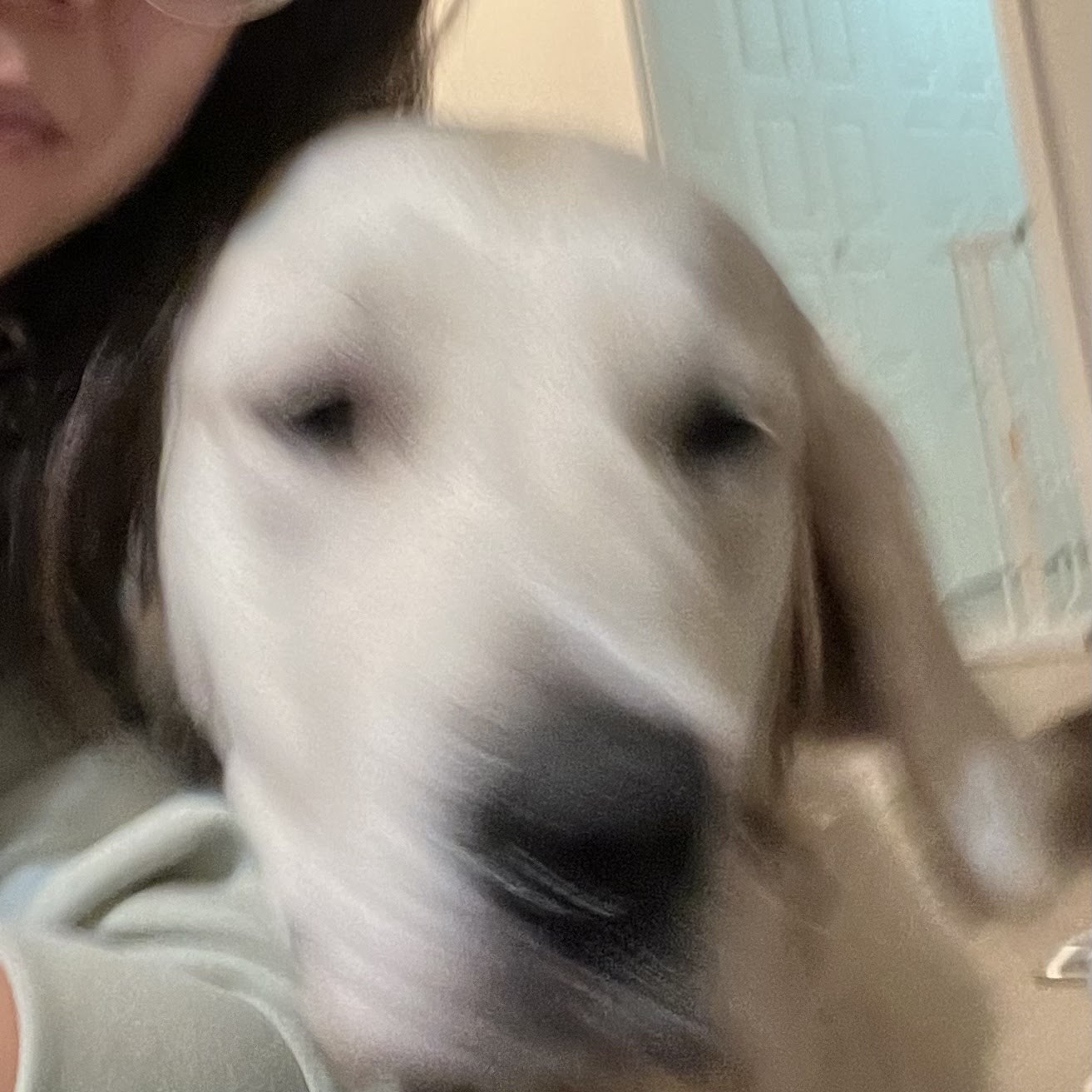}} &
        \raisebox{-.0\height}{\includegraphics[width=\imW]{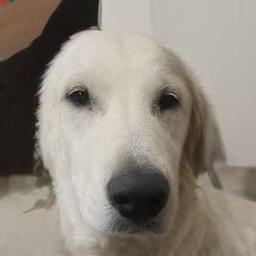}} &
        \raisebox{-.0\height}{\includegraphics[width=\imW]{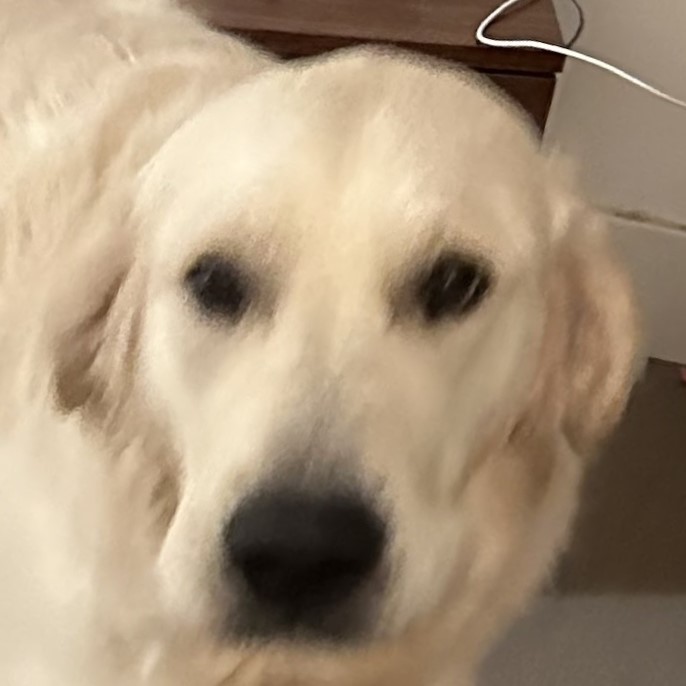}} &
        \raisebox{-.0\height}{\includegraphics[width=\imW]{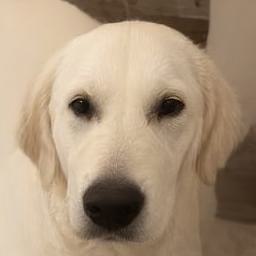}} \\
        \raisebox{-.0\height}{\includegraphics[width=\imW]{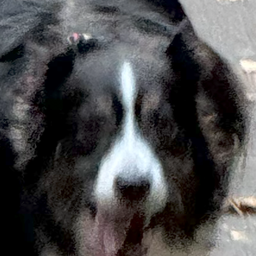}} &
        \raisebox{-.0\height}{\includegraphics[width=\imW]{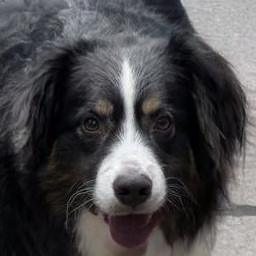}} &
        \raisebox{-.0\height}{\includegraphics[width=\imW]{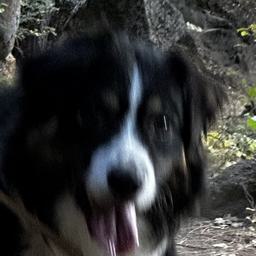}} &
        \raisebox{-.0\height}{\includegraphics[width=\imW]{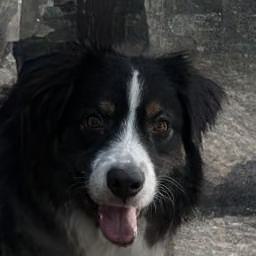}} \\
        Input & Ours & Input & Ours \\
    \end{tabular}
    \vspace{-7pt}
    \caption{\textbf{Results on real-world cat/dog restoration.} Our method easily extends to other categories with corresponding pre-trained diffusion models. We show results on cats and dogs where we can reconstruct high-frequency details while preserving the identity.}
    
    \label{fig:pet}
    \vspace{-4mm}
\end{figure}

Our model does not make any assumptions about the type of degradation or image contents, allowing it to be easily extended to other categories of data where a generative model is available. Specifically, we evaluate our approach's ability to generalize to restoring dog and cat images. We pre-train two diffusion models with the same architecture, one for dogs and one for cats, on the AFHQ Dog and Cat datasets \cite{choi2020stargan}. Our testing involves three subjects: a gray cat, an English golden retriever, and an Australian shepherd. For each subject, we fine-tune the pre-trained diffusion model using an album of around 20 images. Once fine-tuned, given a low-quality image, we add noise to it and then denoise it using the fine-tuned model. Qualitative results in Figure~\ref{fig:pet} demonstrate that our method can effectively reconstruct high-frequency details such as fur, while preserving the identity.

\vspace{-2mm}
\section{Ablation Studies}

\begin{figure}[h]
    \centering
    \setlength{\tabcolsep}{1pt}
    \def\imW{0.23\linewidth}
    \scalebox{0.95}{
    \begin{tabular}{ccccc}
        \multirow[c]{2}{*}[-2mm]{{\includegraphics[width=\imW]{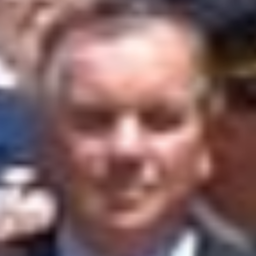}}} &
        \raisebox{-.5\height}{\includegraphics[width=\imW]{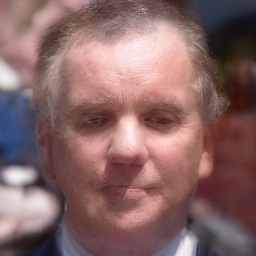}} &
        \raisebox{-.5\height}{\includegraphics[width=\imW]{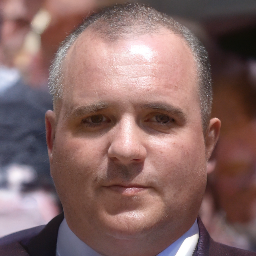}} &
        \raisebox{-.5\height}{\includegraphics[width=\imW]{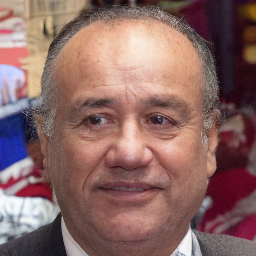}} & \rotatebox[origin=c]{90}{\footnotesize w/o Constraining} \vspace{3pt} \\
        &
        \raisebox{-.5\height}{\includegraphics[width=\imW]{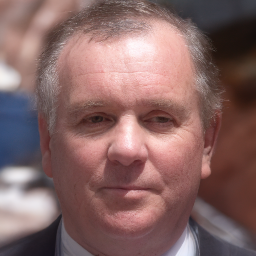}} &
        \raisebox{-.5\height}{\includegraphics[width=\imW]{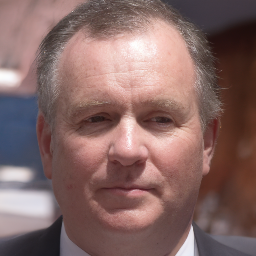}} &
        \raisebox{-.5\height}{\includegraphics[width=\imW]{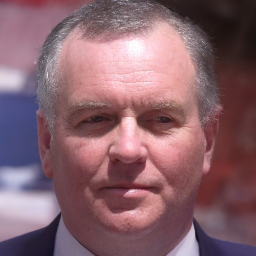}} & \rotatebox[origin=c]{90}{\footnotesize w/ Constraining} \vspace{3pt} \\
        Input & $K=200$  & $K=400$ & $K=600$ \\
    \end{tabular}
    }
    \vspace{-7pt}
    \caption{\textbf{Ablation on Noise Step \( K \) and Constraining with Generative Album.} As \( K \) increases, quality of images sampled from \( y_K \) improves, but alignment with the input reduces. Fine-tuning with a generative album notably enhances both image quality and input fidelity.}
    \label{fig:ab_generative}
    \vspace{-1mm}
\end{figure}

\paragraph{Noise Step $K$.} Our restoration-by-generation approach is predicated on the observation that sufficient noise added to a degraded image $y_0$ and subsequent denoising of the noisy image $y_K$ with a pre-trained diffusion model yields a high-quality, realistic image. Here, we demonstrate this observation and analyze the effect of the choice of $K$, which determines the noise level added to initiate the sampling process. Figure~\ref{fig:ab_generative} displays sampled images from $y_K$ for varying $K$ values. A smaller $K$ leads to a $y_K$ that falls outside the typical diffusion process's training trajectory, resulting in lower-quality sampled output. Conversely, while a larger $K$ enhances sample quality as hypothesized, it may also produce outputs less faithful to the input.
\vspace{-4mm}

\paragraph{Constraining Prior with Generative Album.} In the same Figure~\ref{fig:ab_generative}, we illustrate the significance of prior constraining and the effectiveness of using a generative album. As shown, a generative space that is too diverse increases the difficulty of sampling high-quality images from a given input, especially when $K$ is small. Conversely, for large $K$ values, the sampled image can deviate significantly from the input. Constraining the generative space with an album close to the input ensures preservation of input information in the output for large $K$, while still allowing high-quality sampling from small $K$. Ablation on Skip Guidance is included in the supplementary.
\vspace{-4mm}

\begin{figure}[h]
    \centering
    \setlength{\tabcolsep}{1pt}
    \def\imW{0.24\linewidth}
    \begin{tabular}{cccc}
        \raisebox{-.0\height}{\includegraphics[width=\imW]{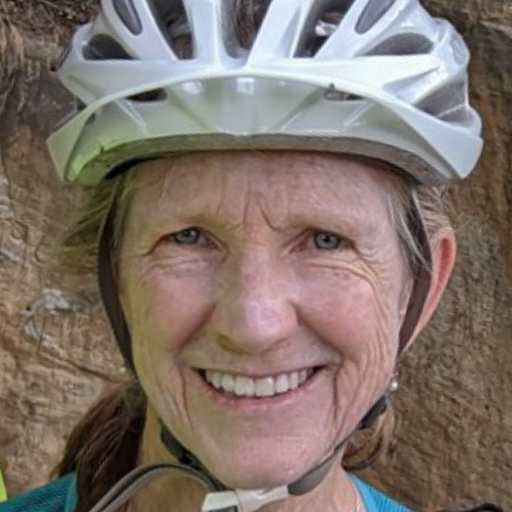}} &
        \raisebox{-.0\height}{\includegraphics[width=\imW]{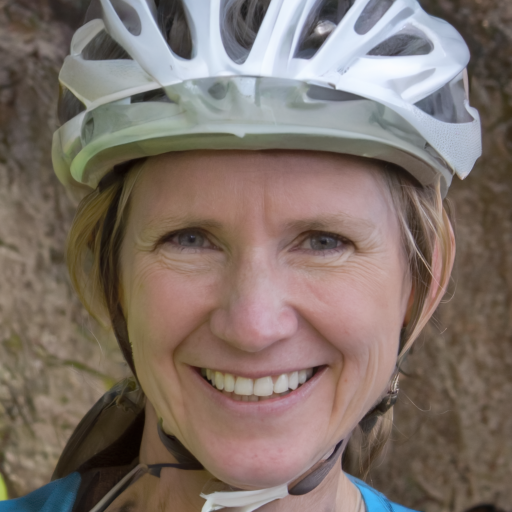}} &
        \raisebox{-.0\height}{\includegraphics[width=\imW]{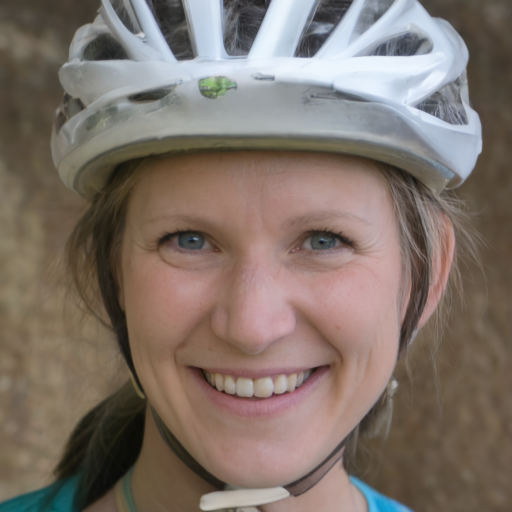}}&
        \raisebox{-.0\height}{\includegraphics[width=\imW]{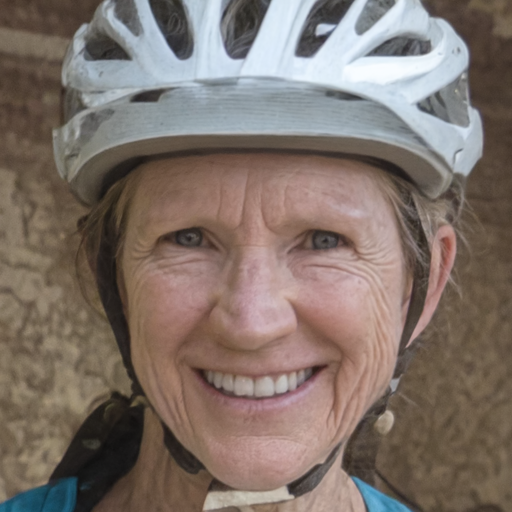}} \\
        \raisebox{-.0\height}{\includegraphics[width=\imW]{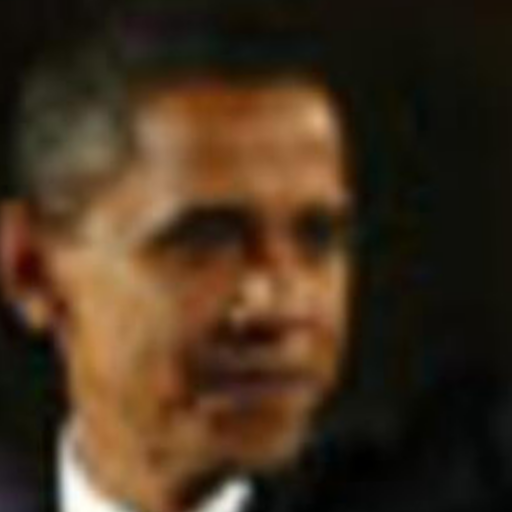}} &
        \raisebox{-.0\height}{\includegraphics[width=\imW]{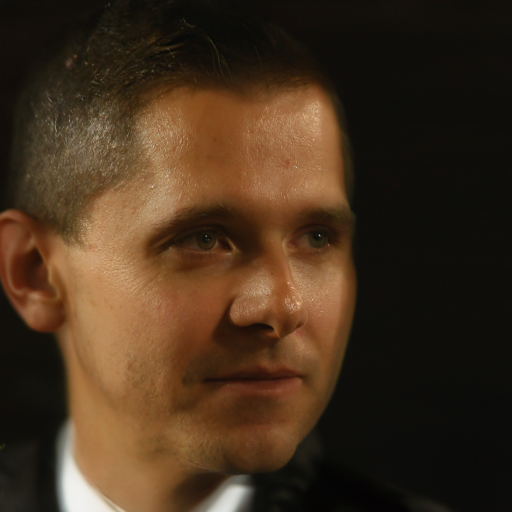}} &
        \raisebox{-.0\height}{\includegraphics[width=\imW]{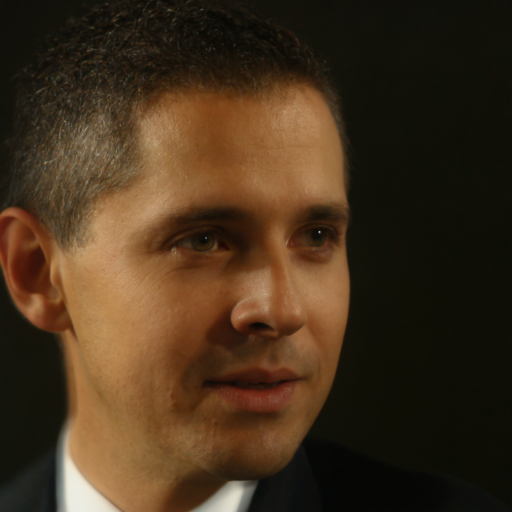}} &
        \raisebox{-.0\height}{\includegraphics[width=\imW]{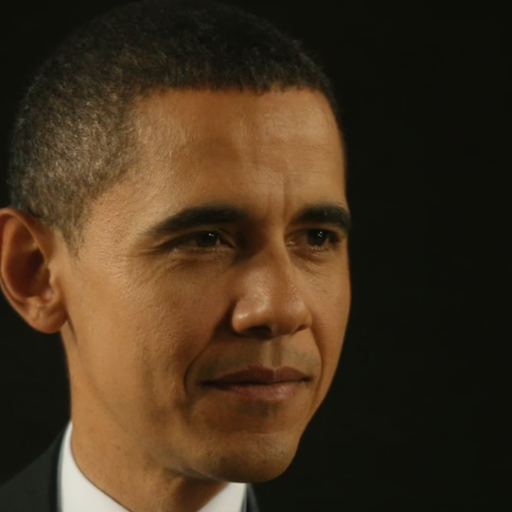}} \\
        {\footnotesize Input} & {\footnotesize w/o constraining} & {\footnotesize w/ generative} & {\footnotesize w/ album} \vspace{-1mm}\\
        & & {\footnotesize constraining} & {\footnotesize constraining}
    \end{tabular}
    \vspace{-5pt}
    \caption{\textbf{Constraining with personal album.} Personalized model produces higher-quality images that better preserve identity compared to the model without constraining.}
    \label{fig:ab_person}
    \vspace{-3mm}
\end{figure}

\paragraph{Constraining Prior with Personal Album.} When a personal album is available, we directly constrain the generative space with this album. This not only improves output quality and faithfulness, as with the generative album, but also aids in recovering information absent in the input. As demonstrated in Figure~\ref{fig:ab_person}, compared to an unconstrained model (i.e., the pre-trained diffusion model), the personalized model produces higher-quality images that better preserve identity.

\section{Conclusion}

We propose a method for image restoration that involves simply adding noise to a degraded input and then denoising it with a diffusion model. The key to our approach is constraining the generative space with a set of anchor images. We demonstrate in single-image restoration tasks that this method yields high-quality restoration results, surpassing previous supervised approaches.
Furthermore, we show that constraining the generative space with a personal album leads to a personalized restoration-by-generation model that is effective for any image of the same subject, producing results with high quality and faithful details.
\vspace{-4mm}

\paragraph{Limitations and Future Work.} Unlike the personalization case, 
for single-image restoration, our approach requires fine-tuning for each input image. This is relatively slow compared to feed-forward approaches. Investigating methods to constrain the generative space without fine-tuning could be interesting. Furthermore, we have primarily validated our approach on class-specific image restoration tasks, largely due to the absence of a high-quality pre-trained diffusion model for natural images. Exploring whether our approach remains effective within a more diverse generative space would be intriguing.
Such exploration could potentially address the challenge of blind restoration for general images.

\vspace{-2mm}
\paragraph{Acknowledgment} We thank Marc Levoy for providing valuable feedback, and everyone whose photos appear in the paper, including our furry friends, Chuchu, Nobi and Panghu.

{
    \small
    \bibliographystyle{ieeenat_fullname}
    \bibliography{main}

\begin{thebibliography}{60}
\providecommand{\natexlab}[1]{#1}
\providecommand{\url}[1]{\texttt{#1}}
\expandafter\ifx\csname urlstyle\endcsname\relax
  \providecommand{\doi}[1]{doi: #1}\else
  \providecommand{\doi}{doi: \begingroup \urlstyle{rm}\Url}\fi

\bibitem[Bansal et~al.(2023)Bansal, Chu, Schwarzschild, Sengupta, Goldblum, Geiping, and Goldstein]{Bansal_2023_CVPR}
Arpit Bansal, Hong-Min Chu, Avi Schwarzschild, Soumyadip Sengupta, Micah Goldblum, Jonas Geiping, and Tom Goldstein.
\newblock Universal guidance for diffusion models.
\newblock In \emph{Proceedings of the IEEE/CVF Conference on Computer Vision and Pattern Recognition (CVPR) Workshops}, pages 843--852, 2023.

\bibitem[Cho et~al.(2021)Cho, Ji, Hong, Jung, and Ko]{cho2021rethinking}
Sung-Jin Cho, Seo-Won Ji, Jun-Pyo Hong, Seung-Won Jung, and Sung-Jea Ko.
\newblock Rethinking coarse-to-fine approach in single image deblurring.
\newblock In \emph{Proceedings of the IEEE/CVF international conference on computer vision}, pages 4641--4650, 2021.

\bibitem[Choi et~al.(2021)Choi, Kim, Jeong, Gwon, and Yoon]{choi2021ilvr}
Jooyoung Choi, Sungwon Kim, Yonghyun Jeong, Youngjune Gwon, and Sungroh Yoon.
\newblock Ilvr: Conditioning method for denoising diffusion probabilistic models.
\newblock \emph{arXiv preprint arXiv:2108.02938}, 2021.

\bibitem[Choi et~al.(2020)Choi, Uh, Yoo, and Ha]{choi2020stargan}
Yunjey Choi, Youngjung Uh, Jaejun Yoo, and Jung-Woo Ha.
\newblock Stargan v2: Diverse image synthesis for multiple domains.
\newblock In \emph{Proceedings of the IEEE/CVF conference on computer vision and pattern recognition}, pages 8188--8197, 2020.

\bibitem[Chung et~al.(2022)Chung, Kim, Mccann, Klasky, and Ye]{chung2022diffusion}
Hyungjin Chung, Jeongsol Kim, Michael~T Mccann, Marc~L Klasky, and Jong~Chul Ye.
\newblock Diffusion posterior sampling for general noisy inverse problems.
\newblock \emph{arXiv preprint arXiv:2209.14687}, 2022.

\bibitem[Deng et~al.(2019)Deng, Guo, Xue, and Zafeiriou]{arcface}
Jiankang Deng, Jia Guo, Niannan Xue, and Stefanos Zafeiriou.
\newblock Arcface: Additive angular margin loss for deep face recognition.
\newblock In \emph{Proceedings of the IEEE/CVF conference on computer vision and pattern recognition}, pages 4690--4699, 2019.

\bibitem[Dhariwal and Nichol(2021{\natexlab{a}})]{adm}
Prafulla Dhariwal and Alexander Nichol.
\newblock Diffusion models beat gans on image synthesis.
\newblock \emph{Advances in neural information processing systems}, 34:\penalty0 8780--8794, 2021{\natexlab{a}}.

\bibitem[Dhariwal and Nichol(2021{\natexlab{b}})]{dhariwal2021diffusion}
Prafulla Dhariwal and Alexander Nichol.
\newblock Diffusion models beat gans on image synthesis.
\newblock \emph{Advances in neural information processing systems}, 34:\penalty0 8780--8794, 2021{\natexlab{b}}.

\bibitem[Ding et~al.(2023)Ding, Zhang, Xia, Jebe, Tu, and Zhang]{diffusionrig}
Zheng Ding, Xuaner Zhang, Zhihao Xia, Lars Jebe, Zhuowen Tu, and Xiuming Zhang.
\newblock Diffusionrig: Learning personalized priors for facial appearance editing.
\newblock In \emph{Proceedings of the IEEE/CVF Conference on Computer Vision and Pattern Recognition}, pages 12736--12746, 2023.

\bibitem[Gal et~al.(2022)Gal, Alaluf, Atzmon, Patashnik, Bermano, Chechik, and Cohen-Or]{gal2022image}
Rinon Gal, Yuval Alaluf, Yuval Atzmon, Or Patashnik, Amit~H Bermano, Gal Chechik, and Daniel Cohen-Or.
\newblock An image is worth one word: Personalizing text-to-image generation using textual inversion.
\newblock \emph{arXiv preprint arXiv:2208.01618}, 2022.

\bibitem[Gal et~al.(2023)Gal, Arar, Atzmon, Bermano, Chechik, and Cohen-Or]{gal2023encoder}
Rinon Gal, Moab Arar, Yuval Atzmon, Amit~H Bermano, Gal Chechik, and Daniel Cohen-Or.
\newblock Encoder-based domain tuning for fast personalization of text-to-image models.
\newblock \emph{ACM Transactions on Graphics (TOG)}, 42\penalty0 (4):\penalty0 1--13, 2023.

\bibitem[Gu et~al.(2022)Gu, Wang, Xie, Dong, Li, Shan, and Cheng]{gu2022vqfr}
Yuchao Gu, Xintao Wang, Liangbin Xie, Chao Dong, Gen Li, Ying Shan, and Ming-Ming Cheng.
\newblock Vqfr: Blind face restoration with vector-quantized dictionary and parallel decoder.
\newblock In \emph{European Conference on Computer Vision}, pages 126--143. Springer, 2022.

\bibitem[Guo et~al.(2019)Guo, Yan, Zhang, Zuo, and Zhang]{guo2019toward}
Shi Guo, Zifei Yan, Kai Zhang, Wangmeng Zuo, and Lei Zhang.
\newblock Toward convolutional blind denoising of real photographs.
\newblock In \emph{Proceedings of the IEEE/CVF conference on computer vision and pattern recognition}, pages 1712--1722, 2019.

\bibitem[Heusel et~al.(2017)Heusel, Ramsauer, Unterthiner, Nessler, and Hochreiter]{fid}
Martin Heusel, Hubert Ramsauer, Thomas Unterthiner, Bernhard Nessler, and Sepp Hochreiter.
\newblock Gans trained by a two time-scale update rule converge to a local nash equilibrium.
\newblock \emph{Advances in neural information processing systems}, 30, 2017.

\bibitem[Ho et~al.(2020)Ho, Jain, and Abbeel]{ddpm}
Jonathan Ho, Ajay Jain, and Pieter Abbeel.
\newblock Denoising diffusion probabilistic models.
\newblock \emph{Advances in neural information processing systems}, 33:\penalty0 6840--6851, 2020.

\bibitem[Jia et~al.(2023)Jia, Zhao, Chan, Li, Zhang, Gong, Hou, Wang, and Su]{jia2023taming}
Xuhui Jia, Yang Zhao, Kelvin~CK Chan, Yandong Li, Han Zhang, Boqing Gong, Tingbo Hou, Huisheng Wang, and Yu-Chuan Su.
\newblock Taming encoder for zero fine-tuning image customization with text-to-image diffusion models.
\newblock \emph{arXiv preprint arXiv:2304.02642}, 2023.

\bibitem[Kadkhodaie and Simoncelli(2021)]{kadkhodaie2021stochastic}
Zahra Kadkhodaie and Eero Simoncelli.
\newblock Stochastic solutions for linear inverse problems using the prior implicit in a denoiser.
\newblock \emph{Advances in Neural Information Processing Systems}, 34:\penalty0 13242--13254, 2021.

\bibitem[Karras et~al.(2019)Karras, Laine, and Aila]{ffhq}
Tero Karras, Samuli Laine, and Timo Aila.
\newblock A style-based generator architecture for generative adversarial networks.
\newblock In \emph{Proceedings of the IEEE/CVF conference on computer vision and pattern recognition}, pages 4401--4410, 2019.

\bibitem[Kawar et~al.(2021{\natexlab{a}})Kawar, Vaksman, and Elad]{kawar2021snips}
Bahjat Kawar, Gregory Vaksman, and Michael Elad.
\newblock Snips: Solving noisy inverse problems stochastically.
\newblock \emph{Advances in Neural Information Processing Systems}, 34:\penalty0 21757--21769, 2021{\natexlab{a}}.

\bibitem[Kawar et~al.(2021{\natexlab{b}})Kawar, Vaksman, and Elad]{kawar2021stochastic}
Bahjat Kawar, Gregory Vaksman, and Michael Elad.
\newblock Stochastic image denoising by sampling from the posterior distribution.
\newblock In \emph{Proceedings of the IEEE/CVF International Conference on Computer Vision}, pages 1866--1875, 2021{\natexlab{b}}.

\bibitem[Kawar et~al.(2022)Kawar, Elad, Ermon, and Song]{kawar2022denoising}
Bahjat Kawar, Michael Elad, Stefano Ermon, and Jiaming Song.
\newblock Denoising diffusion restoration models.
\newblock \emph{Advances in Neural Information Processing Systems}, 35:\penalty0 23593--23606, 2022.

\bibitem[Ke et~al.(2021)Ke, Wang, Wang, Milanfar, and Yang]{musiq}
Junjie Ke, Qifei Wang, Yilin Wang, Peyman Milanfar, and Feng Yang.
\newblock Musiq: Multi-scale image quality transformer.
\newblock In \emph{Proceedings of the IEEE/CVF International Conference on Computer Vision}, pages 5148--5157, 2021.

\bibitem[Kumari et~al.(2023)Kumari, Zhang, Zhang, Shechtman, and Zhu]{kumari2023multi}
Nupur Kumari, Bingliang Zhang, Richard Zhang, Eli Shechtman, and Jun-Yan Zhu.
\newblock Multi-concept customization of text-to-image diffusion.
\newblock In \emph{Proceedings of the IEEE/CVF Conference on Computer Vision and Pattern Recognition}, pages 1931--1941, 2023.

\bibitem[Kupyn et~al.(2018)Kupyn, Budzan, Mykhailych, Mishkin, and Matas]{kupyn2018deblurgan}
Orest Kupyn, Volodymyr Budzan, Mykola Mykhailych, Dmytro Mishkin, and Ji{\v{r}}{\'\i} Matas.
\newblock Deblurgan: Blind motion deblurring using conditional adversarial networks.
\newblock In \emph{Proceedings of the IEEE conference on computer vision and pattern recognition}, pages 8183--8192, 2018.

\bibitem[Kupyn et~al.(2019)Kupyn, Martyniuk, Wu, and Wang]{kupyn2019deblurgan}
Orest Kupyn, Tetiana Martyniuk, Junru Wu, and Zhangyang Wang.
\newblock Deblurgan-v2: Deblurring (orders-of-magnitude) faster and better.
\newblock In \emph{Proceedings of the IEEE/CVF international conference on computer vision}, pages 8878--8887, 2019.

\bibitem[Lai et~al.(2022)Lai, Shih, Chu, Wu, Tsai, Krainin, Sun, and Liang]{lai2022face}
Wei-Sheng Lai, Yichang Shih, Lun-Cheng Chu, Xiaotong Wu, Sung-Fang Tsai, Michael Krainin, Deqing Sun, and Chia-Kai Liang.
\newblock Face deblurring using dual camera fusion on mobile phones.
\newblock \emph{ACM Transactions on Graphics (TOG)}, 41\penalty0 (4):\penalty0 1--16, 2022.

\bibitem[Ledig et~al.(2017)Ledig, Theis, Husz{\'a}r, Caballero, Cunningham, Acosta, Aitken, Tejani, Totz, Wang, et~al.]{srgan}
Christian Ledig, Lucas Theis, Ferenc Husz{\'a}r, Jose Caballero, Andrew Cunningham, Alejandro Acosta, Andrew Aitken, Alykhan Tejani, Johannes Totz, Zehan Wang, et~al.
\newblock Photo-realistic single image super-resolution using a generative adversarial network.
\newblock In \emph{Proceedings of the IEEE conference on computer vision and pattern recognition}, pages 4681--4690, 2017.

\bibitem[Li et~al.(2018)Li, Liu, Ye, Zuo, Lin, and Yang]{li2018learning}
Xiaoming Li, Ming Liu, Yuting Ye, Wangmeng Zuo, Liang Lin, and Ruigang Yang.
\newblock Learning warped guidance for blind face restoration.
\newblock In \emph{Proceedings of the European conference on computer vision (ECCV)}, pages 272--289, 2018.

\bibitem[Li et~al.(2020)Li, Li, Ren, Zhang, Wang, and Zuo]{li2020enhanced}
Xiaoming Li, Wenyu Li, Dongwei Ren, Hongzhi Zhang, Meng Wang, and Wangmeng Zuo.
\newblock Enhanced blind face restoration with multi-exemplar images and adaptive spatial feature fusion.
\newblock In \emph{Proceedings of the IEEE/CVF Conference on Computer Vision and Pattern Recognition}, pages 2706--2715, 2020.

\bibitem[Liang et~al.(2021)Liang, Cao, Sun, Zhang, Van~Gool, and Timofte]{swinir}
Jingyun Liang, Jiezhang Cao, Guolei Sun, Kai Zhang, Luc Van~Gool, and Radu Timofte.
\newblock Swinir: Image restoration using swin transformer.
\newblock In \emph{Proceedings of the IEEE/CVF international conference on computer vision}, pages 1833--1844, 2021.

\bibitem[Nichol and Dhariwal(2021)]{nichol2021improved}
Alexander~Quinn Nichol and Prafulla Dhariwal.
\newblock Improved denoising diffusion probabilistic models.
\newblock In \emph{International Conference on Machine Learning}, pages 8162--8171. PMLR, 2021.

\bibitem[Pan et~al.(2021)Pan, Zhan, Dai, Lin, Loy, and Luo]{pan2021exploiting}
Xingang Pan, Xiaohang Zhan, Bo Dai, Dahua Lin, Chen~Change Loy, and Ping Luo.
\newblock Exploiting deep generative prior for versatile image restoration and manipulation.
\newblock \emph{IEEE Transactions on Pattern Analysis and Machine Intelligence}, 44\penalty0 (11):\penalty0 7474--7489, 2021.

\bibitem[Romano et~al.(2017)Romano, Elad, and Milanfar]{red}
Yaniv Romano, Michael Elad, and Peyman Milanfar.
\newblock The little engine that could: Regularization by denoising (red).
\newblock \emph{SIAM Journal on Imaging Sciences}, 10\penalty0 (4):\penalty0 1804--1844, 2017.

\bibitem[Rudin et~al.(1992)Rudin, Osher, and Fatemi]{tv}
Leonid~I Rudin, Stanley Osher, and Emad Fatemi.
\newblock Nonlinear total variation based noise removal algorithms.
\newblock \emph{Physica D: nonlinear phenomena}, 60\penalty0 (1-4):\penalty0 259--268, 1992.

\bibitem[Ruiz et~al.(2023)Ruiz, Li, Jampani, Pritch, Rubinstein, and Aberman]{dreambooth}
Nataniel Ruiz, Yuanzhen Li, Varun Jampani, Yael Pritch, Michael Rubinstein, and Kfir Aberman.
\newblock Dreambooth: Fine tuning text-to-image diffusion models for subject-driven generation.
\newblock In \emph{Proceedings of the IEEE/CVF Conference on Computer Vision and Pattern Recognition}, pages 22500--22510, 2023.

\bibitem[Saharia et~al.(2022{\natexlab{a}})Saharia, Chan, Chang, Lee, Ho, Salimans, Fleet, and Norouzi]{saharia2022palette}
Chitwan Saharia, William Chan, Huiwen Chang, Chris Lee, Jonathan Ho, Tim Salimans, David Fleet, and Mohammad Norouzi.
\newblock Palette: Image-to-image diffusion models.
\newblock In \emph{ACM SIGGRAPH 2022 Conference Proceedings}, pages 1--10, 2022{\natexlab{a}}.

\bibitem[Saharia et~al.(2022{\natexlab{b}})Saharia, Ho, Chan, Salimans, Fleet, and Norouzi]{saharia2022image}
Chitwan Saharia, Jonathan Ho, William Chan, Tim Salimans, David~J Fleet, and Mohammad Norouzi.
\newblock Image super-resolution via iterative refinement.
\newblock \emph{IEEE Transactions on Pattern Analysis and Machine Intelligence}, 45\penalty0 (4):\penalty0 4713--4726, 2022{\natexlab{b}}.

\bibitem[Shi et~al.(2023)Shi, Xiong, Lin, and Jung]{shi2023instantbooth}
Jing Shi, Wei Xiong, Zhe Lin, and Hyun~Joon Jung.
\newblock Instantbooth: Personalized text-to-image generation without test-time finetuning.
\newblock \emph{arXiv preprint arXiv:2304.03411}, 2023.

\bibitem[Song et~al.(2020)Song, Meng, and Ermon]{ddim}
Jiaming Song, Chenlin Meng, and Stefano Ermon.
\newblock Denoising diffusion implicit models.
\newblock \emph{arXiv preprint arXiv:2010.02502}, 2020.

\bibitem[Song et~al.(2023)Song, Zhang, Yin, Mardani, Liu, Kautz, Chen, and Vahdat]{song2023loss}
Jiaming Song, Qinsheng Zhang, Hongxu Yin, Morteza Mardani, Ming-Yu Liu, Jan Kautz, Yongxin Chen, and Arash Vahdat.
\newblock Loss-guided diffusion models for plug-and-play controllable generation.
\newblock In \emph{Proceedings of the 40th International Conference on Machine Learning}, pages 32483--32498. PMLR, 2023.

\bibitem[Tian et~al.(2020)Tian, Xu, Li, Zuo, Fei, and Liu]{tian2020attention}
Chunwei Tian, Yong Xu, Zuoyong Li, Wangmeng Zuo, Lunke Fei, and Hong Liu.
\newblock Attention-guided cnn for image denoising.
\newblock \emph{Neural Networks}, 124:\penalty0 117--129, 2020.

\bibitem[Wang et~al.(2021{\natexlab{a}})Wang, Li, Zhang, and Shan]{wang2021gfpgan}
Xintao Wang, Yu Li, Honglun Zhang, and Ying Shan.
\newblock Towards real-world blind face restoration with generative facial prior.
\newblock In \emph{The IEEE Conference on Computer Vision and Pattern Recognition (CVPR)}, 2021{\natexlab{a}}.

\bibitem[Wang et~al.(2021{\natexlab{b}})Wang, Li, Zhang, and Shan]{wang2021towards}
Xintao Wang, Yu Li, Honglun Zhang, and Ying Shan.
\newblock Towards real-world blind face restoration with generative facial prior.
\newblock In \emph{Proceedings of the IEEE/CVF conference on computer vision and pattern recognition}, pages 9168--9178, 2021{\natexlab{b}}.

\bibitem[Wang et~al.(2021{\natexlab{c}})Wang, Xie, Dong, and Shan]{wang2021real}
Xintao Wang, Liangbin Xie, Chao Dong, and Ying Shan.
\newblock Real-esrgan: Training real-world blind super-resolution with pure synthetic data.
\newblock In \emph{Proceedings of the IEEE/CVF international conference on computer vision}, pages 1905--1914, 2021{\natexlab{c}}.

\bibitem[Wang et~al.(2022{\natexlab{a}})Wang, Yu, and Zhang]{wang2022zero}
Yinhuai Wang, Jiwen Yu, and Jian Zhang.
\newblock Zero-shot image restoration using denoising diffusion null-space model.
\newblock In \emph{The Eleventh International Conference on Learning Representations}, 2022{\natexlab{a}}.

\bibitem[Wang et~al.(2022{\natexlab{b}})Wang, Cun, Bao, Zhou, Liu, and Li]{wang2022uformer}
Zhendong Wang, Xiaodong Cun, Jianmin Bao, Wengang Zhou, Jianzhuang Liu, and Houqiang Li.
\newblock Uformer: A general u-shaped transformer for image restoration.
\newblock In \emph{Proceedings of the IEEE/CVF conference on computer vision and pattern recognition}, pages 17683--17693, 2022{\natexlab{b}}.

\bibitem[Wang et~al.(2023)Wang, Zhang, Zhang, Zheng, Zhou, Zhang, and Wang]{wang2023dr2}
Zhixin Wang, Ziying Zhang, Xiaoyun Zhang, Huangjie Zheng, Mingyuan Zhou, Ya Zhang, and Yanfeng Wang.
\newblock Dr2: Diffusion-based robust degradation remover for blind face restoration.
\newblock In \emph{Proceedings of the IEEE/CVF Conference on Computer Vision and Pattern Recognition}, pages 1704--1713, 2023.

\bibitem[Whang et~al.(2022)Whang, Delbracio, Talebi, Saharia, Dimakis, and Milanfar]{whang2022deblurring}
Jay Whang, Mauricio Delbracio, Hossein Talebi, Chitwan Saharia, Alexandros~G Dimakis, and Peyman Milanfar.
\newblock Deblurring via stochastic refinement.
\newblock In \emph{Proceedings of the IEEE/CVF Conference on Computer Vision and Pattern Recognition}, pages 16293--16303, 2022.

\bibitem[Xia and Chakrabarti(2020)]{xia2020identifying}
Zhihao Xia and Ayan Chakrabarti.
\newblock Identifying recurring patterns with deep neural networks for natural image denoising.
\newblock In \emph{Proceedings of the IEEE/CVF Winter Conference on Applications of Computer Vision}, pages 2426--2434, 2020.

\bibitem[Xiao et~al.(2023)Xiao, Yin, Freeman, Durand, and Han]{xiao2023fastcomposer}
Guangxuan Xiao, Tianwei Yin, William~T Freeman, Fr{\'e}do Durand, and Song Han.
\newblock Fastcomposer: Tuning-free multi-subject image generation with localized attention.
\newblock \emph{arXiv preprint arXiv:2305.10431}, 2023.

\bibitem[Yang et~al.(2016)Yang, Luo, Loy, and Tang]{yang2016wider}
Shuo Yang, Ping Luo, Chen-Change Loy, and Xiaoou Tang.
\newblock Wider face: A face detection benchmark.
\newblock In \emph{Proceedings of the IEEE conference on computer vision and pattern recognition}, pages 5525--5533, 2016.

\bibitem[Yang et~al.(2021)Yang, Ren, Xie, and Zhang]{yang2021gan}
Tao Yang, Peiran Ren, Xuansong Xie, and Lei Zhang.
\newblock Gan prior embedded network for blind face restoration in the wild.
\newblock In \emph{Proceedings of the IEEE/CVF Conference on Computer Vision and Pattern Recognition}, pages 672--681, 2021.

\bibitem[Zamir et~al.(2022)Zamir, Arora, Khan, Hayat, Khan, and Yang]{restormer}
Syed~Waqas Zamir, Aditya Arora, Salman Khan, Munawar Hayat, Fahad~Shahbaz Khan, and Ming-Hsuan Yang.
\newblock Restormer: Efficient transformer for high-resolution image restoration.
\newblock In \emph{Proceedings of the IEEE/CVF conference on computer vision and pattern recognition}, pages 5728--5739, 2022.

\bibitem[Zhang et~al.(2017{\natexlab{a}})Zhang, Zuo, Chen, Meng, and Zhang]{DnCNN}
Kai Zhang, Wangmeng Zuo, Yunjin Chen, Deyu Meng, and Lei Zhang.
\newblock Beyond a {Gaussian} denoiser: Residual learning of deep {CNN} for image denoising.
\newblock \emph{IEEE Transactions on Image Processing}, 26\penalty0 (7):\penalty0 3142--3155, 2017{\natexlab{a}}.

\bibitem[Zhang et~al.(2017{\natexlab{b}})Zhang, Zuo, Gu, and Zhang]{zhang2017learning}
Kai Zhang, Wangmeng Zuo, Shuhang Gu, and Lei Zhang.
\newblock Learning deep cnn denoiser prior for image restoration.
\newblock In \emph{IEEE Conference on Computer Vision and Pattern Recognition}, pages 3929--3938, 2017{\natexlab{b}}.

\bibitem[Zhang et~al.(2018)Zhang, Zuo, and Zhang]{zhang2018ffdnet}
Kai Zhang, Wangmeng Zuo, and Lei Zhang.
\newblock Ffdnet: Toward a fast and flexible solution for cnn-based image denoising.
\newblock \emph{IEEE Transactions on Image Processing}, 27\penalty0 (9):\penalty0 4608--4622, 2018.

\bibitem[Zhang et~al.(2021)Zhang, Li, Zuo, Zhang, Van~Gool, and Timofte]{zhang2021plug}
Kai Zhang, Yawei Li, Wangmeng Zuo, Lei Zhang, Luc Van~Gool, and Radu Timofte.
\newblock Plug-and-play image restoration with deep denoiser prior.
\newblock \emph{IEEE Transactions on Pattern Analysis and Machine Intelligence}, 44\penalty0 (10):\penalty0 6360--6376, 2021.

\bibitem[Zhang et~al.(2019)Zhang, Li, Li, Zhong, and Fu]{zhang2019residual}
Yulun Zhang, Kunpeng Li, Kai Li, Bineng Zhong, and Yun Fu.
\newblock Residual non-local attention networks for image restoration.
\newblock \emph{arXiv preprint arXiv:1903.10082}, 2019.

\bibitem[Zhao et~al.(2023)Zhao, Hou, Su, Jia, Li, and Grundmann]{zhao2023towards}
Yang Zhao, Tingbo Hou, Yu-Chuan Su, Xuhui Jia, Yandong Li, and Matthias Grundmann.
\newblock Towards authentic face restoration with iterative diffusion models and beyond.
\newblock In \emph{Proceedings of the IEEE/CVF International Conference on Computer Vision}, pages 7312--7322, 2023.

\bibitem[Zhou et~al.(2022)Zhou, Chan, Li, and Loy]{zhou2022towards}
Shangchen Zhou, Kelvin Chan, Chongyi Li, and Chen~Change Loy.
\newblock Towards robust blind face restoration with codebook lookup transformer.
\newblock \emph{Advances in Neural Information Processing Systems}, 35:\penalty0 30599--30611, 2022.

\end{thebibliography}
}

\clearpage

\pagenumbering{arabic}
\renewcommand*{\thepage}{A\arabic{page}}

\setcounter{figure}{0}
\renewcommand\thefigure{S\arabic{figure}}

\setcounter{table}{0}
\renewcommand\thetable{S\arabic{table}}   

\newcommand{\appendixhead}
{\centering{\Large \bf Supplementary Material}
\vspace{20mm}}

\twocolumn[\appendixhead]

\appendix

\section{Additional Results on Blind Face Restoration}
\vspace{1mm}

\subsection{Standard Benchmark}

\begin{figure*}[ht]
    \centering
    \setlength{\tabcolsep}{1pt}
    \def\imW{0.15\linewidth}
    \begin{tabular}{cccccc}
        \raisebox{-.0\height}{\includegraphics[width=\imW]{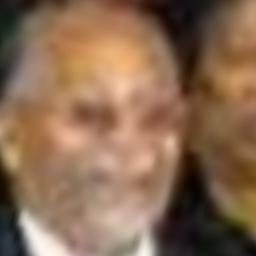}} &
        \raisebox{-.0\height}{\includegraphics[width=\imW]{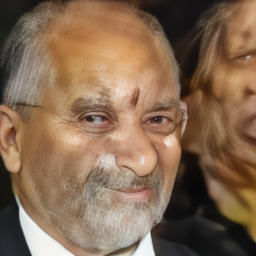}} &
        \raisebox{-.0\height}{\includegraphics[width=\imW]{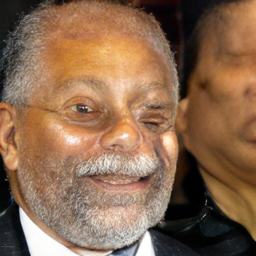}} &
        \raisebox{-.0\height}{\includegraphics[width=\imW]{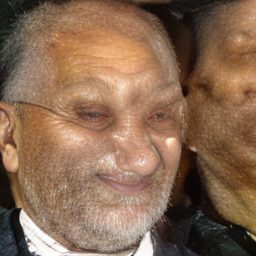}} &
        \raisebox{-.0\height}{\includegraphics[width=\imW]{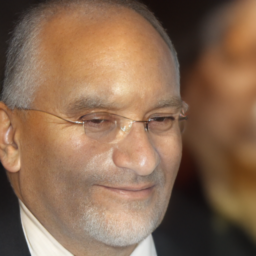}} &
        \raisebox{-.0\height}{\includegraphics[width=\imW]{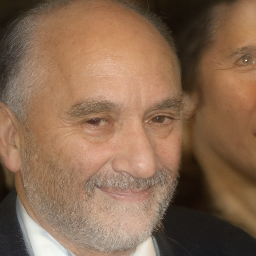}} \\
        \raisebox{-.0\height}{\includegraphics[width=\imW]{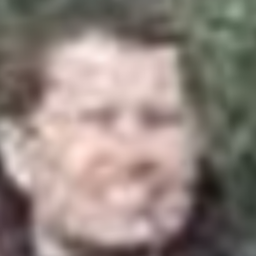}} &
        \raisebox{-.0\height}{\includegraphics[width=\imW]{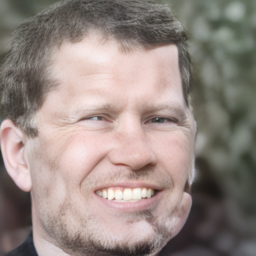}} &
        \raisebox{-.0\height}{\includegraphics[width=\imW]{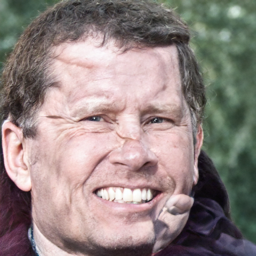}} &
        \raisebox{-.0\height}{\includegraphics[width=\imW]{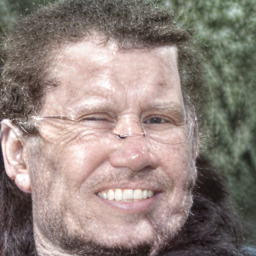}} &
        \raisebox{-.0\height}{\includegraphics[width=\imW]{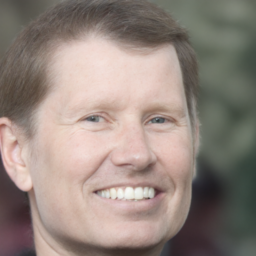}} &
        \raisebox{-.0\height}{\includegraphics[width=\imW]{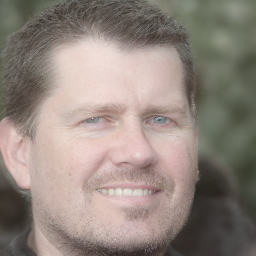}} \\
        \raisebox{-.0\height}{\includegraphics[width=\imW]{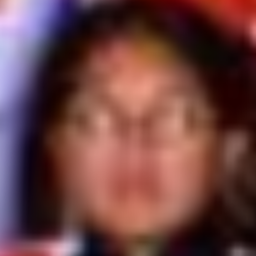}} &
        \raisebox{-.0\height}{\includegraphics[width=\imW]{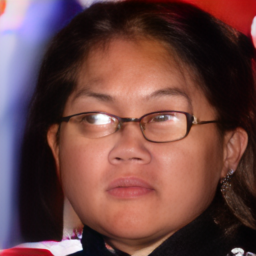}} &
        \raisebox{-.0\height}{\includegraphics[width=\imW]{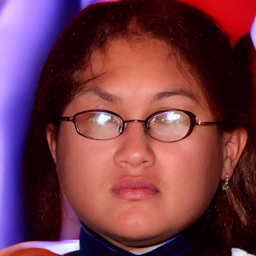}} &
        \raisebox{-.0\height}{\includegraphics[width=\imW]{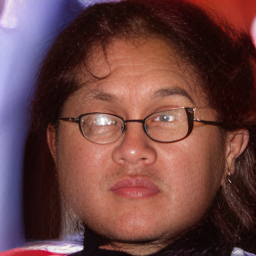}} &
        \raisebox{-.0\height}{\includegraphics[width=\imW]{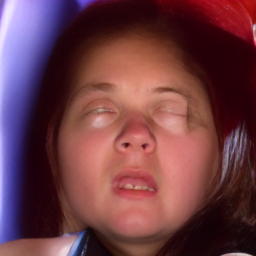}} &
        \raisebox{-.0\height}{\includegraphics[width=\imW]{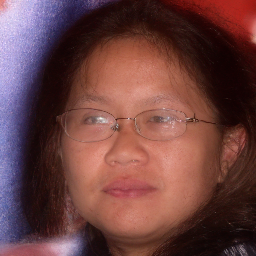}} \\
        \raisebox{-.0\height}{\includegraphics[width=\imW]{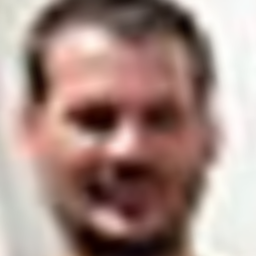}} &
        \raisebox{-.0\height}{\includegraphics[width=\imW]{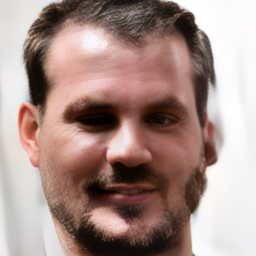}} &
        \raisebox{-.0\height}{\includegraphics[width=\imW]{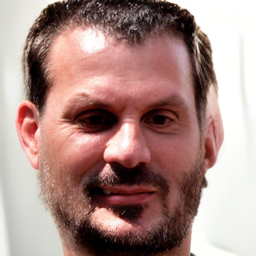}} &
        \raisebox{-.0\height}{\includegraphics[width=\imW]{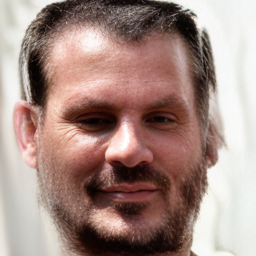}} &
        \raisebox{-.0\height}{\includegraphics[width=\imW]{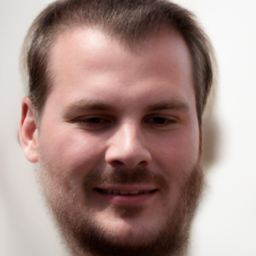}} &
        \raisebox{-.0\height}{\includegraphics[width=\imW]{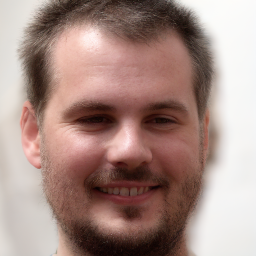}} \\
        \raisebox{-.0\height}{\includegraphics[width=\imW]{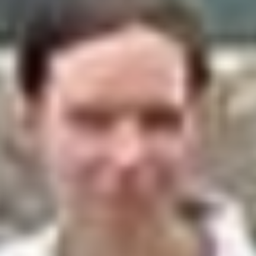}} &
        \raisebox{-.0\height}{\includegraphics[width=\imW]{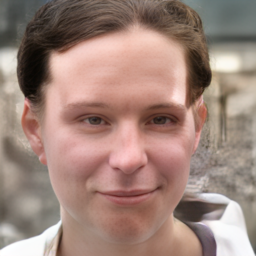}} &
        \raisebox{-.0\height}{\includegraphics[width=\imW]{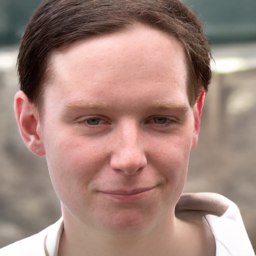}} &
        \raisebox{-.0\height}{\includegraphics[width=\imW]{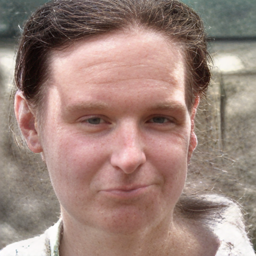}} &
        \raisebox{-.0\height}{\includegraphics[width=\imW]{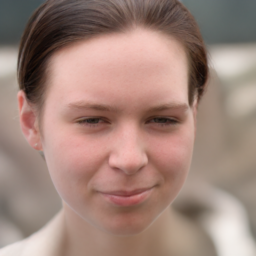}} &
        \raisebox{-.0\height}{\includegraphics[width=\imW]{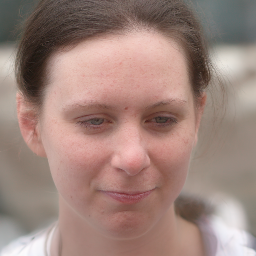}} \\
        \raisebox{-.0\height}{\includegraphics[width=\imW]{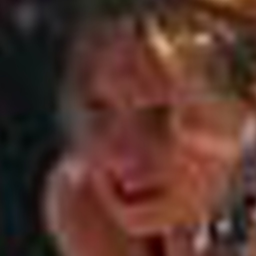}} &
        \raisebox{-.0\height}{\includegraphics[width=\imW]{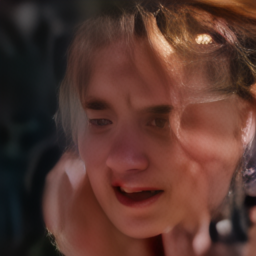}} &
        \raisebox{-.0\height}{\includegraphics[width=\imW]{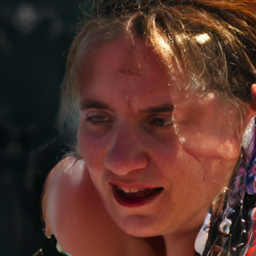}} &
        \raisebox{-.0\height}{\includegraphics[width=\imW]{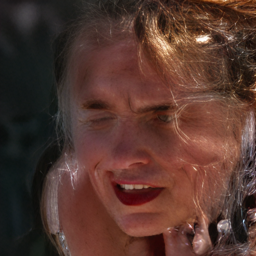}} &
        \raisebox{-.0\height}{\includegraphics[width=\imW]{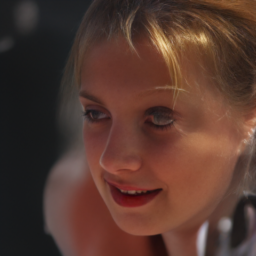}} &
        \raisebox{-.0\height}{\includegraphics[width=\imW]{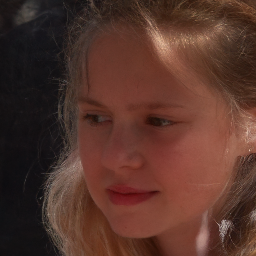}} \\
        \raisebox{-.0\height}{\includegraphics[width=\imW]{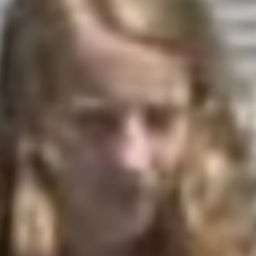}} &
        \raisebox{-.0\height}{\includegraphics[width=\imW]{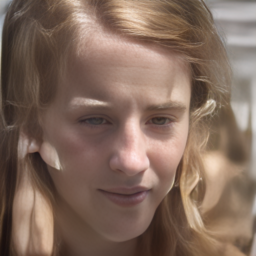}} &
        \raisebox{-.0\height}{\includegraphics[width=\imW]{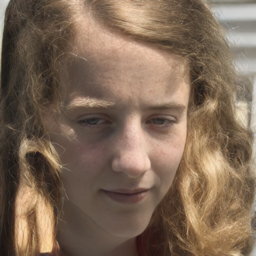}} &
        \raisebox{-.0\height}{\includegraphics[width=\imW]{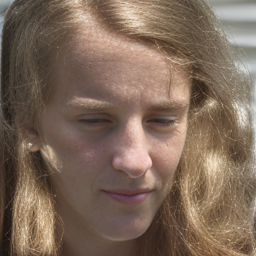}} &
        \raisebox{-.0\height}{\includegraphics[width=\imW]{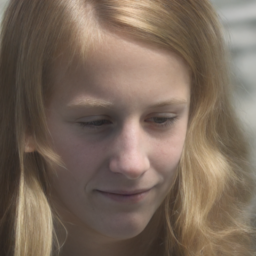}} &
        \raisebox{-.0\height}{\includegraphics[width=\imW]{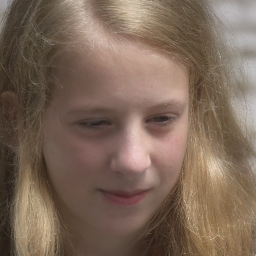}} \\
        Input & GFPGAN\cite{wang2021gfpgan} & VQFR\cite{gu2022vqfr} & CodeFormer\cite{zhou2022towards} & DR2(+VQFR)\cite{wang2023dr2}  & Ours \\
    \end{tabular}
    \vspace{-7pt} 
    \caption{\textbf{More qualitative comparison with previous methods on Wider-Test.} }
    \vspace{-4mm} 
    \label{fig:sup_widertest}
\end{figure*}

\begin{figure*}[ht]
    \centering
    \setlength{\tabcolsep}{1pt}
    \def\imW{0.25\linewidth}
    \scalebox{0.8}{
    \begin{tabular}{cccc}
        \raisebox{-.0\height}{\includegraphics[width=\imW]{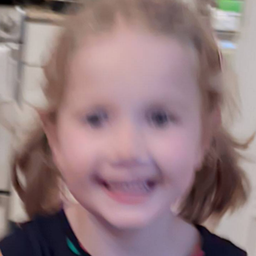}} &
        \raisebox{-.0\height}{\includegraphics[width=\imW]{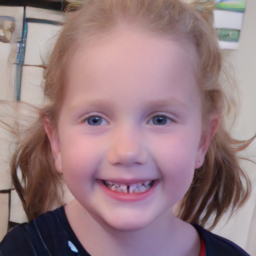}} &
        \raisebox{-.0\height}{\includegraphics[width=\imW]{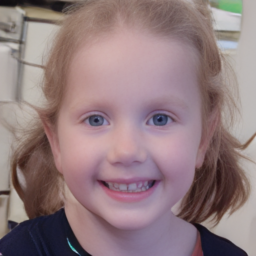}} &
        \raisebox{-.0\height}{\includegraphics[width=\imW]{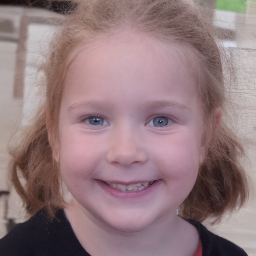}} \\
        \raisebox{-.0\height}{\includegraphics[width=\imW]{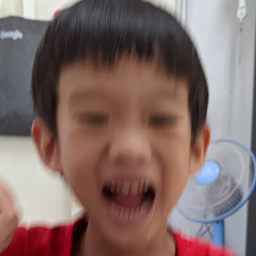}} &
        \raisebox{-.0\height}{\includegraphics[width=\imW]{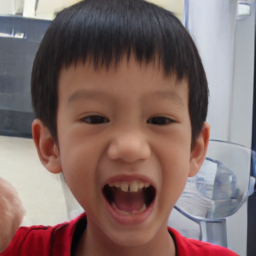}} &
        \raisebox{-.0\height}{\includegraphics[width=\imW]{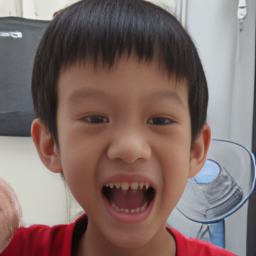}} &
        \raisebox{-.0\height}{\includegraphics[width=\imW]{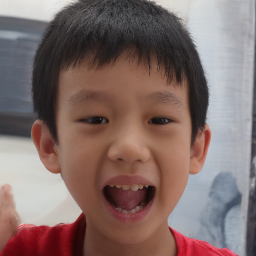}} \\
        \raisebox{-.0\height}{\includegraphics[width=\imW]{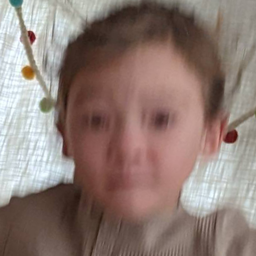}} &
        \raisebox{-.0\height}{\includegraphics[width=\imW]{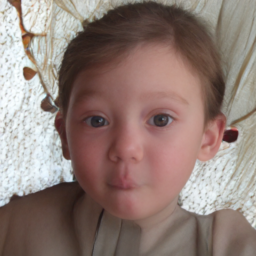}} &
        \raisebox{-.0\height}{\includegraphics[width=\imW]{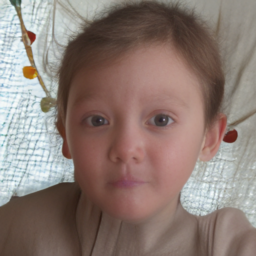}} &
        \raisebox{-.0\height}{\includegraphics[width=\imW]{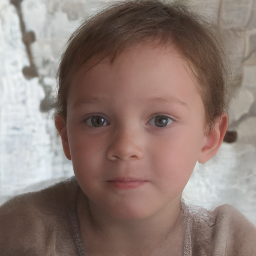}} \\
        \raisebox{-.0\height}{\includegraphics[width=\imW]{Figs/fusiondeblur/fd1_input.png}} &
        \raisebox{-.0\height}{\includegraphics[width=\imW]{Figs/fusiondeblur/fd1_vqfr.png}} &
        \raisebox{-.0\height}{\includegraphics[width=\imW]{Figs/fusiondeblur/fd1_codeformer.png}} &
        \raisebox{-.0\height}{\includegraphics[width=\imW]{Figs/fusiondeblur/fd1_ours.png}} \\
        Input & VQFR\cite{gu2022vqfr} & CodeFormer\cite{zhou2022towards}  & Ours \\
    \end{tabular}
    }
    \caption{\textbf{More qualitative comparison with previous methods on Deblur-Test.}}
    \label{fig:sup_deblur}
\end{figure*}

We provide additional qualitative comparisons on Wider-Test dataset in Figure~\ref{fig:sup_widertest} and the Deblur-Test dataset in Figure~\ref{fig:sup_deblur}. Wider-Test contains 970 images selected from the Wider-Face\cite{yang2016wider} dataset which is initially collected for face detection that contains many real-world low-quality face images. The Deblur-Test dataset contains 67 real-world motion-blur images from \cite{lai2022face}. Both of the datasets are aligned using the same way in FFHQ\cite{ffhq}. All the previous methods we compare are synthetic-data-based methods. As our method does not utilize synthetic data which previous methods rely on, our method shows good generalizability in handling different kinds of real-world degraded images.

\subsection{More Distortion Types}

\begin{figure}[ht]
\vspace{-3mm}
\centering
\setlength{\tabcolsep}{0.2pt}
\def\imW{0.22\linewidth}
\begin{tabular}{cccc}
\includegraphics[width=\imW]{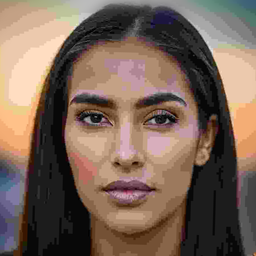}
&
\includegraphics[width=\imW]{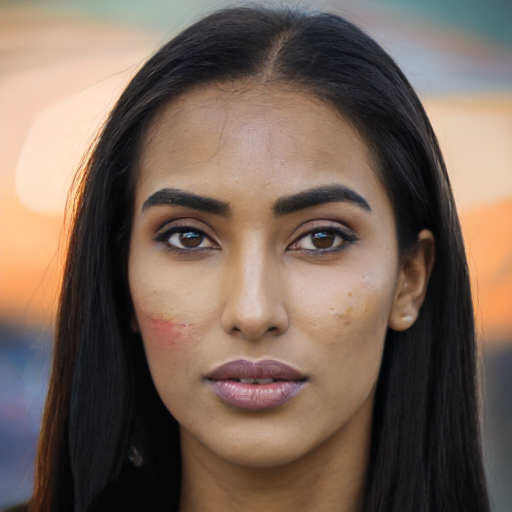}
&
\includegraphics[width=\imW]{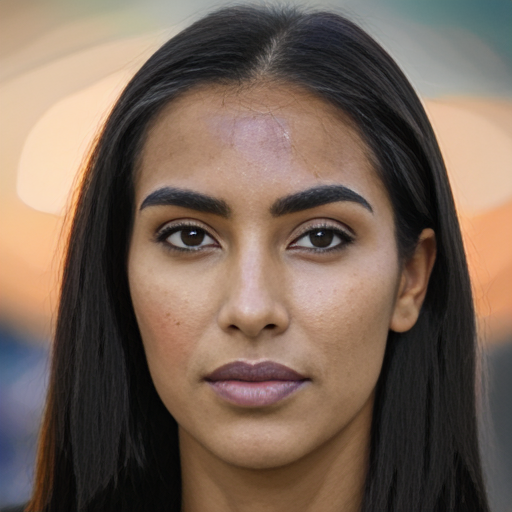}
&
\includegraphics[width=\imW]{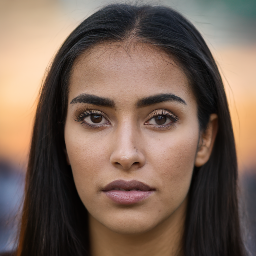}
\\
\includegraphics[width=\imW]{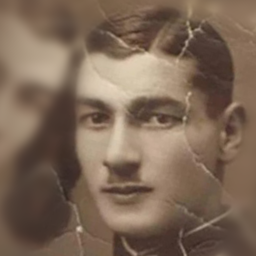}
&
\includegraphics[width=\imW]{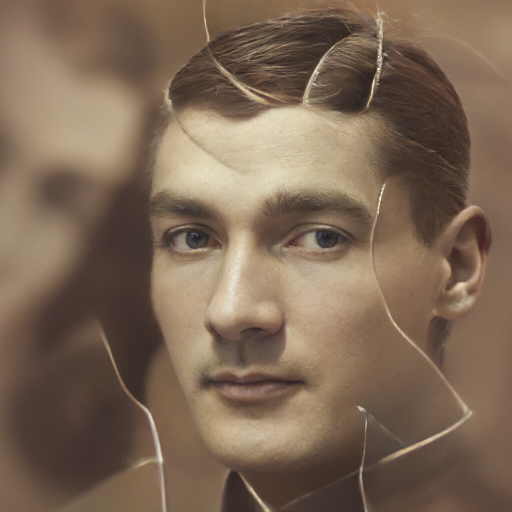}
&
\includegraphics[width=\imW]{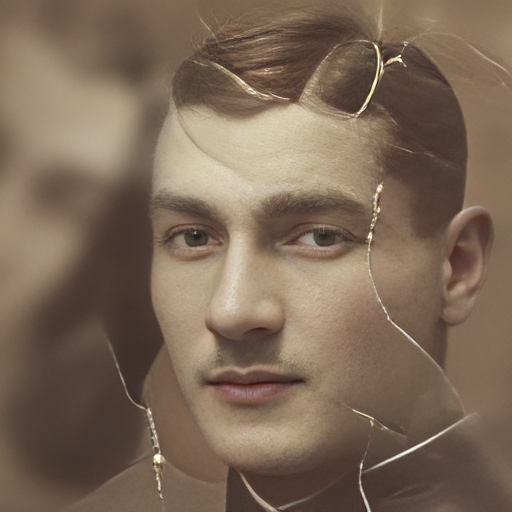}
&
\includegraphics[width=\imW]{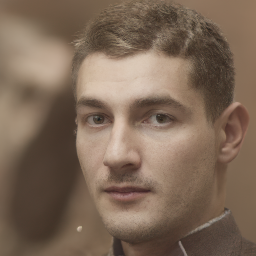}
\\
Input & VQFR & CodeFormer & Ours
\end{tabular}
\vspace{-2mm}
\caption{\scriptsize Results on JPEG compression(top) and scratches(bottom).}
\label{fig:more_dist}
\vspace{-4mm}
\end{figure}

We further provide results on more distortion types e.g., JPEG compression and scratches in Figure \ref{fig:more_dist}. This further demonstrates that our method is able to perform better on out-of-distribution input low-quality images as we don't utilize synthetic data for training.

\section{Additional Results on Personalized Blind Face Restoration}
\vspace{1mm}

\begin{figure*}[ht]
    \centering
    \setlength{\tabcolsep}{1pt}
    \def\imW{0.17\linewidth}
    \scalebox{1.0}{
    \begin{tabular}{ccccccc}
        \raisebox{-.0\height}{\includegraphics[width=\imW]{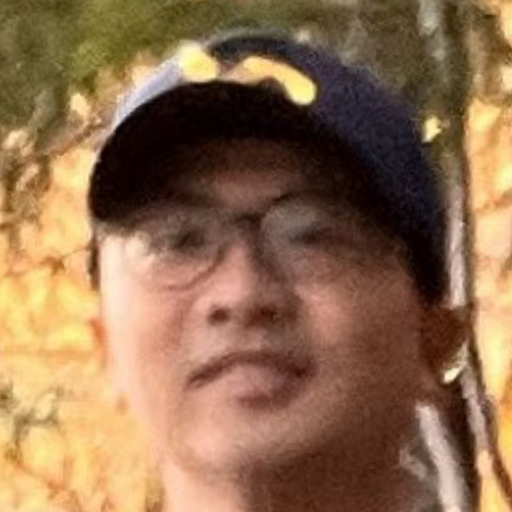}} &
        \raisebox{-.0\height}{\includegraphics[width=\imW]{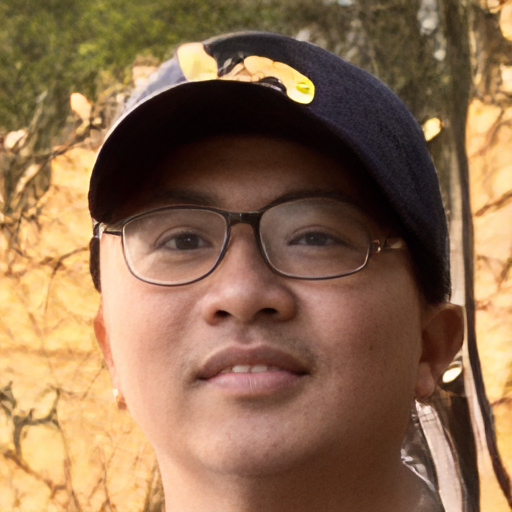}} &
        \raisebox{-.0\height}{\includegraphics[width=\imW]{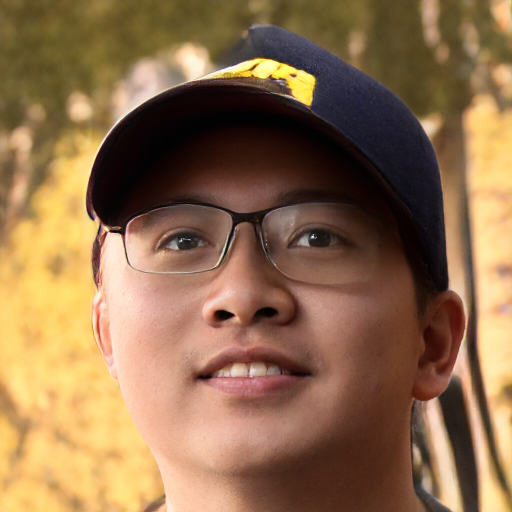}} &
        \raisebox{-.0\height}{\includegraphics[width=\imW]{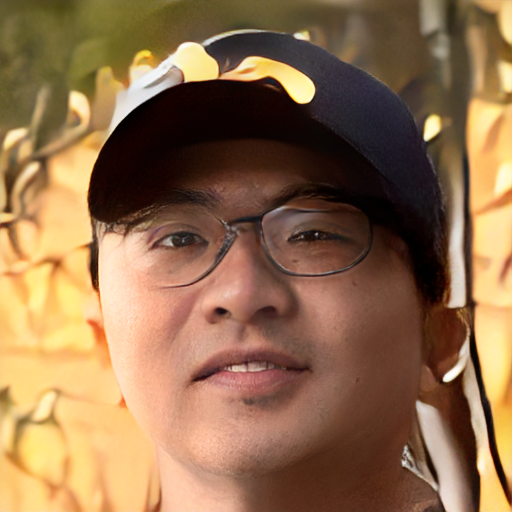}} &
        \raisebox{-.0\height}{\includegraphics[width=\imW]{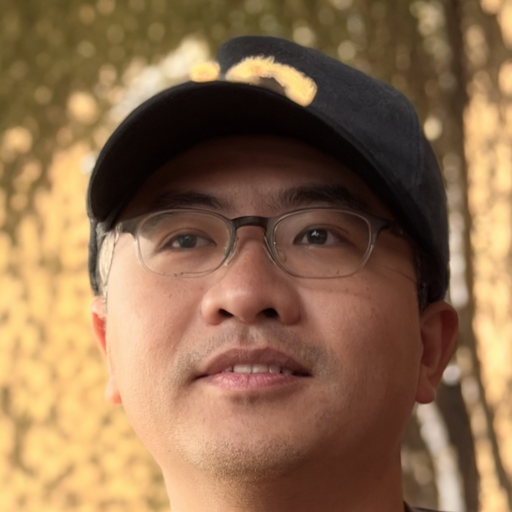}} \\
        \raisebox{-.0\height}{\includegraphics[width=\imW]{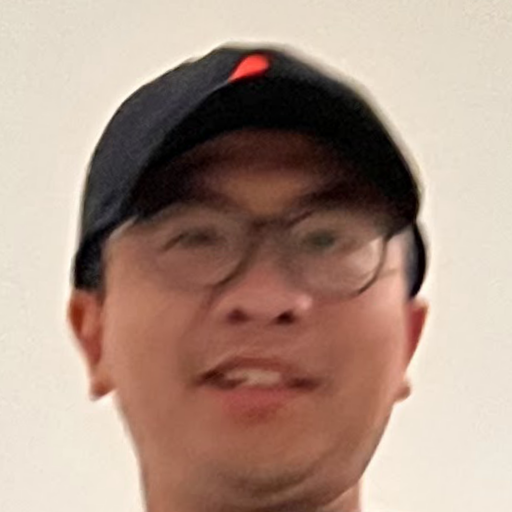}} &
        \raisebox{-.0\height}{\includegraphics[width=\imW]{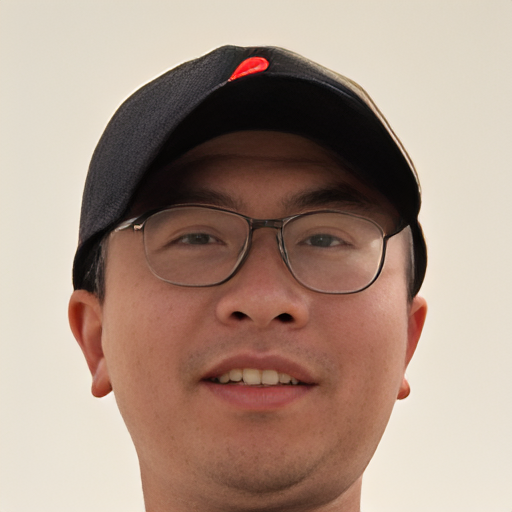}} &
        \raisebox{-.0\height}{\includegraphics[width=\imW]{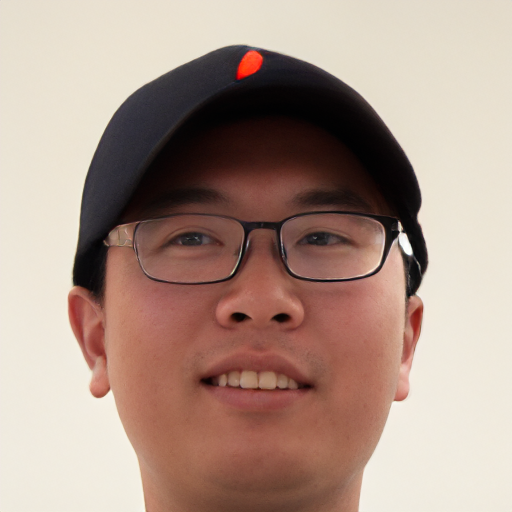}} &
        \raisebox{-.0\height}{\includegraphics[width=\imW]{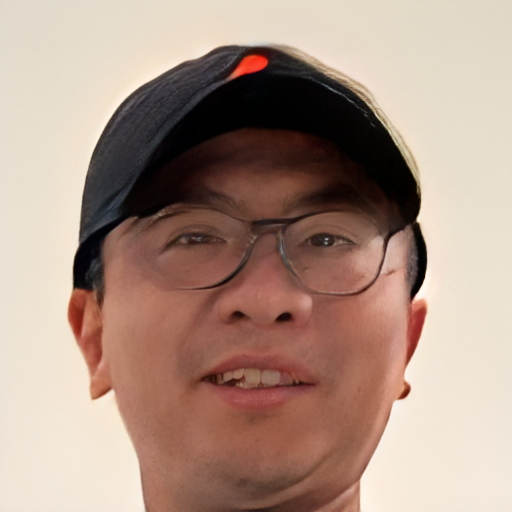}} &
        \raisebox{-.0\height}{\includegraphics[width=\imW]{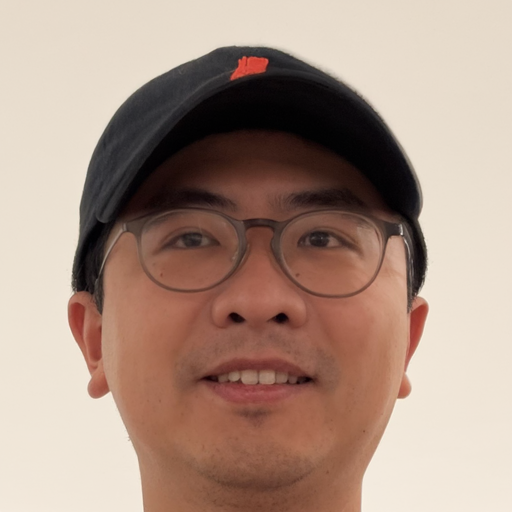}} \vspace{5pt} \\
        \raisebox{-.0\height}{\includegraphics[width=\imW]{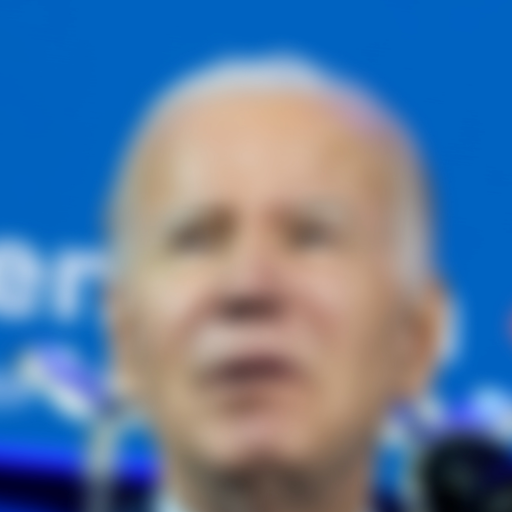}} &
        \raisebox{-.0\height}{\includegraphics[width=\imW]{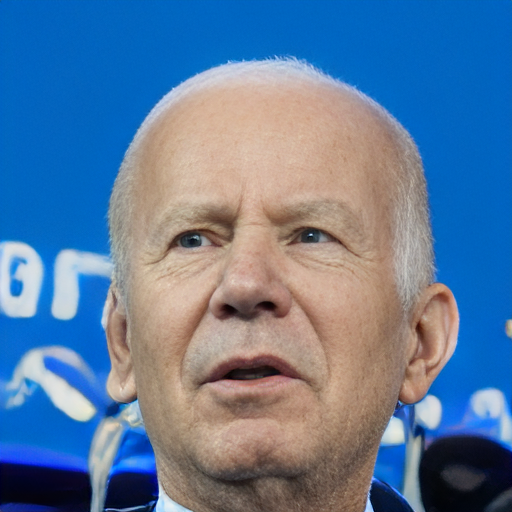}} &
        \raisebox{-.0\height}{\includegraphics[width=\imW]{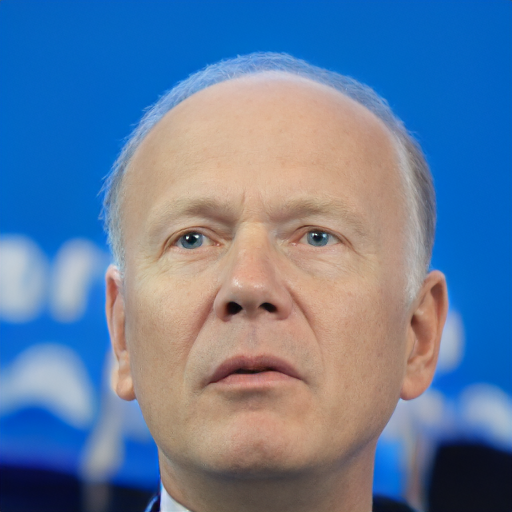}} &
        \raisebox{-.0\height}{\includegraphics[width=\imW]{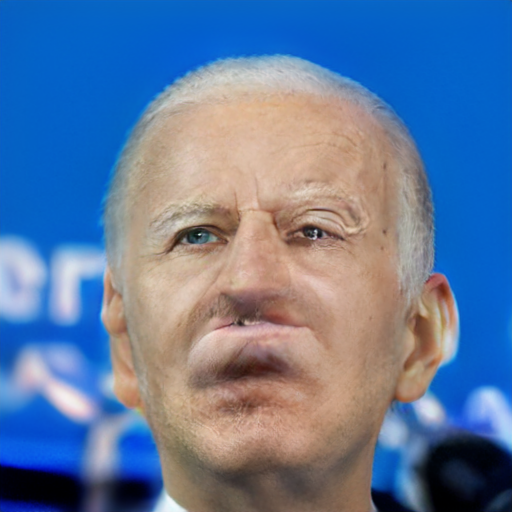}} &
        \raisebox{-.0\height}{\includegraphics[width=\imW]{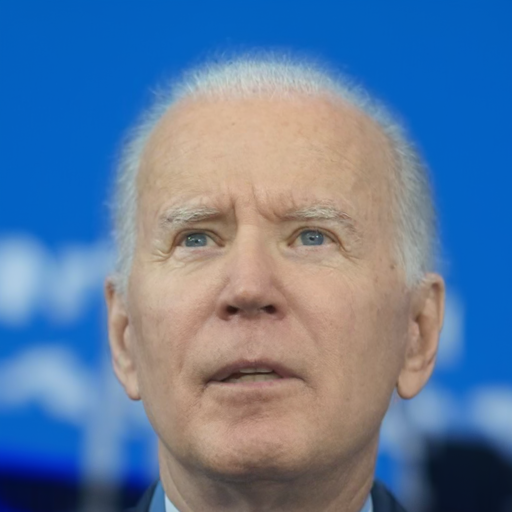}} \\
        \raisebox{-.0\height}{\includegraphics[width=\imW]{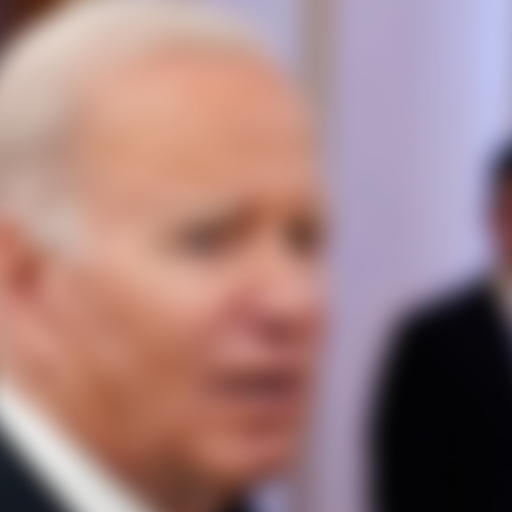}} &
        \raisebox{-.0\height}{\includegraphics[width=\imW]{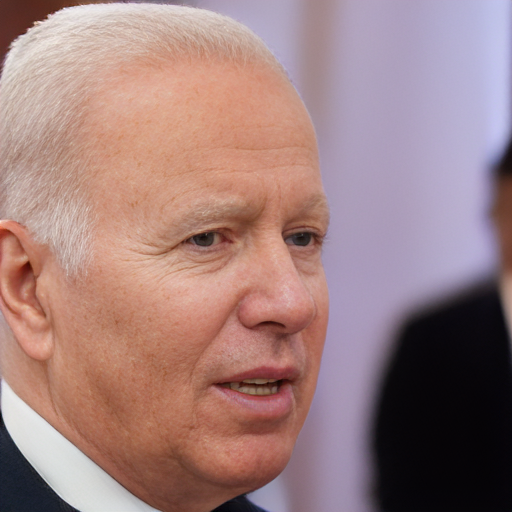}} &
        \raisebox{-.0\height}{\includegraphics[width=\imW]{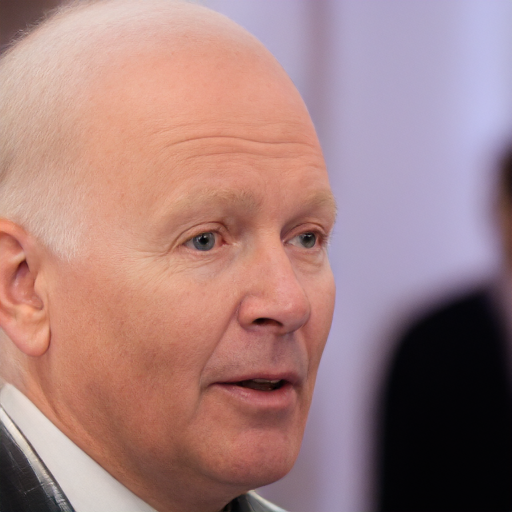}} &
        \raisebox{-.0\height}{\includegraphics[width=\imW]{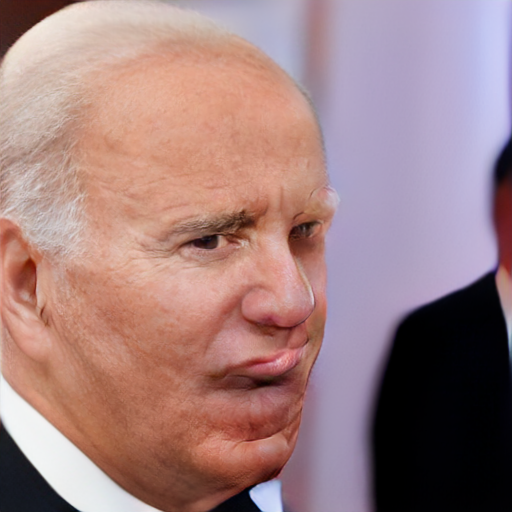}} &
        \raisebox{-.0\height}{\includegraphics[width=\imW]{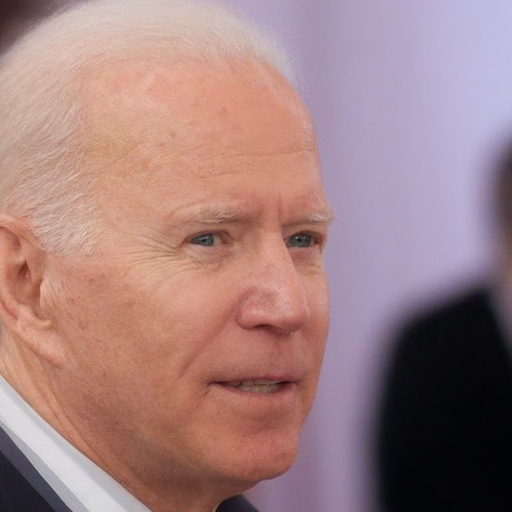}} \vspace{5pt} \\
        \raisebox{-.0\height}{\includegraphics[width=\imW]{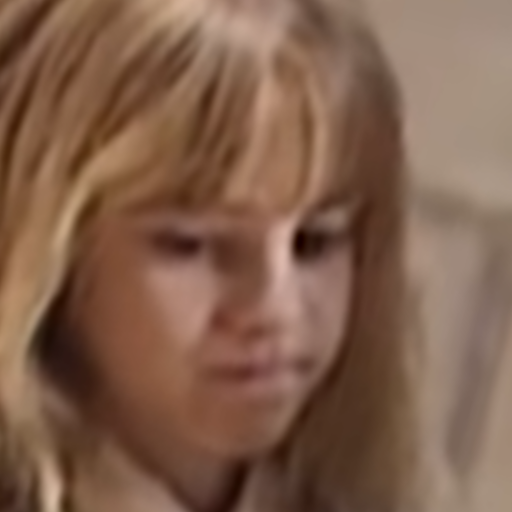}} &
        \raisebox{-.0\height}{\includegraphics[width=\imW]{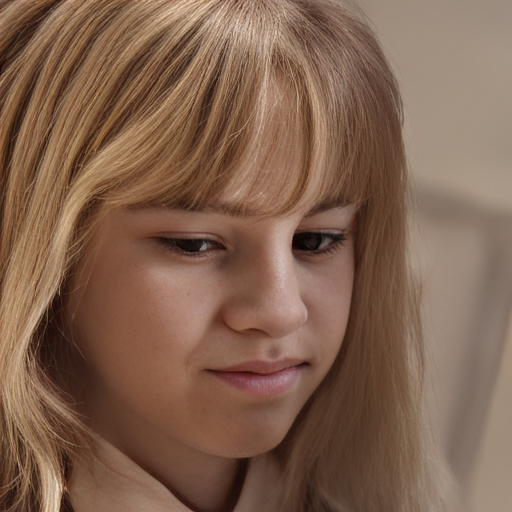}} &
        \raisebox{-.0\height}{\includegraphics[width=\imW]{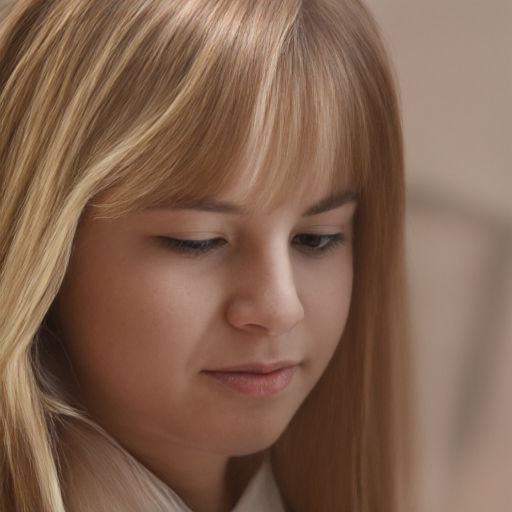}} &
        \raisebox{-.0\height}{\includegraphics[width=\imW]{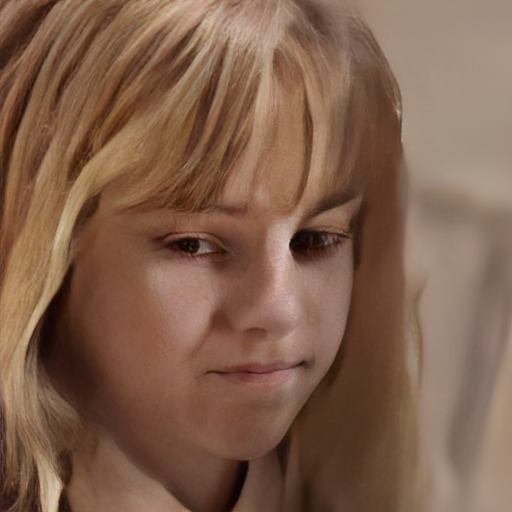}} &
        \raisebox{-.0\height}{\includegraphics[width=\imW]{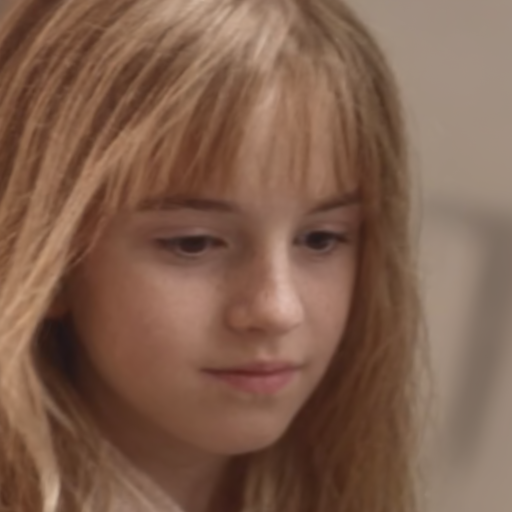}} \\
        \raisebox{-.0\height}{\includegraphics[width=\imW]{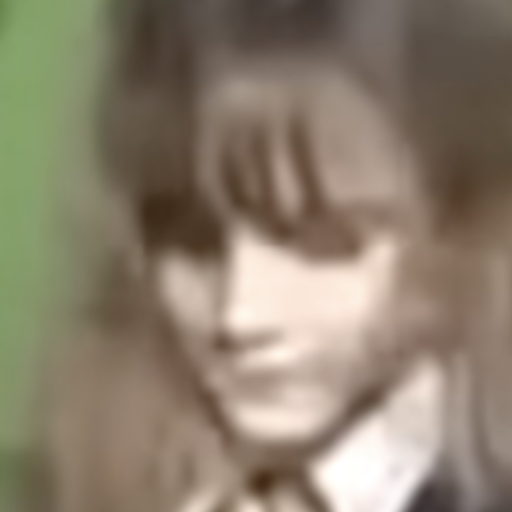}} &
        \raisebox{-.0\height}{\includegraphics[width=\imW]{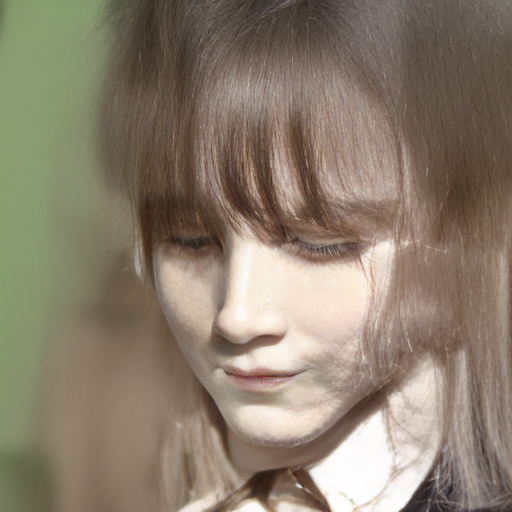}} &
        \raisebox{-.0\height}{\includegraphics[width=\imW]{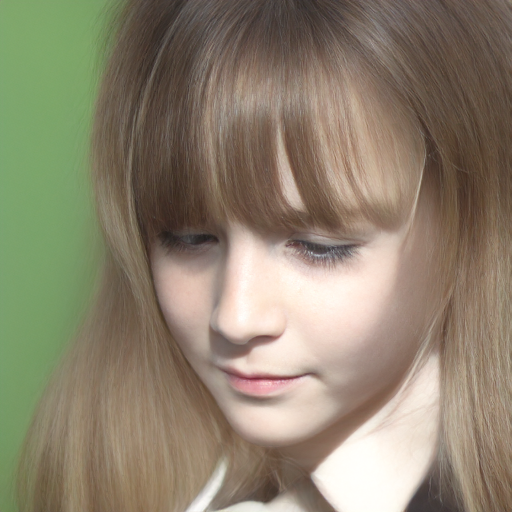}} &
        \raisebox{-.0\height}{\includegraphics[width=\imW]{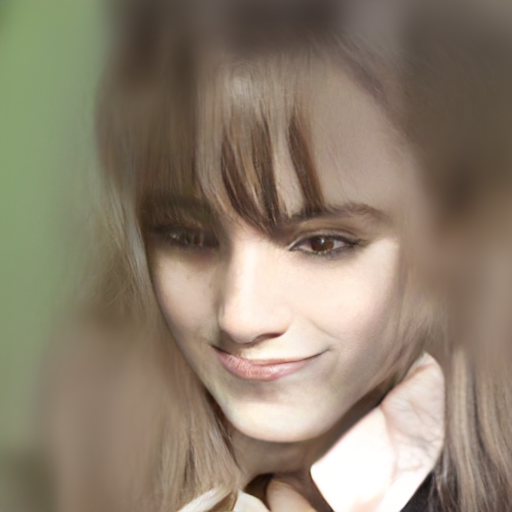}} &
        \raisebox{-.0\height}{\includegraphics[width=\imW]{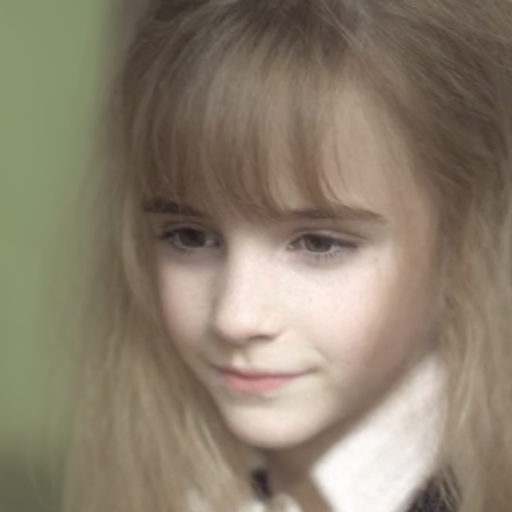}} \\
        Input  & CodeFormer\cite{zhou2022towards} & DR2(+VQFR)\cite{wang2023dr2} & ASFFNet\cite{li2020enhanced} & Ours \\
    \end{tabular}
    }
    \caption{\textbf{More Qualitative Comparison on Personalized Face Restoration.} We present three subjects here. For each subjects, we compare two real-world low-quality images with previous methods. The subjects from the top to bottom are: subject B referenced in the main paper, Biden and Hermione. }
    \label{fig:sup_personal}
\end{figure*}

In this section, we present more personalized restoration results involving additional subjects. This includes Subject B (a man), as well as public figures such as Biden and Hermione. These results are showcased in Figure \ref{fig:sup_personal}. We compare our method with previous methods CodeFormer\cite{zhou2022towards} and DR2(+VQFR) \cite{wang2023dr2} which are two single-image-based restoration methods that rely on synthetic data as well as ASFFNet\cite{li2020enhanced} that requires a reference dataset. We use the same personal album in ASFFNet and our method. Our findings demonstrate a superior preservation of identity, along with a high level of quality in the results.

\section{Anchor Images and Constrained Generative Space}
\vspace{1mm}

In this section, we discuss both the generative album and the personalized album utilized by the model to restrict the generative space for restoration. Additionally, we visualize this constrained space through unconditional generation from the fine-tuned model.

\begin{figure*}[ht]
    \centering
    \setlength{\tabcolsep}{1pt}
    \def\imW{0.103\linewidth}
    \def\rbx{-.5\height}
    \scalebox{1.0}{
    \begin{tabular}{c@{\hskip 6pt}cccc@{\hskip 6pt}cccc}
        \multirow[c]{2}{*}[-2mm]{\raisebox{0pt}{\includegraphics[width=\imW]{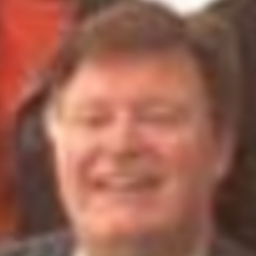}}} &
        \raisebox{\rbx}{\includegraphics[width=\imW]{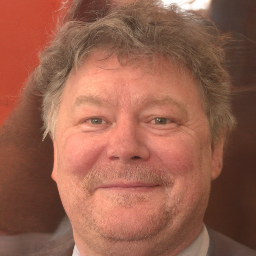}} &
        \raisebox{\rbx}{\includegraphics[width=\imW]{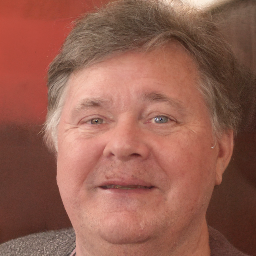}} &
        \raisebox{\rbx}{\includegraphics[width=\imW]{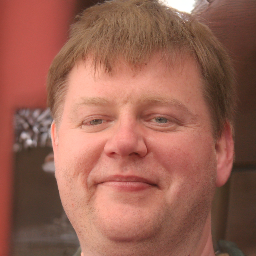}} &
        \raisebox{\rbx}{\includegraphics[width=\imW]{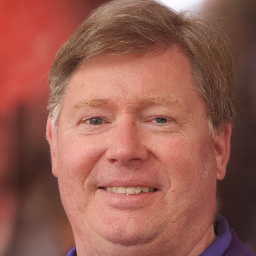}} &
        \raisebox{\rbx}{\includegraphics[width=\imW]{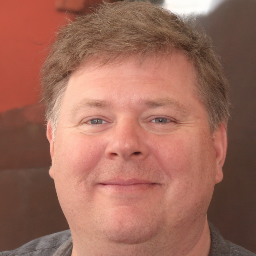}} &
        \raisebox{\rbx}{\includegraphics[width=\imW]{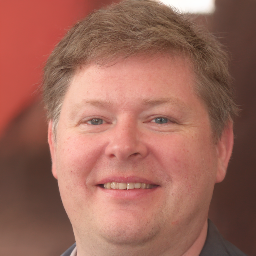}} &
        \raisebox{\rbx}{\includegraphics[width=\imW]{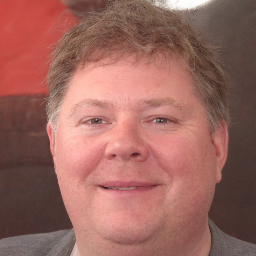}} &
        \raisebox{\rbx}{\includegraphics[width=\imW]{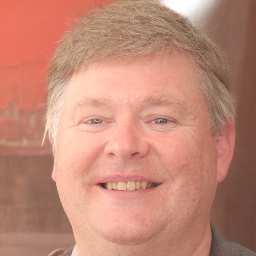}} \vspace{2pt} \\
        &
        \raisebox{\rbx}{\includegraphics[width=\imW]{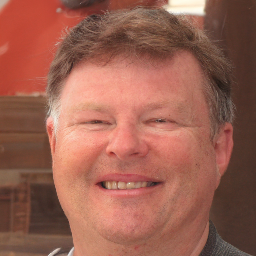}} &
        \raisebox{\rbx}{\includegraphics[width=\imW]{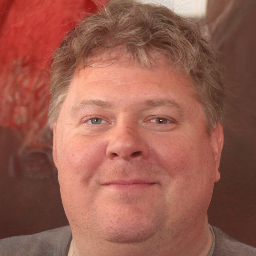}} &
        \raisebox{\rbx}{\includegraphics[width=\imW]{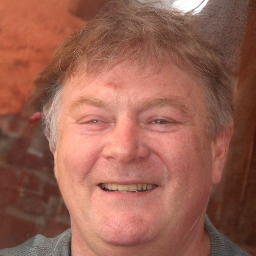}} &
        \raisebox{\rbx}{\includegraphics[width=\imW]{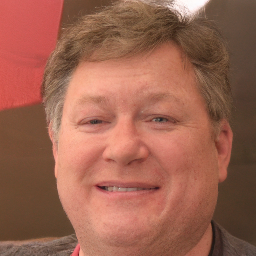}} \vspace{3pt} &
        \raisebox{\rbx}{\includegraphics[width=\imW]{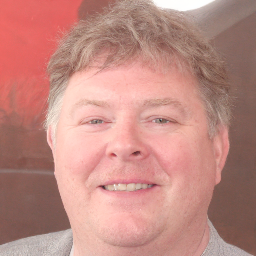}} &
        \raisebox{\rbx}{\includegraphics[width=\imW]{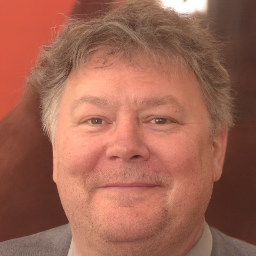}} &
        \raisebox{\rbx}{\includegraphics[width=\imW]{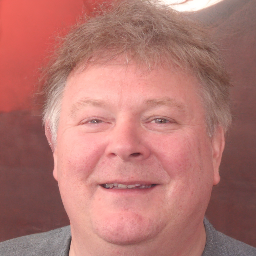}} &
        \raisebox{\rbx}{\includegraphics[width=\imW]{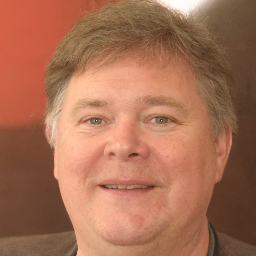}} \vspace{3pt} \\
        \multirow[c]{2}{*}[-2mm]{\raisebox{0pt}{\includegraphics[width=\imW]{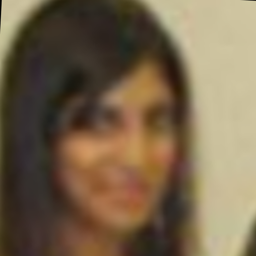}}} &
        \raisebox{\rbx}{\includegraphics[width=\imW]{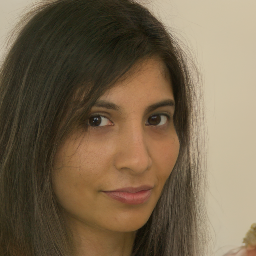}} &
        \raisebox{\rbx}{\includegraphics[width=\imW]{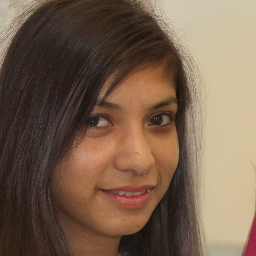}} &
        \raisebox{\rbx}{\includegraphics[width=\imW]{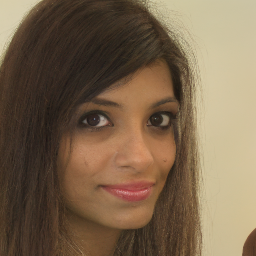}} &
        \raisebox{\rbx}{\includegraphics[width=\imW]{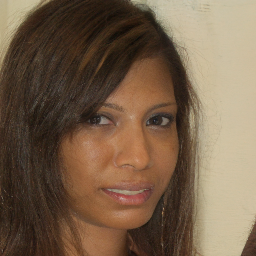}} &
        \raisebox{\rbx}{\includegraphics[width=\imW]{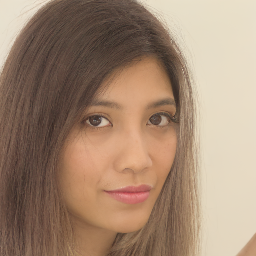}} &
        \raisebox{\rbx}{\includegraphics[width=\imW]{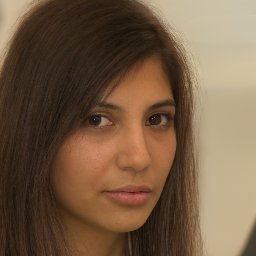}} &
        \raisebox{\rbx}{\includegraphics[width=\imW]{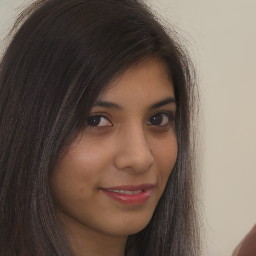}} &
        \raisebox{\rbx}{\includegraphics[width=\imW]{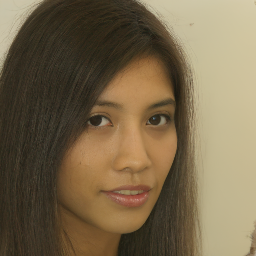}} 
        \vspace{2pt} \\
        &
        \raisebox{\rbx}{\includegraphics[width=\imW]{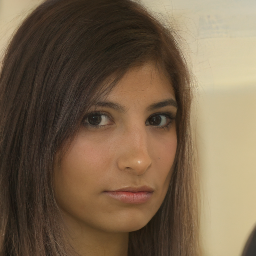}} &
        \raisebox{\rbx}{\includegraphics[width=\imW]{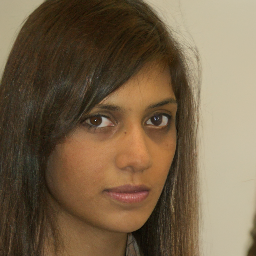}} &
        \raisebox{\rbx}{\includegraphics[width=\imW]{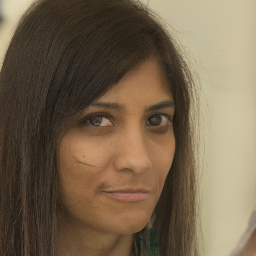}} &
        \raisebox{\rbx}{\includegraphics[width=\imW]{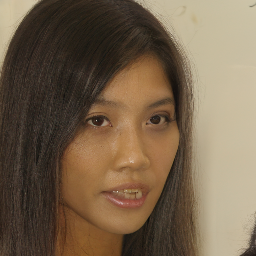}} &
        \raisebox{\rbx}{\includegraphics[width=\imW]{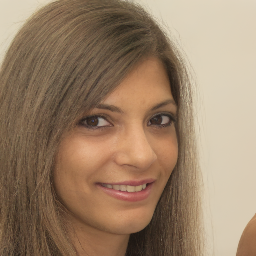}} &
        \raisebox{\rbx}{\includegraphics[width=\imW]{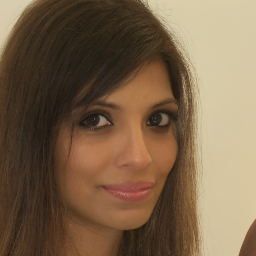}} &
        \raisebox{\rbx}{\includegraphics[width=\imW]{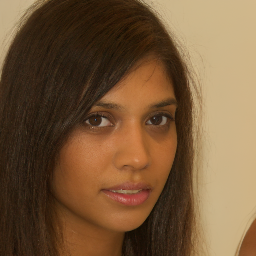}} &
        \raisebox{\rbx}{\includegraphics[width=\imW]{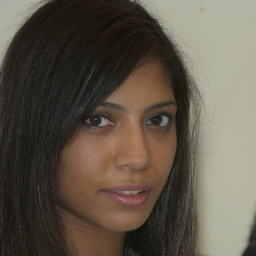}} \\
         Input & \multicolumn{4}{c}{Generative Album for constraining the prior}
         & \multicolumn{4}{c}{Generated Images using the constrained prior} \\
    \end{tabular}
    }
    \caption{\textbf{Generative Album \& Unconditional Generation from Fine-tuned Model.} The generative album is generated with the input image as guidance. Model fine-tuned with this album can then generate high-quality images that are close to the original input.}
    \label{fig:anchor_generative}
\end{figure*}

For the generative album, we employ the input low-quality image with skip guidance to produce anchor images. These images are then used to fine-tune the model. We display the generated album, along with randomly generated images from the fine-tuned model, in Figure~\ref{fig:anchor_generative}. The generated album contains images similar to the input but with enhanced quality, though not as high as those produced by the pre-trained model. This is likely due to the influence of skip guidance. Nevertheless, the model fine-tuned with this album is capable of generating high-quality images that still bear resemblance to the original input.

\begin{figure*}[ht]
    \centering
    \setlength{\tabcolsep}{1pt}
    \def\imW{0.11\linewidth}
    \def\rbx{-.0\height}
    \scalebox{1.0}{
    \begin{tabular}{cccc@{\hskip 6pt}cccc}
        \raisebox{\rbx}{\includegraphics[width=\imW]{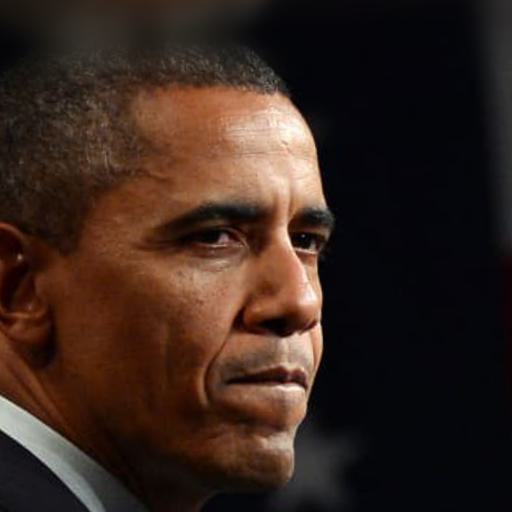}} &
        \raisebox{\rbx}{\includegraphics[width=\imW]{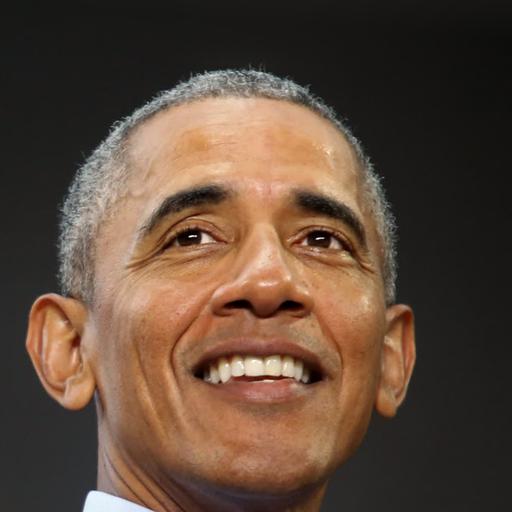}} &
        \raisebox{\rbx}{\includegraphics[width=\imW]{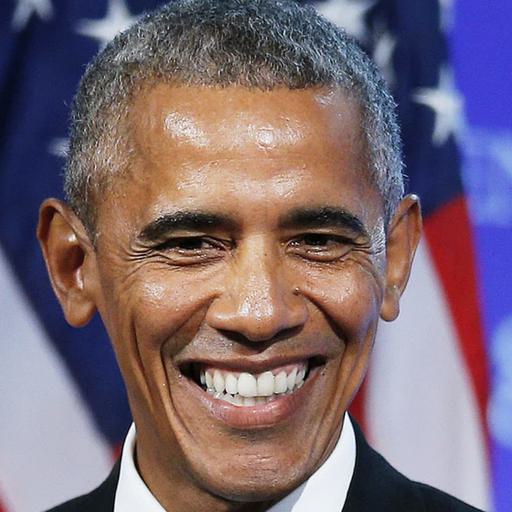}} &
        \raisebox{\rbx}{\includegraphics[width=\imW]{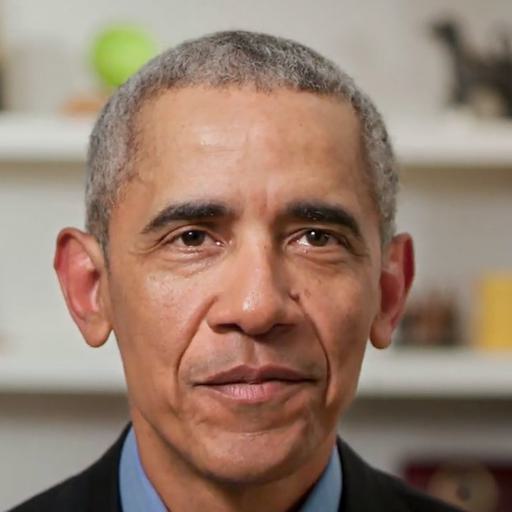}} &
        \raisebox{\rbx}{\includegraphics[width=\imW]{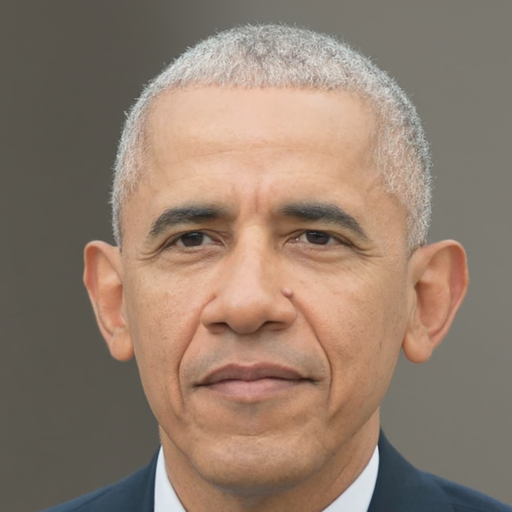}} &
        \raisebox{\rbx}{\includegraphics[width=\imW]{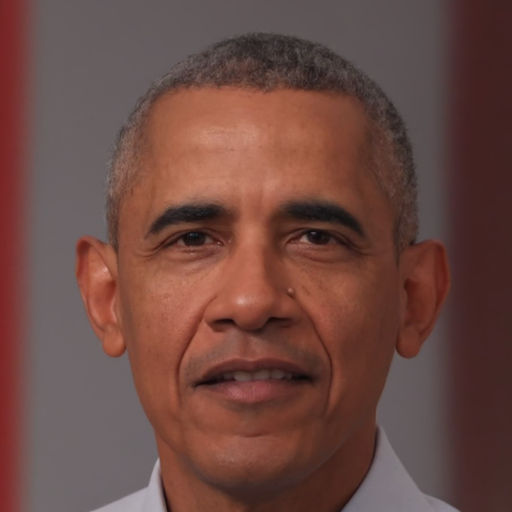}} &
        \raisebox{\rbx}{\includegraphics[width=\imW]{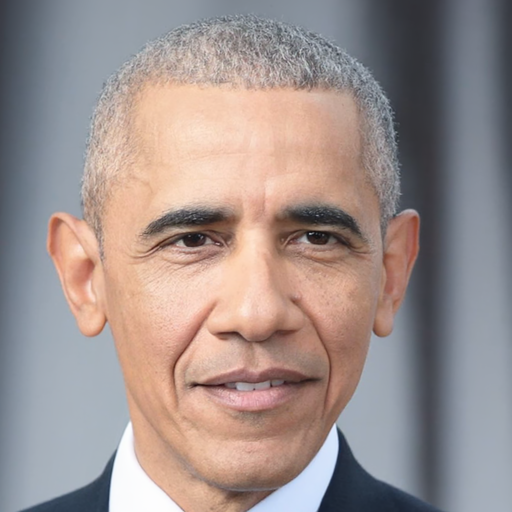}} &
        \raisebox{\rbx}{\includegraphics[width=\imW]{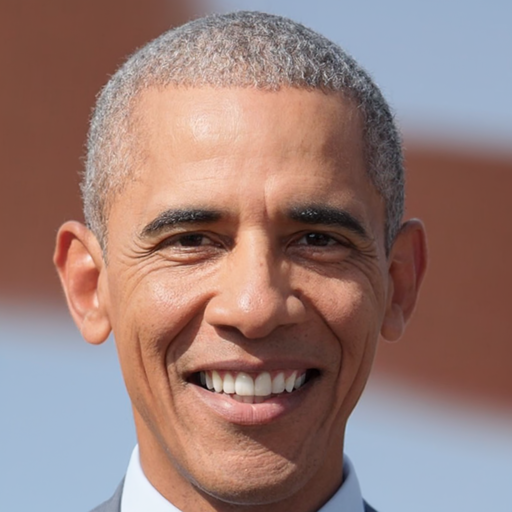}} \\
        \raisebox{\rbx}{\includegraphics[width=\imW]{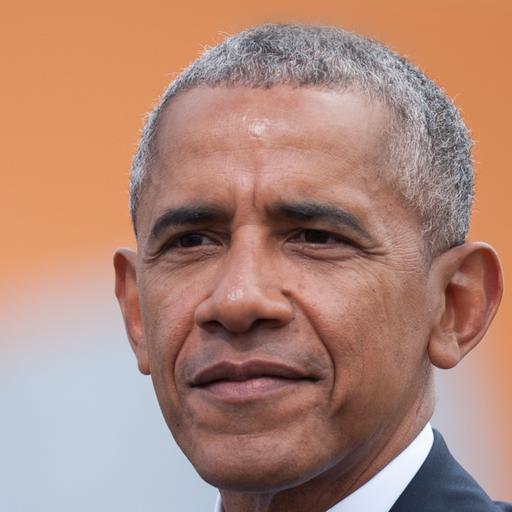}} &
        \raisebox{\rbx}{\includegraphics[width=\imW]{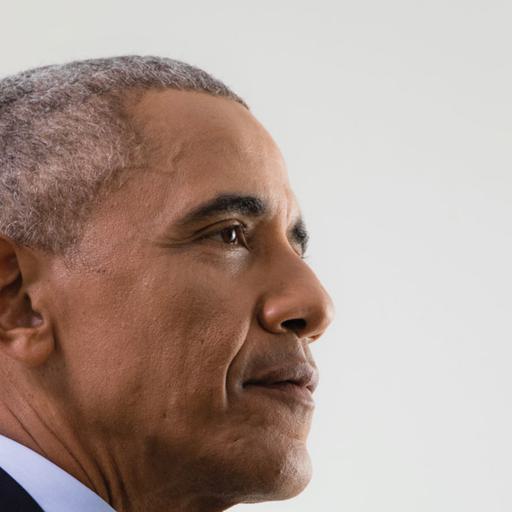}} &
        \raisebox{\rbx}{\includegraphics[width=\imW]{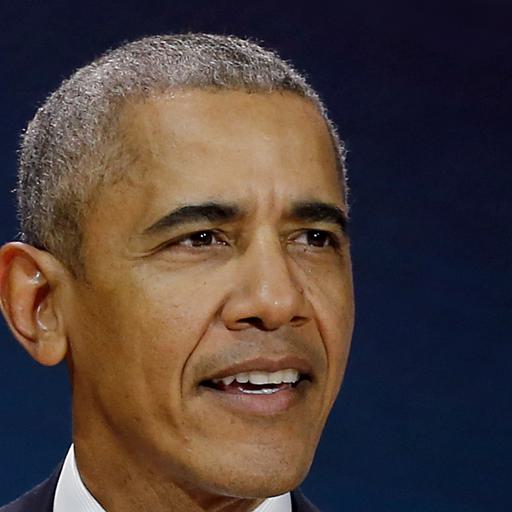}} &
        \raisebox{\rbx}{\includegraphics[width=\imW]{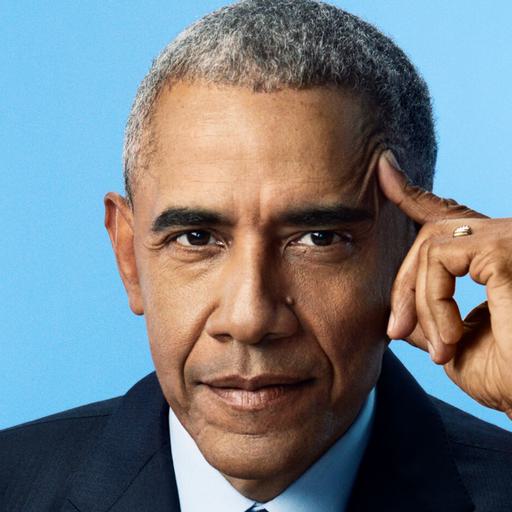}}& 
        \raisebox{\rbx}{\includegraphics[width=\imW]{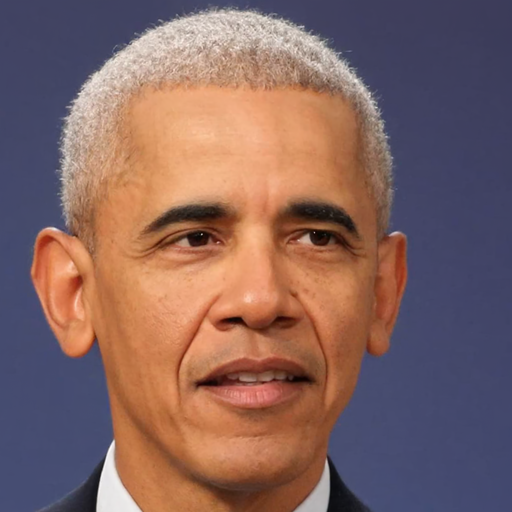}} &
        \raisebox{\rbx}{\includegraphics[width=\imW]{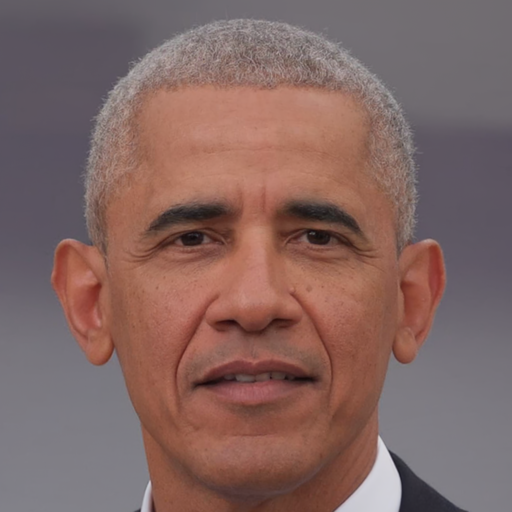}} &
        \raisebox{\rbx}{\includegraphics[width=\imW]{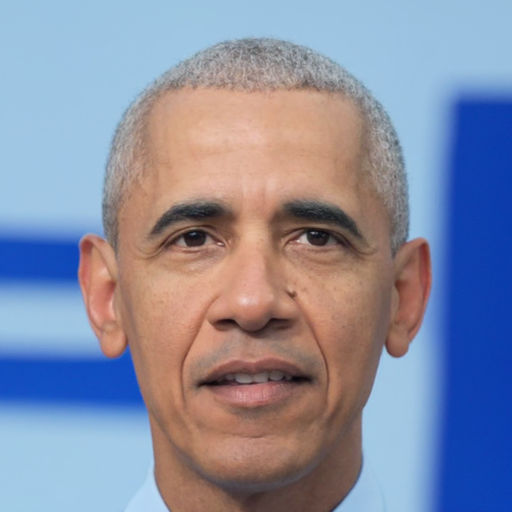}} &
        \raisebox{\rbx}{\includegraphics[width=\imW]{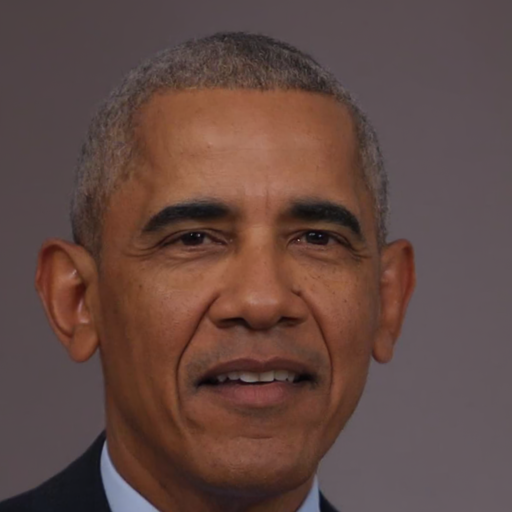}} \vspace{3mm} \\
        \raisebox{\rbx}{\includegraphics[width=\imW]{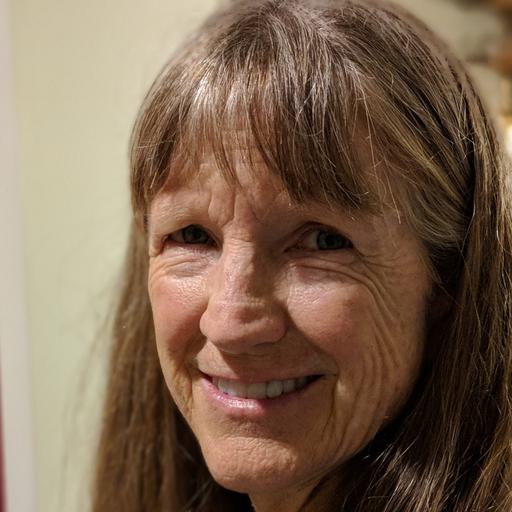}} &
        \raisebox{\rbx}{\includegraphics[width=\imW]{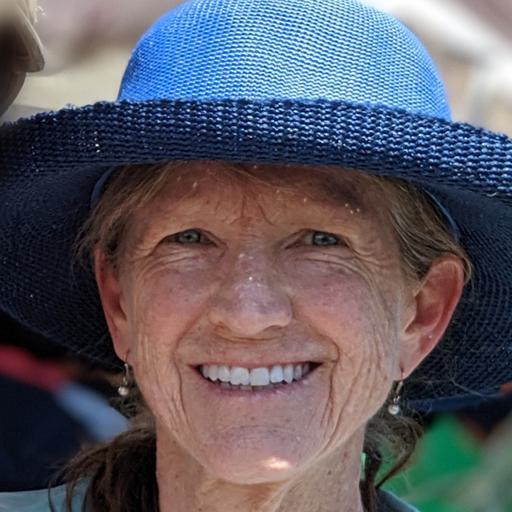}} &
        \raisebox{\rbx}{\includegraphics[width=\imW]{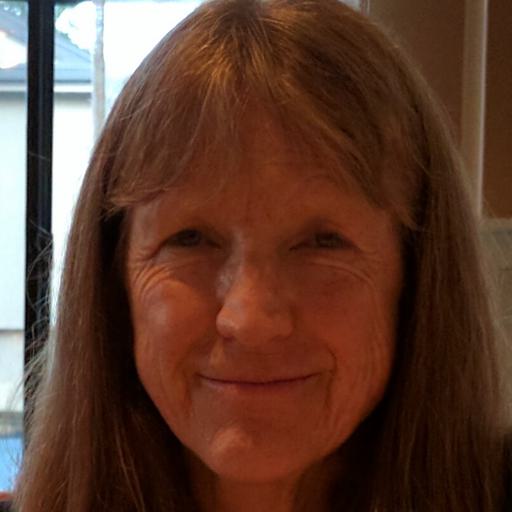}} &
        \raisebox{\rbx}{\includegraphics[width=\imW]{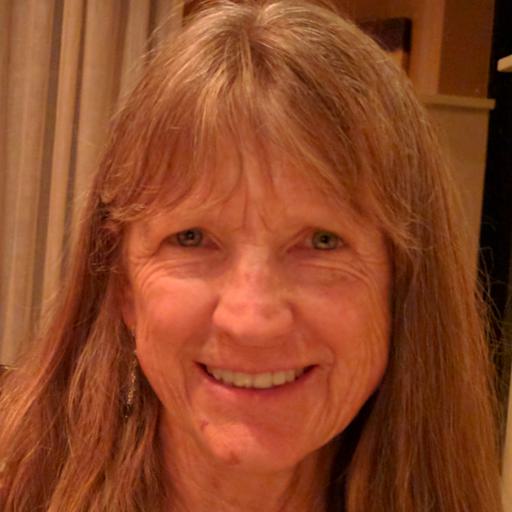}} &
                \raisebox{\rbx}{\includegraphics[width=\imW]{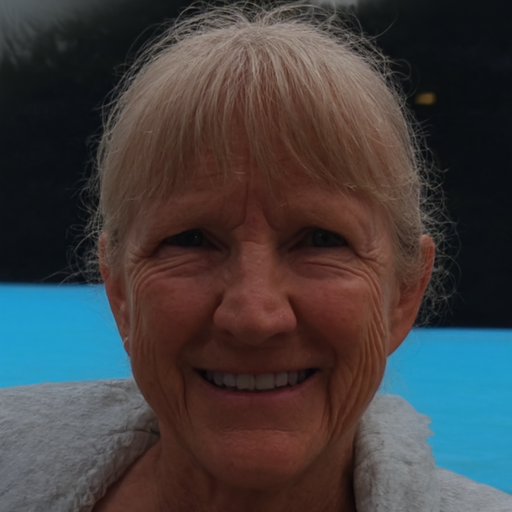}} &
        \raisebox{\rbx}{\includegraphics[width=\imW]{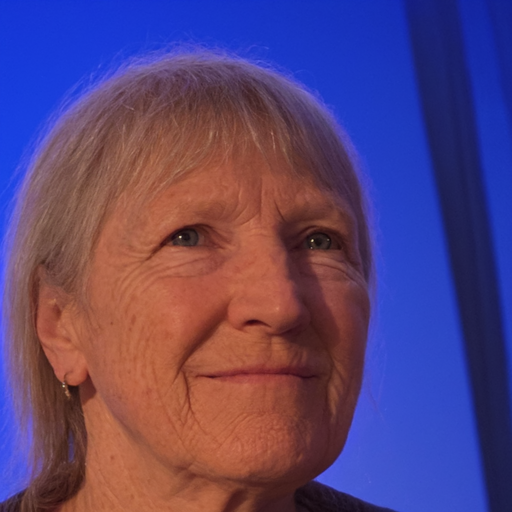}} &
        \raisebox{\rbx}{\includegraphics[width=\imW]{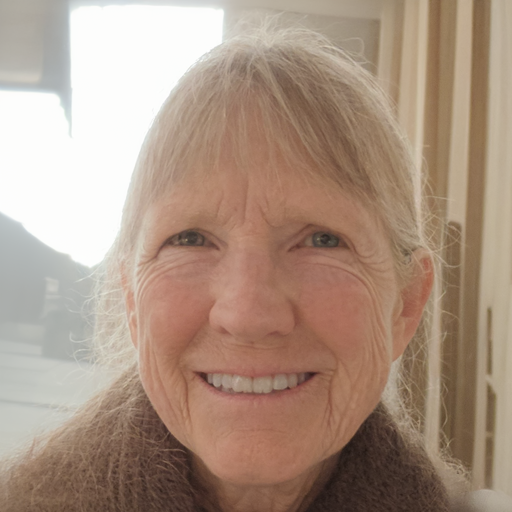}} &
        \raisebox{\rbx}{\includegraphics[width=\imW]{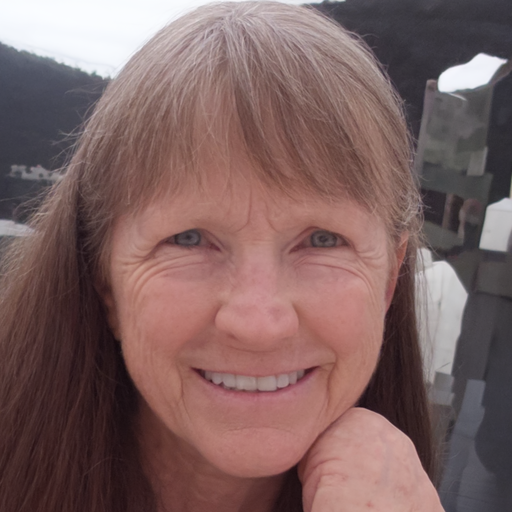}} 
 \\
         \raisebox{\rbx}{\includegraphics[width=\imW]{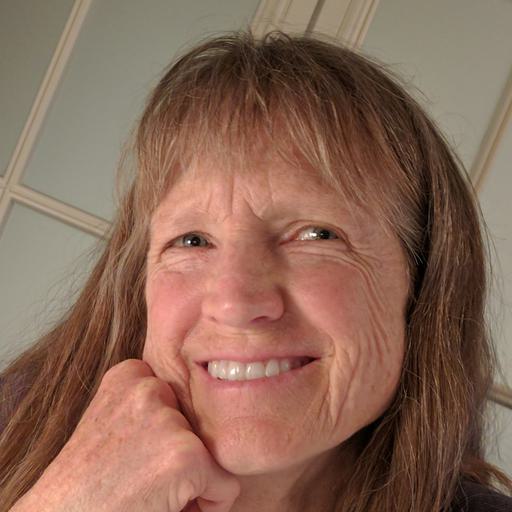}} &
        \raisebox{\rbx}{\includegraphics[width=\imW]{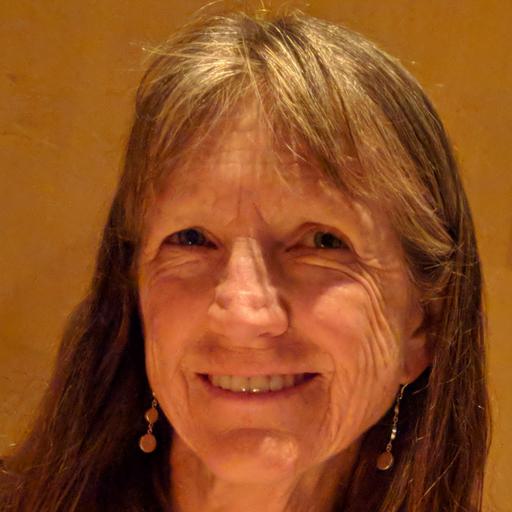}} &
        \raisebox{\rbx}{\includegraphics[width=\imW]{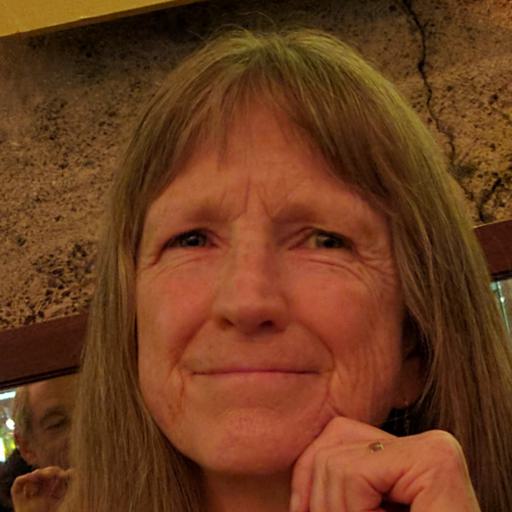}} &
        \raisebox{\rbx}{\includegraphics[width=\imW]{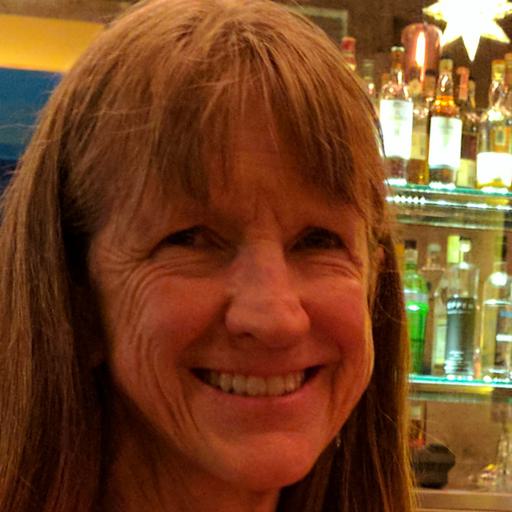}} &
        \raisebox{\rbx}{\includegraphics[width=\imW]{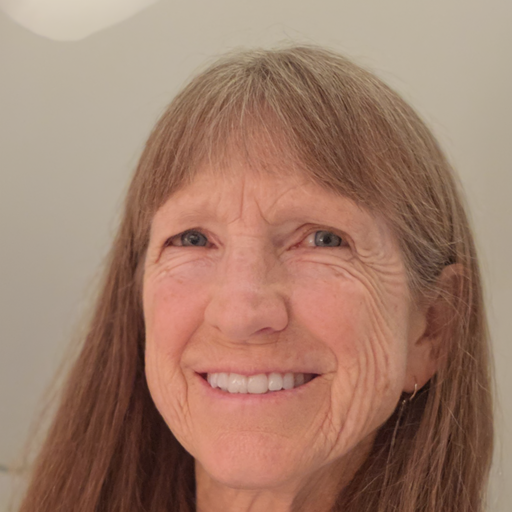}} &
        \raisebox{\rbx}{\includegraphics[width=\imW]{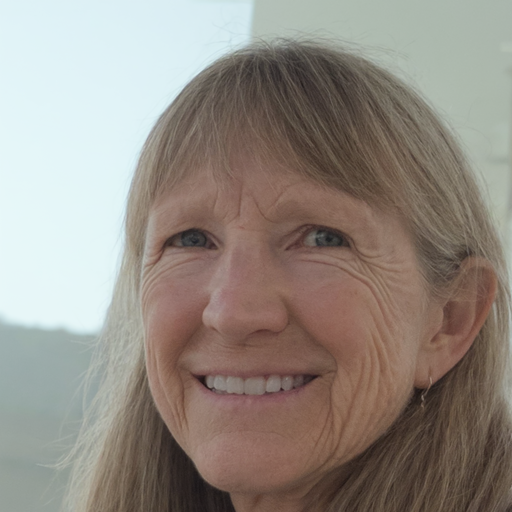}} &
        \raisebox{\rbx}{\includegraphics[width=\imW]{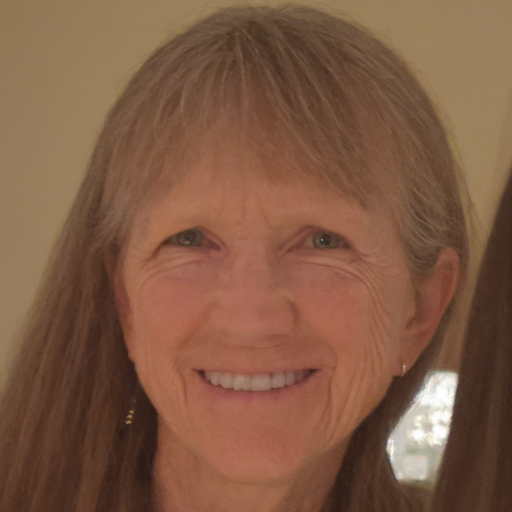}} &
        \raisebox{\rbx}{\includegraphics[width=\imW]{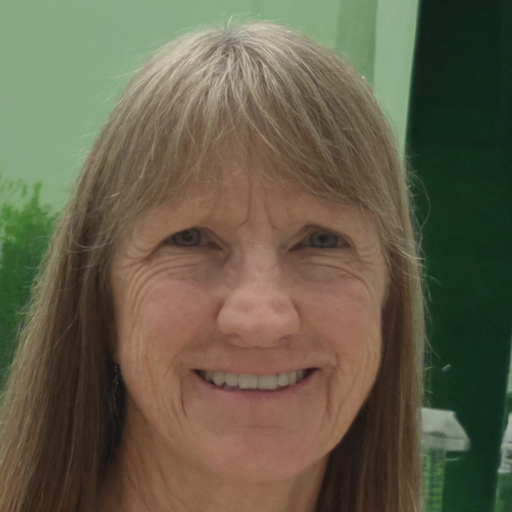}} \\
        \multicolumn{4}{c}{Personal Album for constraining the prior}
         & \multicolumn{4}{c}{Generated Images using the constrained prior} \\
    \end{tabular}
    }
    \caption{\textbf{Personal Album \& Unconditional Generation from Personalized Model.} We provide two sets of results.}
    \label{fig:anchor_personal}
\end{figure*}

Regarding the personalized album, we collect around 20 real high-quality images to act as anchor images. Examples of these images, along with images randomly generated by the personalized model, are presented in Figure~\ref{fig:anchor_personal}. The images produced by the personalized model exhibit both diversity and identity preservation, attributes learned from the personal album.

\section{More Ablation Studies}
\vspace{1mm}

\begin{figure*}[h]
    \centering
    \setlength{\tabcolsep}{1pt}
    \def\imW{0.14\linewidth}
    \scalebox{0.9}{
    \begin{tabular}{ccccccccc}
        \multirow[c]{2}{*}[-2mm]{{\includegraphics[width=\imW]{Figs/ablation/0000_input.png}}} &
        \raisebox{-.5\height}{\includegraphics[width=\imW]{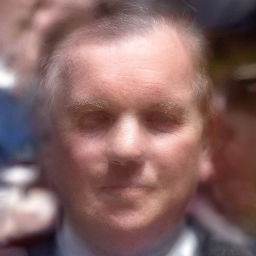}} &
        \raisebox{-.5\height}{\includegraphics[width=\imW]{Figs/ablation/0000_n200.png}} &
        \raisebox{-.5\height}{\includegraphics[width=\imW]{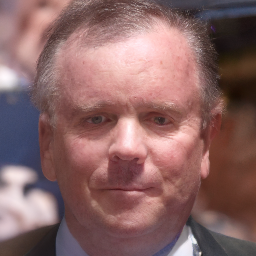}} &
        \raisebox{-.5\height}{\includegraphics[width=\imW]{Figs/ablation/0000_n400.png}} &
        \raisebox{-.5\height}{\includegraphics[width=\imW]{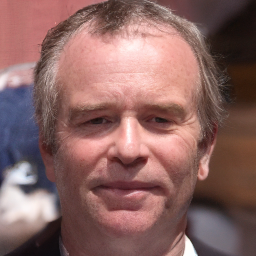}} &
        \raisebox{-.5\height}{\includegraphics[width=\imW]{Figs/ablation/0000_n600.png}} & \rotatebox[origin=c]{90}{\footnotesize w/o Constraining} \vspace{3pt} \\
        &
        \raisebox{-.5\height}{\includegraphics[width=\imW]{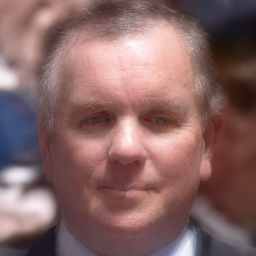}} &
        \raisebox{-.5\height}{\includegraphics[width=\imW]{Figs/ablation/0000_c200.png}} &
        \raisebox{-.5\height}{\includegraphics[width=\imW]{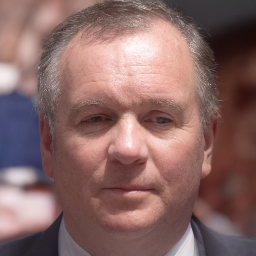}} &
        \raisebox{-.5\height}{\includegraphics[width=\imW]{Figs/ablation/0000_c400.png}} &
        \raisebox{-.5\height}{\includegraphics[width=\imW]{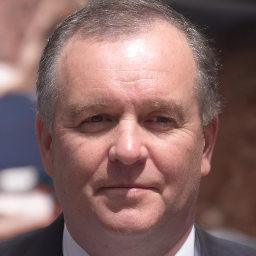}} &
        \raisebox{-.5\height}{\includegraphics[width=\imW]{Figs/ablation/0000_c600.png}} & \rotatebox[origin=c]{90}{\footnotesize w/ Constraining} \vspace{3pt} \\
        \multirow[c]{2}{*}[-2mm]{{\includegraphics[width=\imW]{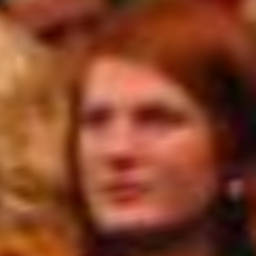}}} &
        \raisebox{-.5\height}{\includegraphics[width=\imW]{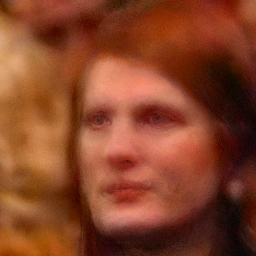}} &
        \raisebox{-.5\height}{\includegraphics[width=\imW]{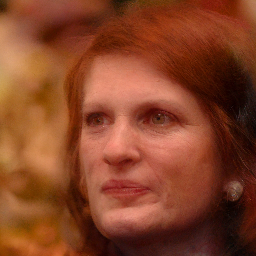}} &
        \raisebox{-.5\height}{\includegraphics[width=\imW]{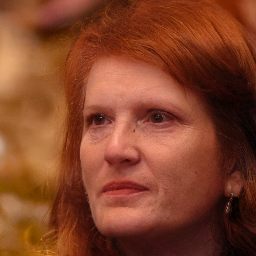}} &
        \raisebox{-.5\height}{\includegraphics[width=\imW]{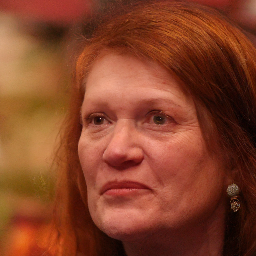}} &
        \raisebox{-.5\height}{\includegraphics[width=\imW]{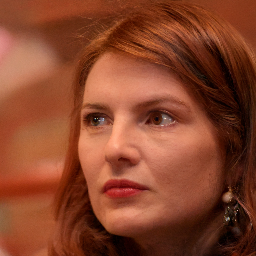}} &
        \raisebox{-.5\height}{\includegraphics[width=\imW]{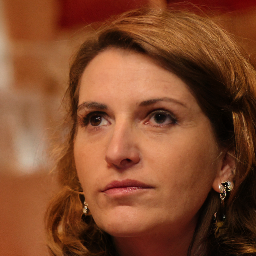}} & \rotatebox[origin=c]{90}{\footnotesize w/o Constraining} \vspace{3pt} \\
        &
        \raisebox{-.5\height}{\includegraphics[width=\imW]{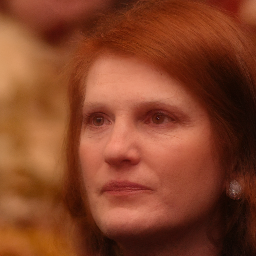}} &
        \raisebox{-.5\height}{\includegraphics[width=\imW]{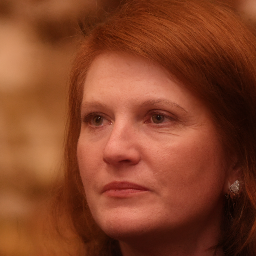}} &
        \raisebox{-.5\height}{\includegraphics[width=\imW]{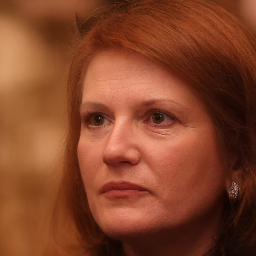}} &
        \raisebox{-.5\height}{\includegraphics[width=\imW]{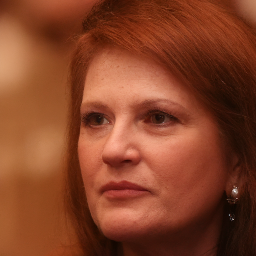}} &
        \raisebox{-.5\height}{\includegraphics[width=\imW]{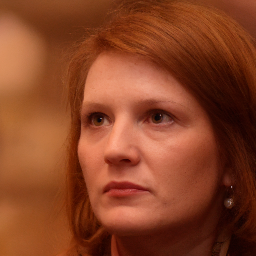}} &
        \raisebox{-.5\height}{\includegraphics[width=\imW]{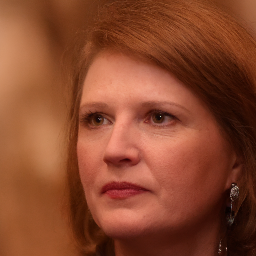}} & \rotatebox[origin=c]{90}{\footnotesize w/ Constraining} \vspace{3pt} \\
        \multirow[c]{2}{*}[-2mm]{{\includegraphics[width=\imW]{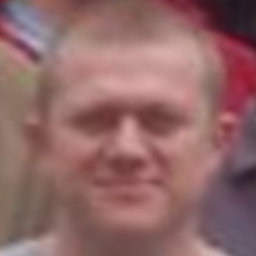}}} &
        \raisebox{-.5\height}{\includegraphics[width=\imW]{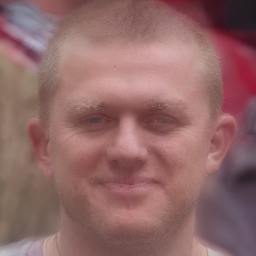}} &
        \raisebox{-.5\height}{\includegraphics[width=\imW]{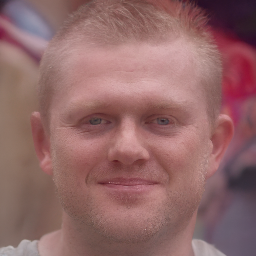}} &
        \raisebox{-.5\height}{\includegraphics[width=\imW]{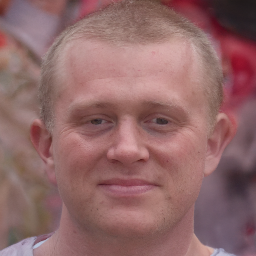}} &
        \raisebox{-.5\height}{\includegraphics[width=\imW]{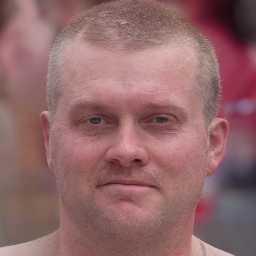}} &
        \raisebox{-.5\height}{\includegraphics[width=\imW]{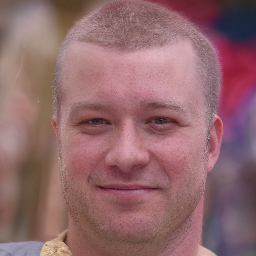}} &
        \raisebox{-.5\height}{\includegraphics[width=\imW]{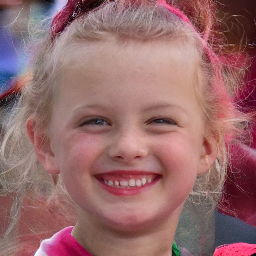}} & \rotatebox[origin=c]{90}{\footnotesize w/o Constraining} \vspace{3pt} \\
        &
        \raisebox{-.5\height}{\includegraphics[width=\imW]{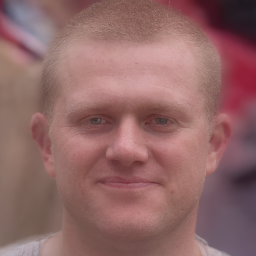}} &
        \raisebox{-.5\height}{\includegraphics[width=\imW]{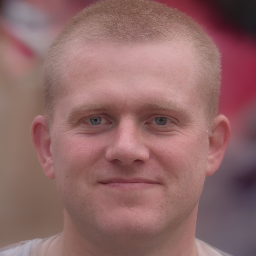}} &
        \raisebox{-.5\height}{\includegraphics[width=\imW]{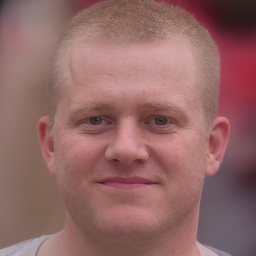}} &
        \raisebox{-.5\height}{\includegraphics[width=\imW]{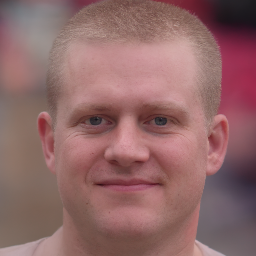}} &
        \raisebox{-.5\height}{\includegraphics[width=\imW]{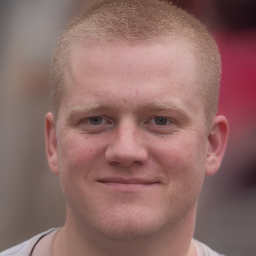}} &
        \raisebox{-.5\height}{\includegraphics[width=\imW]{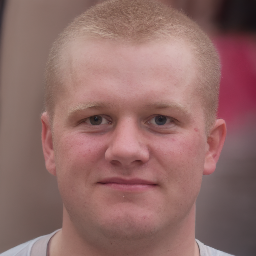}} & \rotatebox[origin=c]{90}{\footnotesize w/ Constraining} \vspace{3pt} \\
        Input & $K=100$ & $K=200$ & $K=300$ & $K=400$ & $K=500$ & $K=600$ \\
    \end{tabular}
    }
    \caption{\textbf{More Results of Ablation on Noise Step \( K \) and Constraining with Generative Album.}}
    \label{fig:sup_ab_generative}
\end{figure*}

\paragraph{Noise Step $K$ and Constraining Prior with Generative Album.} Here we provide supplemental results to Figure~\textcolor{red}{8} from the main paper, which analyzes the effect of noise step $K$ and the effectiveness of using a generative album to constrain the prior. Results are shown in Figure~\ref{fig:sup_ab_generative}. From the results we can see that as K increases both results using either the constrained prior or not would have better quality but would not be less faithful to the input image. However, with constrained prior, we can see that the loss in faithfulness is considerably less than the ones not using the constrained prior. 

\begin{figure}[ht]
\centering
\setlength{\tabcolsep}{0.2pt}
\def\imW{0.19\linewidth}
\begin{tabular}{ccccc}
\includegraphics[width=\imW]{Figs/Personal_Obama/00_input.png}
& 
\includegraphics[width=\imW]{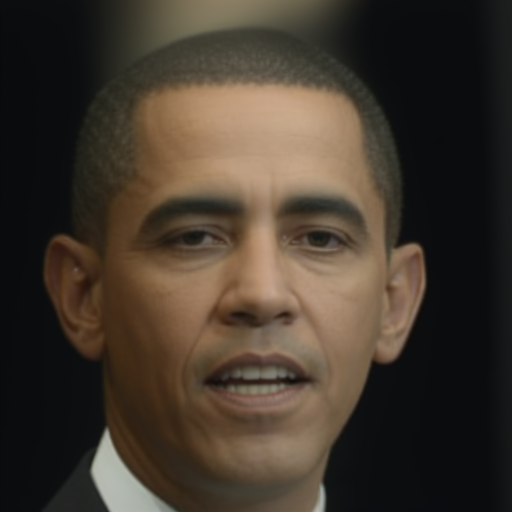}
&
\includegraphics[width=\imW]{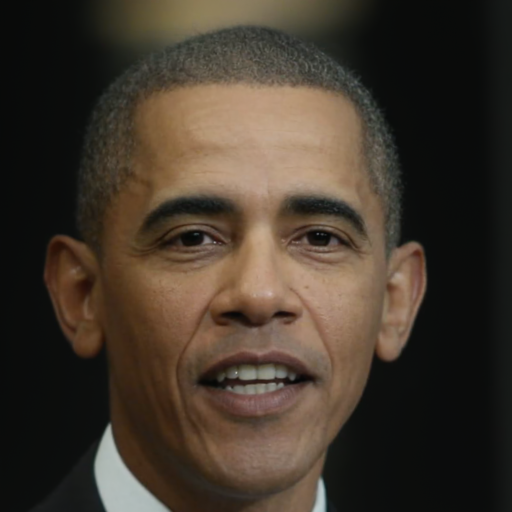}
&
\includegraphics[width=\imW]{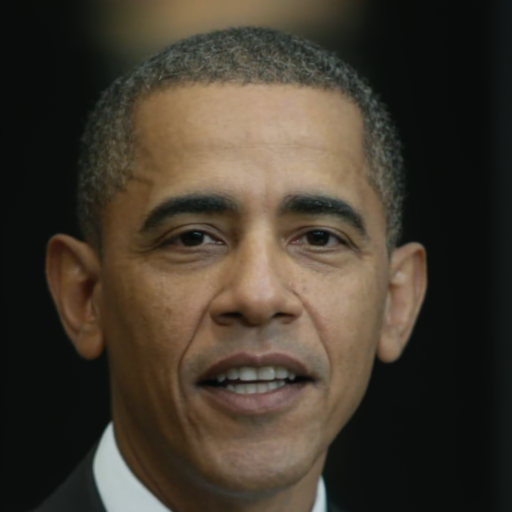}
&
\includegraphics[width=\imW]{Figs/Personal_Obama/00_ours.png}
\\
Input & Size=1 & Size=4 & Size=8 & Size=16
\end{tabular}
\vspace{-2mm}
\caption{Ablation on Size of Personal Album.}
\label{fig:ab_size_personal_album}
\vspace{-2mm}
\end{figure}

\begin{figure*}[h]
    \centering
    \setlength{\tabcolsep}{1pt}
    \def\imW{0.13\linewidth}
    \scalebox{0.9}{
    \begin{tabular}{cccccccc}
        \raisebox{-.0\height}{\includegraphics[width=\imW]{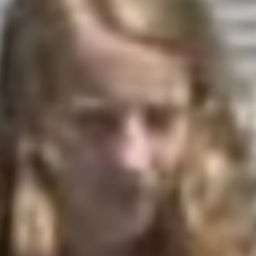}} & 
        \raisebox{-.0\height}{\includegraphics[width=\imW]{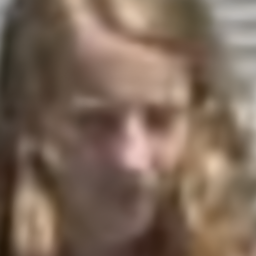}}
        &
        \raisebox{-.0\height}{\includegraphics[width=\imW]{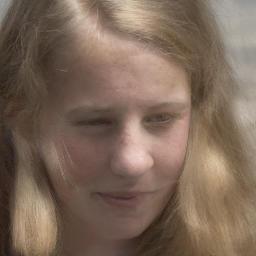}} 
        &
        \raisebox{-.0\height}{\includegraphics[width=\imW]{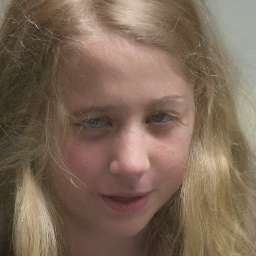}} 
        &
        \raisebox{-.0\height}{\includegraphics[width=\imW]{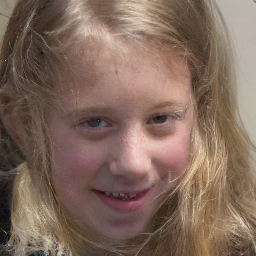}} 
        &
        \raisebox{-.0\height}{\includegraphics[width=\imW]{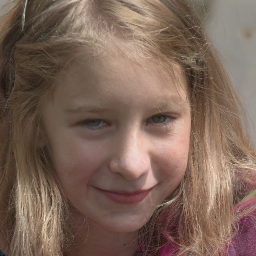}} 
        &
        \raisebox{-.0\height}{\includegraphics[width=\imW]{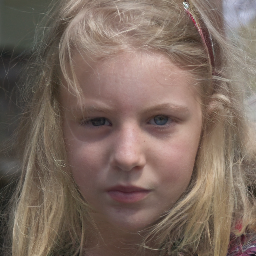}} 
        &
        \raisebox{-.0\height}{\includegraphics[width=\imW]{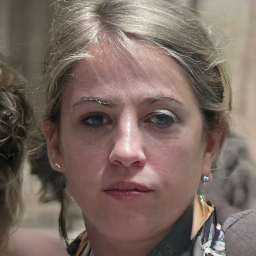}}
        \\
        \raisebox{-.0\height}{\includegraphics[width=\imW]{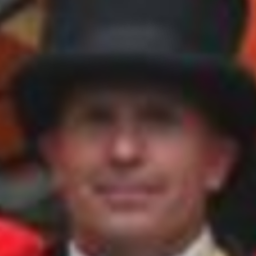}} &
        \raisebox{-.0\height}{\includegraphics[width=\imW]{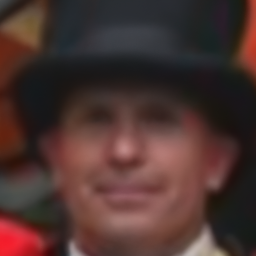}}
        &
        \raisebox{-.0\height}{\includegraphics[width=\imW]{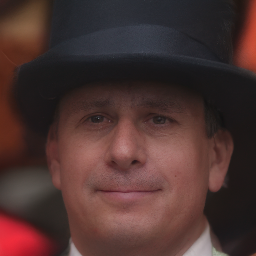}} 
        &
        \raisebox{-.0\height}{\includegraphics[width=\imW]{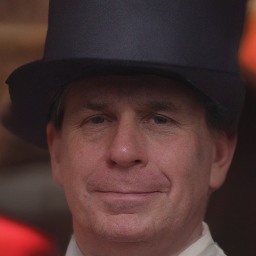}} 
        &
        \raisebox{-.0\height}{\includegraphics[width=\imW]{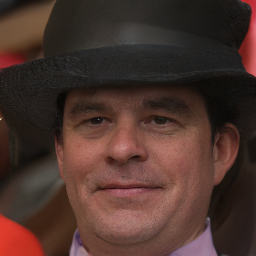}} 
        &
        \raisebox{-.0\height}{\includegraphics[width=\imW]{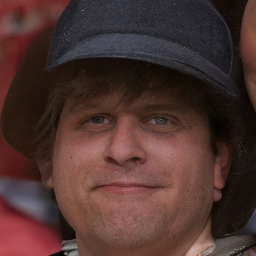}} 
        &
        \raisebox{-.0\height}{\includegraphics[width=\imW]{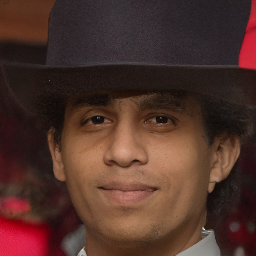}} 
        &
        \raisebox{-.0\height}{\includegraphics[width=\imW]{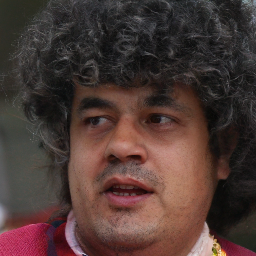}} 
        \\
        {\footnotesize Input} & {\footnotesize $n=1$} & {\footnotesize $n=5$} & {\footnotesize $n=10$} & {\footnotesize $n=20$} & {\footnotesize $n=30$}  & {\footnotesize $n=50$} & {\footnotesize No Guidance}\\
    \end{tabular}
    }
    \caption{\textbf{The impact of skip guidance frequency on the generative album.} Absence of skip guidance results in divergence from the input, while overly frequent guidance produces low-quality anchor images.}
    
    \label{fig:sup_ab_skip}
\end{figure*}

\paragraph{Skip Guidance for Generative Album.} We analyze the effectiveness of our proposed Skip Guidance in generating a generative album from a degraded input image. The album should contain images close to the input yet of high quality, serving as anchor images for the constrained generative space. Figure~\ref{fig:sup_ab_skip} shows that without guidance (i.e., direct sampling of the album from $y_K$), we obtain high-quality images that do not closely resemble the input, thus failing to effectively constrain the generative space. Conversely, applying skip guidance too frequently lowers sample quality due to the guidance's approximate nature, potentially leading to a constrained space filled with low-quality images.
\vspace{-3mm}
\paragraph{Size of Personal Album.} In Fig.\ref{fig:ab_size_personal_album}, we present results using personal albums of varying sizes. Generally, larger albums enable the model to better preserve identity and details, though the improvements diminish with size increase.

\section{Theoretical Analysis}

In this section, we provide the theoretical analysis for the intuition we claimed in the paper. 
\begin{compactenum}
    \item \textit{Adding noise to high-quality and low-quality images can progressively align their distributions, making them more similar over time.}

For a clean image $x_0$ and low-quality image $y_0$, we can analyze the distribution of their noisy versions: $q(x_t|x_0)$ and $q(y_t|y_0)$ which are
\begin{align}
    q(x_t|x_0) &\sim N(\sqrt{\Bar{\alpha}_t}x_0, (1-\Bar\alpha_t)\mathbf{I})\\
    q^\prime(y_t|y_0) &\sim N(\sqrt{\Bar{\alpha}_t}y_0, (1-\Bar\alpha_t)\mathbf{I})
\end{align}
in which $\Bar\alpha_t$ is a hyper-parameter in the diffusion process and will decrease as $t$ increases. Therefore we can compute the KL divergence between these two distributions:
\begin{align}
    KL(q, q^\prime)=\frac{\Bar{\alpha}_t}{2(1-\Bar{\alpha}_t)}(x_0-y_0)^2.
\end{align}
When more noise is added ($t\uparrow$), $KL(q,q^\prime)$ will decrease. Therefore the two distributions get more similar as more noise is added. 
    \item \textit{The larger t is, the larger the generative space $p(x_0|x_t)$ spans.} 

    In this case, we compute the entropy $H$ of $p(x_0|x_t)$ to show how large the generative space spans. Let's consider $q(x_t|x_0)$ first:
    \begin{align}
        q(x_{t}|x_0) \sim N(\sqrt{\Bar{\alpha}_{t}}x_0, (1-\Bar\alpha_{t})\mathbf{I})
    \end{align}
    Thus we can compute the entropy:
    \begin{align}
        H(q(x_{t}|x_0))=\frac{1}{2}\log(2\pi(1-\Bar{\alpha}_t))+\frac{1}{2}
    \end{align}
    For a deterministic denoising process (such as DDIM), we can have $H(p(x_{0}|x_t))=H(q(x_{t}|x_0))$. Therefore $H(p(x_{0}|x_t))$ increases as $t$ increases, showing the generative space spans larger.
\end{compactenum}

\section{Implementation Details}
\vspace{1mm}

We provide the model details trained on datasets (256$\times$256 and 512$\times$512) along with the training/inference parameters in Table \ref{tab:sup_details}. Due to the lack of 512$\times$512 model trained with diffusion models and its slow speed in both training and inference, we use 256$\times$256 for standard benchmarks while for personalized restoration, we utilize a 512$\times$512 model. 

We first train an unconditional generative model using the model architecture based on \cite{dhariwal2021diffusion, nichol2021improved}. After this, we will get a powerful generative prior that can output high-quality images. Then we finetune the model using either the generative album or the personal album. For the generative album, we first generate the images with the skip guidance to ensure that our images follow the input. Then we finetune the model using either the generative album or the personal album. Finally, we restore the images using the constrained prior.

\begin{table}[htbp]
    \centering
    \scalebox{0.7}{
    \begin{tabular}{lcc}
    \toprule
        & $256\times 256$ & $512\times 512$ \\
        \hline 
    \textbf{Model Details} & & \\
    Diffusion Steps & $1000$ & $1000$ \\
    Channels & $128$ & $256$ \\
    Channels Multiple & $1,1,2,2,4,4$ & $0.5,1,1,2,2,4,4$ \\
    Heads Channels & $128$ & $64$ \\
    Attention Resolution & $16$ & $32,16,8$ \\
    Dropout & $0.1$ & $0.1$ \\
    \textbf{Training Details} & & \\
    Batch Size$^{[1]}$ & $256$ & $32$ \\
    Iterations & $200$k & $2320$k \\
    Learning Rate & $10^{-4}$ & $10^{-4}$ \\
    Optimizer & Adam & Adam \\
    Weight Decay & $0.0$ & $0.0$ \\
    \textbf{Generative Album} & & \\
    Noise Step $K$ & 600 & - \\
    Skip Guidance & 20 & - \\ 
    \textbf{Finetuning Details} & & \\
    Batch Size & $4$ & $4$ \\
    Iterations$^{[2]}$ & $3000$ & $5000$ \\
    Learning Rate & $10^{-5}$ & $10^{-5}$ \\
    \textbf{Inference Details} & & \\
    Noise Step $K$ & 200 & 300 \\
    \bottomrule
    \end{tabular}}
    \caption{\textbf{Implementation details}. ${[1]}$ for AFHQ-Dog and AFHQ-Cat (256$\times$256), the iterations are 50k and 100k respectively. ${[2]}$ for personalized finetuning, we use 5000 iterations. }
    \label{tab:sup_details}
\end{table}

\newpage

\end{document}